\documentclass[11pt]{article}

\usepackage[]{acl}

\usepackage{times}
\usepackage{latexsym}
\usepackage[T1]{fontenc}
\usepackage[utf8]{inputenc}
\usepackage{microtype}
\usepackage{graphicx}

\DeclareUnicodeCharacter{00D7}{\ensuremath{\times}}
\DeclareUnicodeCharacter{03BB}{\ensuremath{\lambda}}
\DeclareUnicodeCharacter{0394}{\ensuremath{\Delta}}
\DeclareUnicodeCharacter{03A9}{\ensuremath{\Omega}}
\DeclareUnicodeCharacter{2115}{\ensuremath{\mathbb{N}}}
\DeclareUnicodeCharacter{211D}{\ensuremath{\mathbb{R}}}
\DeclareUnicodeCharacter{2190}{\ensuremath{\leftarrow}}
\DeclareUnicodeCharacter{2192}{\ensuremath{\to}}
\DeclareUnicodeCharacter{21D2}{\ensuremath{\Rightarrow}}
\DeclareUnicodeCharacter{2200}{\ensuremath{\forall}}
\DeclareUnicodeCharacter{2194}{\ensuremath{\leftrightarrow}}
\DeclareUnicodeCharacter{03B1}{\ensuremath{\alpha}}
\DeclareUnicodeCharacter{2026}{\ensuremath{\ldots}}
\DeclareUnicodeCharacter{2203}{\ensuremath{\exists}}
\DeclareUnicodeCharacter{2208}{\ensuremath{\in}}
\DeclareUnicodeCharacter{2218}{\ensuremath{\circ}}
\DeclareUnicodeCharacter{2227}{\ensuremath{\land}}
\DeclareUnicodeCharacter{2264}{\ensuremath{\leq}}
\DeclareUnicodeCharacter{27E8}{\ensuremath{\langle}}
\DeclareUnicodeCharacter{27E9}{\ensuremath{\rangle}}
\DeclareUnicodeCharacter{27F6}{\ensuremath{\longrightarrow}}
\DeclareUnicodeCharacter{2014}{---}
\DeclareUnicodeCharacter{00A7}{\S}
\DeclareUnicodeCharacter{2016}{\ensuremath{\|}}
\DeclareUnicodeCharacter{207B}{\ensuremath{^{-}}}
\DeclareUnicodeCharacter{00B9}{\ensuremath{^{1}}}
\DeclareUnicodeCharacter{222B}{\ensuremath{\int}}
\DeclareUnicodeCharacter{2022}{\ensuremath{\bullet}}
\DeclareUnicodeCharacter{2202}{\ensuremath{\partial}}
\DeclareUnicodeCharacter{03BC}{\ensuremath{\mu}}
\DeclareUnicodeCharacter{2102}{\ensuremath{\mathbb{C}}}
\DeclareUnicodeCharacter{03C0}{\ensuremath{\pi}}
\DeclareUnicodeCharacter{2229}{\ensuremath{\cap}}
\DeclareUnicodeCharacter{2260}{\ensuremath{\neq}}

\usepackage{xcolor,colortbl,caption}
\usepackage{fontawesome5}             
\definecolor{affilSNU}{HTML}{000080}  
\definecolor{affilSejong}{HTML}{D6001C}  
\definecolor{affilETRI}{HTML}{0B6E4F} 
\definecolor{affilUMD}{HTML}{9E1B32}  
\definecolor{affilAWS}{HTML}{B8620A}  

\usepackage[most]{tcolorbox}
\usepackage{comment}
\usepackage[record,abbreviations]{glossaries-extra}
\glsdisablehyper
\usepackage{amsmath}
\usepackage{pifont}
\usepackage{tabularx}
\usepackage{caption, booktabs}
\usepackage{algorithm}
\usepackage{amssymb}
\usepackage{arydshln}
\usepackage{makecell}

\usepackage{cleveref}

\crefformat{section}{\S#2#1#3}
\crefformat{subsection}{\S#2#1#3}

\usepackage{todonotes}

\usepackage{stmaryrd}

\definecolor{RevRed}{RGB}{180,30,30}
\definecolor{RevBlue}{RGB}{30,60,180}
\definecolor{RevGreen}{RGB}{20,120,60}

\newif\ifarxivonly
\arxivonlytrue

\makeatletter
\newcommand\footnoteref[1]{\protected@xdef\@thefnmark{\ref{#1}}\@footnotemark}
\makeatother
\newcolumntype{P}[1]{>{\centering\arraybackslash}p{#1}}

\usepackage{xspace}

\newcommand{\ours}[0]{\textsc{ShadowBench}\xspace}
\DeclareRobustCommand{\ourrelaxed}{%
  \ifmmode
    \operatorname{SA\text{-}Pass}_{\mathrm{soft}}%
  \else
    \textsc{SA-Pass}$_{\mathrm{soft}}$\xspace
  \fi
}
\DeclareRobustCommand{\ourmetric}{%
  \ifmmode
    \operatorname{SA\text{-}Pass}%
  \else
    \textsc{SA-Pass}\xspace
  \fi
}
\newtcbox{\codebox}{on line, boxrule=0pt, boxsep=0pt, colback=gray!10, colframe=gray!10, arc=2pt, left=2pt, right=2pt, top=1pt, bottom=1pt}
\newcommand{\code}[1]{\codebox{\texttt{#1}}}

\usepackage{graphicx}
\usepackage{adjustbox}
\makeatletter
\newcommand{\thickhline}{
\noalign {\ifnum 0=`}\fi \hrule height 1pt
    \futurelet \reserved@a \@xhline
}
\newcolumntype{"}{@{\hskip\tabcolsep\vrule width 1pt\hskip\tabcolsep}}
\makeatother
\usepackage{amsfonts}
\usepackage{multirow}
\newcommand{\edit}[1]{#1}

\usepackage{mathtools}

\definecolor{my_blue}{RGB}{0,112,192}

\usepackage{placeins}

\usepackage{tikz}

\usepackage{fvextra}
\usepackage{minted}
\makeatletter
\@ifundefined{setminted}{%
  \usemintedstyle{colorful}
}{%
  \setminted{style=colorful}
}
\makeatother

\DefineVerbatimEnvironment{WideMinted}{Verbatim}
{breaklines, fontsize=\small, frame=lines, linenos, breakanywhere, xleftmargin=0pt, xrightmargin=0pt, width=\textwidth}

\usepackage{stackengine}

\usepackage{amsthm}
\newtheorem{definition}{Definition}

\theoremstyle{definition}
\newtheorem{example}{Example}
\crefname{example}{Example}{Examples}
\Crefname{example}{Example}{Examples}
\theoremstyle{plain}

\newcommand{\HPE}{\texttt{HasProjectiveEmbedding}}
\usetikzlibrary{arrows.meta,positioning,calc,fit,backgrounds}

\makeatletter
\def\th@definition{%
    \normalfont 
    \thm@headpunct{.}
}
\makeatother

\usepackage{enumitem}
\usepackage{wrapfig}
\usepackage{subcaption}

\usepackage{listings}

\definecolor{codegray}{rgb}{0.5,0.5,0.5}
\definecolor{codepurple}{rgb}{0.58,0,0.82}
\definecolor{backcolour}{rgb}{0.95,0.95,0.92}

\lstdefinelanguage{Lean}{
keywords={def, theorem, lemma, example, where, match, with, end, class, instance, structure, inductive},
keywordstyle=\color{blue}\bfseries,
ndkeywords={Type, Prop, Nat},
ndkeywordstyle=\color{teal}\bfseries,
identifierstyle=\color{black},
sensitive=true,
comment=[l]{--},
morecomment=[s]{/-}{-/},
commentstyle=\color{codegray}\ttfamily,
stringstyle=\color{red}\ttfamily,
morestring=[b]",
basicstyle=\ttfamily\footnotesize,
breaklines=true,
keepspaces=true,
showstringspaces=false,
frame=none,
backgroundcolor=\color{backcolour},
numbers=none,
literate={{<}}{{$\langle$}}1 {{>}}{{$\rangle$}}1 
}

\lstdefinestyle{promptstyle}{
    backgroundcolor=\color{backcolour},
    commentstyle=\color{codegray},
    keywordstyle=\color{magenta},
    numberstyle=\tiny\color{codegray},
    stringstyle=\color{codepurple},
    basicstyle=\ttfamily\footnotesize,
    breakatwhitespace=false,
    breaklines=true,
    captionpos=b,
    keepspaces=true,
    numbers=none,
    numbersep=5pt,
    showspaces=false,
    showstringspaces=false,
    showtabs=false,
    tabsize=2,
    literate={α}{{$\alpha$}}1 {λ}{{$\lambda$}}1 {Δ}{{$\Delta$}}1
      {Ω}{{$\Omega$}}1 {μ}{{$\mu$}}1 {π}{{$\pi$}}1
      {ℕ}{{$\mathbb{N}$}}1 {ℝ}{{$\mathbb{R}$}}1 {ℂ}{{$\mathbb{C}$}}1
      {→}{{$\to$}}1 {←}{{$\leftarrow$}}1 {↔}{{$\leftrightarrow$}}1
      {⇒}{{$\Rightarrow$}}1 {⟶}{{$\longrightarrow$}}2
      {∀}{{$\forall$}}1 {∃}{{$\exists$}}1 {∈}{{$\in$}}1
      {∧}{{$\land$}}1 {∘}{{$\circ$}}1 {≤}{{$\leq$}}1 {≠}{{$\neq$}}1
      {∩}{{$\cap$}}1 {×}{{$\times$}}1
      {⟨}{{$\langle$}}1 {⟩}{{$\rangle$}}1
      {‖}{{$\|$}}1 {∫}{{$\int$}}1 {∂}{{$\partial$}}1 {•}{{$\bullet$}}1
      {⁻}{{${}^{-}$}}1 {¹}{{${}^{1}$}}1
      {—}{{\textemdash}}1 {…}{{\ldots}}1
}

\tcbuselibrary{listings, breakable}

\newtcolorbox{systemprompt}{
    colback=red!5!white,
    colframe=red!75!black,
    title=System Prompt,
    fonttitle=\bfseries,
    boxrule=0.5mm,
    arc=2mm,
    left=2mm, right=2mm, top=2mm, bottom=2mm,
    breakable
}

\newtcolorbox{userprompt}{
    colback=blue!5!white,
    colframe=blue!75!black,
    title=User Prompt,
    fonttitle=\bfseries,
    boxrule=0.5mm,
    arc=2mm,
    left=2mm, right=2mm, top=2mm, bottom=2mm,
    breakable
}

\newtcolorbox{modelresponse}{
    colback=green!5!white,
    colframe=green!75!black,
    title=AI Response,
    fonttitle=\bfseries,
    boxrule=0.5mm,
    arc=2mm,
    left=2mm, right=2mm, top=2mm, bottom=2mm,
    breakable
}
\newtcblisting{leancode}{
    listing only,
    breakable,
    colback=white,
    colframe=gray!50,
    colbacktitle=gray!20,
    coltitle=black,
    title={\scriptsize\textbf{Lean 4}},
    fonttitle=\sffamily,
    boxrule=0.5pt,
    arc=1pt,
    left=1mm, right=1mm, top=1mm, bottom=1mm,
    listing options={
            language=Lean,
            style=promptstyle,
        }
}

\newtcblisting{textcode}{
    listing only,
    breakable,
    colback=white,
    colframe=gray!50,
    colbacktitle=gray!20,
    coltitle=black,
    title={\scriptsize\textbf{Natural Language}},
    fonttitle=\sffamily,
    boxrule=0.5pt,
    arc=1pt,
    left=1mm, right=1mm, top=1mm, bottom=1mm,
    listing options={
            style=promptstyle,
        }
}

\newtcblisting{userpromptcode}{
    listing only,
    breakable,
    colback=blue!5!white,
    colframe=blue!75!black,
    title=User Prompt,
    fonttitle=\bfseries,
    boxrule=0.5mm,
    arc=2mm,
    left=2mm, right=2mm, top=2mm, bottom=2mm,
    listing options={
            style=promptstyle,
        }
}

\usepackage{stfloats}

\title{\ours: Toward Reliable Automatic Evaluation of \\Semantic Alignment in Autoformalization}

\newcommand{\blfootnote}[1]{%
  \begingroup
  \renewcommand{\thefootnote}{}%
  \footnote{#1}%
  \addtocounter{footnote}{-1}%
  \endgroup
}

\author{
    \textbf{Hojae Han}\textsuperscript{\textcolor{affilETRI}{\rm 1}}\thanks{These authors contributed equally to this work.}~
    \textbf{Jongyoon Kim}\textsuperscript{\textcolor{affilSNU}{\rm 2}}\footnotemark[1]~
    \textbf{Sanghyeok Park}\textsuperscript{\textcolor{affilSNU}{\rm 2}}\footnotemark[1]~
    \textbf{Dongwook Cheon}\textsuperscript{\textcolor{affilSNU}{\rm 2}}~
    \textbf{Yeachan Park}\textsuperscript{\textcolor{affilSejong}{\rm 3}}\\
    \textbf{Myeong Jae Jeon}\textsuperscript{\textcolor{affilUMD}{\rm 4}}~
    \textbf{Sunjong Choe}\textsuperscript{\textcolor{affilSNU}{\rm 2}}~
    \textbf{Soonho Kong}\textsuperscript{\textcolor{affilAWS}{\rm 5}}~
    \textbf{Wonseok Hur}\textsuperscript{\textcolor{affilSNU}{\rm 2}}\\
    \textbf{Seung-won Hwang}\textsuperscript{\textcolor{affilSNU}{\rm 2}}\thanks{Corresponding authors.}~
    \textbf{Donghoon Hyeon}\textsuperscript{\textcolor{affilSNU}{\rm 2}}\footnotemark[2]\\[3pt]
    {\normalfont
    \textsuperscript{\textcolor{affilETRI}{\rm 1}}Electronics and Telecommunications Research Institute$\quad$
    \textsuperscript{\textcolor{affilSNU}{\rm 2}}Seoul National University}\\
    {\normalfont
    \textsuperscript{\textcolor{affilSejong}{\rm 3}}Sejong University$\quad$
    \textsuperscript{\textcolor{affilUMD}{\rm 4}}University of Maryland, College Park$\quad$
    \textsuperscript{\textcolor{affilAWS}{\rm 5}}Amazon Web Services}\\[3pt]
    {\normalfont
    \textcolor{affilSNU}{\small\faEnvelope}~\texttt{hojae.han@etri.re.kr}}\\
    {\normalfont
    \textcolor{affilSNU}{\small\faEnvelope}~\texttt{\{john.jongyoon.kim,202123018,seungwonh,dhyeon\}@snu.ac.kr}}
}

\begin{document}

\maketitle

\blfootnote{Contact information for all authors is provided in \Cref{app:contact}.}

\begin{abstract}
Autoformalization translates informal mathematical theorems into code for proof assistants such as Lean.
A central challenge is that current evaluation metrics can accept type-correct but misaligned statements or reject correct statements written in a different formulation.
Inspired by Pass@$k$, we propose \ourmetric (\emph{Semantic Alignment Pass}), which tests formal statements using auxiliary statements called \emph{shadows} that characterize the intended statement.
A generated statement receives full credit only when it compiles, implies each shadow (forward check), and is implied by their conjunction (backward check).
We instantiate \ourmetric in \ours, a Lean~4 full autoformalization benchmark of 178 postgraduate- to research-level problems spanning eight mathematical areas.
Claude Code (Opus~4.8) with Numina-Lean-Agent reaches $61.8\%$ compile rate and $11.2\%$ \ourmetric.
\edit{Across outputs generated by six agentic configurations, \ourmetric achieves $98.8\%$ binary agreement with expert judgments.}
An early version of \ours served as the benchmark for Track~4 of the ICML 2026
AI4Math Challenge.\footnote{\url{https://github.com/ldilab/shadowbench}}
\end{abstract}

\section{Introduction}
\label{sec:intro}

Autoformalization translates informal mathematics written in natural language into code for proof assistants
such as Lean.\footnote{\url{https://lean-lang.org/}}
In the full autoformalization setting, the input is an informal theorem statement
and proof, and the output is code containing the corresponding formal statement
and proof.
The proof assistant can then mechanically check whether the formal proof proves
the formal statement.

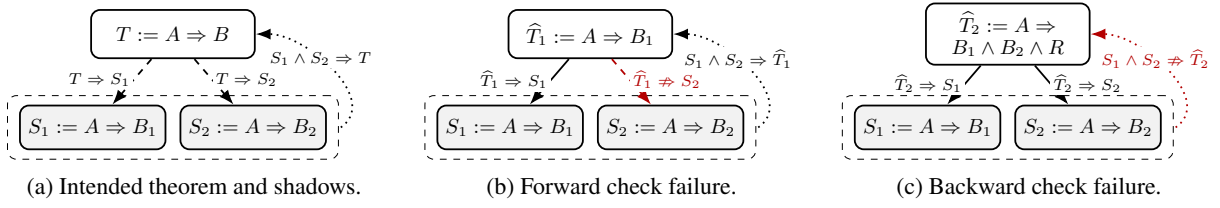
\begin{figure*}[t]
\centering
\begin{subfigure}[t]{0.31\linewidth}
\centering
\resizebox{\linewidth}{!}{%
\begin{tikzpicture}[
  every node/.style={font=\small},
  stmt/.style={draw, rounded corners, align=center, minimum width=2.55cm,
    minimum height=0.75cm},
  shadow/.style={draw, rounded corners, align=center, minimum width=2.2cm,
    minimum height=0.68cm, fill=gray!10},
  edgeLabel/.style={font=\scriptsize, fill=white, inner sep=1.2pt},
  >=Latex, thick
]
  \node[stmt] (T) at (0, 1.45) {$T := A \Rightarrow B$};
  \node[shadow] (S1) at (-1.25, 0) {$S_1 := A \Rightarrow B_1$};
  \node[shadow] (S2) at (1.25, 0) {$S_2 := A \Rightarrow B_2$};
  \begin{scope}[on background layer]
    \node[draw, rounded corners, dashed, inner xsep=0.18cm, inner ysep=0.14cm,
          fit=(S1)(S2)] (group) {};
  \end{scope}
  \draw[->, dashed] (T) -- node[edgeLabel, left, pos=0.45] {$T \Rightarrow S_1$} (S1);
  \draw[->, dashed] (T) -- node[edgeLabel, right, pos=0.45] {$T \Rightarrow S_2$} (S2);
  \draw[->, dotted]
      ($(group.east)+(0.03,0)$) to[out=35, in=0]
      node[edgeLabel, pos=0.5, xshift=-5pt, yshift=5pt] {$S_1 \wedge S_2 \Rightarrow T$} (T.east);
\end{tikzpicture}%
}
\caption{Intended theorem and shadows.}
\label{fig:intro-a}
\end{subfigure}
\hfill
\begin{subfigure}[t]{0.31\linewidth}
\centering
\resizebox{\linewidth}{!}{%
\begin{tikzpicture}[
  every node/.style={font=\small},
  stmt/.style={draw, rounded corners, align=center, minimum width=2.55cm,
    minimum height=0.75cm},
  shadow/.style={draw, rounded corners, align=center, minimum width=2.2cm,
    minimum height=0.68cm, fill=gray!10},
  edgeLabel/.style={font=\scriptsize, fill=white, inner sep=1.2pt},
  >=Latex, thick
]
  \node[stmt] (That) at (0, 1.45) {$\widehat T_1 := A \Rightarrow B_1$};
  \node[shadow] (S1) at (-1.25, 0) {$S_1 := A \Rightarrow B_1$};
  \node[shadow] (S2) at (1.25, 0) {$S_2 := A \Rightarrow B_2$};
  \begin{scope}[on background layer]
    \node[draw, rounded corners, dashed, inner xsep=0.18cm, inner ysep=0.14cm,
          fit=(S1)(S2)] (group) {};
  \end{scope}
  \draw[->] (That) -- node[edgeLabel, left, pos=0.45]
      {$\widehat T_1 \Rightarrow S_1$} (S1);
  \draw[->, red!70!black, dashed] (That) -- node[edgeLabel, right, pos=0.45]
      {$\widehat T_1 \nRightarrow S_2$} (S2);
  \draw[->, dotted]
      ($(group.east)+(0.03,0)$) to[out=35, in=0]
      node[edgeLabel, pos=0.5, xshift=-5pt, yshift=5pt] {$S_1 \wedge S_2 \Rightarrow \widehat T_1$} (That.east);
\end{tikzpicture}%
}
\caption{Forward check failure.}
\label{fig:intro-b}
\end{subfigure}
\hfill
\begin{subfigure}[t]{0.31\linewidth}
\centering
\resizebox{\linewidth}{!}{%
\begin{tikzpicture}[
  every node/.style={font=\small},
  stmt/.style={draw, rounded corners, align=center, minimum width=2.55cm,
    minimum height=0.75cm},
  shadow/.style={draw, rounded corners, align=center, minimum width=2.2cm,
    minimum height=0.68cm, fill=gray!10},
  edgeLabel/.style={font=\scriptsize, fill=white, inner sep=1.2pt},
  >=Latex, thick
]
  \node[stmt] (That) at (0, 1.45)
      {$\widehat T_2 := A \Rightarrow$\\$B_1 \wedge B_2 \wedge R$};
  \node[shadow] (S1) at (-1.25, 0) {$S_1 := A \Rightarrow B_1$};
  \node[shadow] (S2) at (1.25, 0) {$S_2 := A \Rightarrow B_2$};
  \begin{scope}[on background layer]
    \node[draw, rounded corners, dashed, inner xsep=0.18cm, inner ysep=0.14cm,
          fit=(S1)(S2)] (group) {};
  \end{scope}
  \draw[->] (That) -- node[edgeLabel, left, pos=0.45]
      {$\widehat T_2 \Rightarrow S_1$} (S1);
  \draw[->] (That) -- node[edgeLabel, right, pos=0.45]
      {$\widehat T_2 \Rightarrow S_2$} (S2);
  \draw[->, red!70!black, dotted]
      ($(group.east)+(0.03,0)$) to[out=35, in=0]
      node[edgeLabel, pos=0.5, xshift=-5pt, yshift=5pt]
      {$S_1 \wedge S_2 \nRightarrow \widehat T_2$} (That.east);
\end{tikzpicture}%
}
\caption{Backward check failure.}
\label{fig:intro-c}
\end{subfigure}
\caption{Illustrative example of shadow theorem checks.
Here $A,B,B_1,B_2$ are predicates and $B \Leftrightarrow B_1 \wedge B_2$.
The intended theorem statement $T$ implies both shadows ($S_1$, $S_2$), and the two
shadows jointly imply $T$.
The shadow theorem statement checks reject both generated theorems.
The forward check finds that $\widehat{T}_1$ does not imply $S_2$, and the
backward check finds that $S_1 \wedge S_2$ does not imply $\widehat{T}_2$.}
\label{fig:intro}
\end{figure*}

The core challenge for evaluation is to assess whether the generated formal
statement expresses the same mathematics as the informal
statement~\citep{chen2025minif2fv2,ammanamanchi2026faults}.
Existing automatic metrics either rely on compile rate signals that can accept
misaligned statements, assign scores without formally checking the intended
meaning, or require equivalence to a fixed reference formulation that can reject
correct statements expressed differently (\Cref{sec:rel-quality}).
An alternative is manual judgment by mathematicians familiar with Lean,
which is reliable yet too costly to scale for repeated
evaluations~\citep{wu2022autoformalization,wang2025nl2lean,wu2024leanworkbook}.


In this paper, we propose \ourmetric (\emph{Semantic Alignment Pass}), an
automatic metric that scores semantic alignment through forward and backward
implication checks.
\ourmetric follows the Pass@$k$ idea from code generation, where an output is
accepted only if it passes all tests for the task~\citep{chen2021codex}.
For formal statements, we define these checks using a complete set of
\emph{shadow theorems}: auxiliary theorems selected such that each shadow is
implied by the intended theorem and the shadows together characterize it
(\Cref{fig:intro-a}).
During evaluation, a generated formalization receives the maximum \ourmetric score
only if it compiles, its statement implies every shadow in the forward checks
(\Cref{fig:intro-b}), and its statement is implied by the shadows' conjunction
in the backward check (\Cref{fig:intro-c}). 


This design makes expert supervision reusable: a shadow set is constructed once
for each intended theorem rather than separately for every generated
formalization.
Experts construct the set by revising LLM-generated drafts
(\Cref{sec:curation}), and Lean mechanically verifies the set completeness (\Cref{sec:complete-shadow-sets}).
The same shadow checkers can then be reused to evaluate any number of generated
formalizations automatically.

Alongside \ourmetric, we introduce \ours, a Lean~4 full-autoformalization
benchmark comprising 178 postgraduate- to research-level problems across eight
mathematical areas and three difficulty levels.
Each problem is annotated with a complete set of shadow theorems, enabling
automatic evaluation with \ourmetric.
Existing benchmarks such as ProofNet~\cite{azerbayev2023proofnet} primarily
contain short formalizations with a single conclusion.
In contrast, \ours targets multi-conclusion problems whose reference
formalizations are about 1.6 times longer lines on average 
(\Cref{tab:dataset-statistics}).
\ifarxivonly
An early version of \ours, comprising 126 problems, served as the benchmark for Track~4 of the ICML 2026 AI4Math Challenge, 
with analyses of the results provided in \Cref{sec:appendix:challenge}.
\fi

Empirically, Claude Code (Opus~4.8) with
Numina-Lean-Agent~\cite{numina2026leanagent} achieves a $61.8\%$ compile rate on \ours, yet only $11.2\%$ of its outputs pass \ourmetric (\Cref{tab:main-results}).
\edit{\ourmetric achieves $98.8\%$ binary agreement with expert judgments for generated outputs across six agentic configurations, whereas compile rate and BEq+~\citep{li2025rethinking,poiroux-etal-2025-reliable} achieve $17.8\%$ and $80.6\%$, respectively (\Cref{tab:human-correlation}).}


We summarize our contributions as follows.
\begin{itemize}
    \setlength{\itemsep}{0pt}
    \item We introduce \ourmetric, an automated metric that evaluates semantic
    alignment between formal and informal statements through Lean-checkable
    shadow theorems.
    \item We construct \ours, a Lean~4 full autoformalization benchmark that
    uses \ourmetric as its main evaluation metric across {178} postgraduate- to
    research-level problems.
    Claude Code (Opus~4.8) with Numina-Lean-Agent reaches {$61.8\%$} compile rate and \edit{$11.2\%$} \ourmetric.
    \item \edit{\ourmetric achieves $98.8\%$ binary agreement with expert judgments.}
\end{itemize}

\section{Related Work}
\label{sec:rel}

\subsection{Autoformalization Tasks and Benchmarks}
\label{sec:rel-benchmarks}


Autoformalization is a task 
that translates an informal theorem into a formal theorem in a proof assistant such as Lean, 
for either the statement or the proof.
For statement autoformalization, 
various methods have been proposed~\citep{wu2022autoformalization,li2025atlas,lu2025conceptretrieval,wang2025nl2lean,li2026reform} and 
evaluated on benchmarks spanning difficulty levels from high school to undergraduate~\citep{jiang2023herald,wu2024leanworkbook,murphy2024leaneuclid,cheng2025fmc}.
{\citet{lu2024processdriven,li2026proofbridge} have studied full autoformalization, which produces both the statement and its proof from natural language.}
At the proof level, 
this task is mostly discussed as theorem proving, 
where various methods~\citep{yang2023leandojo,xin2024deepseekprover,lin2025goedelprover,wang2025kiminaprover,numina2026leanagent} generate a formal proof given the formal statement, 
evaluated on benchmarks such as miniF2F and ProofNet~\citep{zheng2022minif2f,azerbayev2023proofnet}, and on research-level Lean projects~\citep{poiroux2025rlmeval}.
While these benchmarks establish the task setting, our focus is the complementary problem of evaluating whether generated formal statements are semantically aligned with the intended theorem.

\subsection{Semantic Alignment Evaluation}
\label{sec:rel-quality}

\paragraph{Compile rate.}
Compile rate runs the generated Lean code through the Lean compiler and
counts it as correct when the generated theorem statement and proof type-check \cite{azerbayev2023proofnet}.
It rejects ill-formed or unproved Lean code, but it can produce false positives
when a generated proof type-checks for a statement that differs from the
informal theorem \cite{li2025rethinking, ammanamanchi2026faults}.

\paragraph{Reference-based metrics.}
Several metrics evaluate generated statements against a reference
formalization.
BLEU~\citep{azerbayev2023proofnet,papineni2002bleu} measures lexical overlap
with the reference theorem, TransTED~\citep{ying2026assess} compares structural
similarity to the reference theorem, and FormalAlign~\citep{liu2025formalalign}
and LLM-as-judge~\citep{guo2025autoformalizer} score the generated statement
against the informal statement.
Because these lexical, structural, and model-based scores are not implication
checks, they can produce both false positives and false negatives.
BEq and BEq+~\citep{li2025rethinking,poiroux-etal-2025-reliable} use Lean tactics to prove
equivalence with the reference statement, but they \edit{create
false negatives by rejecting correct statements whose formulations differ from
the fixed reference} (\Cref{tab:human-correlation}).

\paragraph{Expert judgments.}
Manual judgment by mathematicians familiar with Lean remains reliable, but it
does not scale to large or repeated
evaluations~\citep{wu2022autoformalization,wang2025nl2lean,wu2024leanworkbook}.

\paragraph{Test-based metrics.}
Testing Accuracy~\citep{kim2026benchmarking} treats the dependent successor theorems as test cases, 
accepting a generated theorem when all of them compile. 
This approach can produce false positives, since passing all successor theorem tests does not guarantee semantic alignment with the intended informal statement.


\paragraph{Our distinction.}

\ourmetric 
addresses both false positives and
false negatives in semantic alignment 
by evaluating generated statements 
against a complete set of auxiliary theorems.
\ourmetric verifies generated statement with a complete set of both forward and backward implication checks, \edit{obtaining $98.8\%$ binary agreement with expert judgments}
(\Cref{tab:human-correlation}).

\section{\ourmetric: Semantic Alignment Pass}
\label{sec:framework}


\subsection{Task and Shadow Theorems}
\label{sec:shadow-theorems}

\begin{definition}[Full autoformalization task]
Given an informal theorem statement and its proof, the full autoformalization
task is to generate a formal statement $\widehat T$ together with a
machine-checkable proof of $\widehat T$.
Let $T$ denote the intended formal statement.
The semantic alignment objective is
$\widehat T \Leftrightarrow T$.
\end{definition}

\begin{definition}[Shadow theorem]
For an intended formal statement $T$, a shadow theorem consists of an auxiliary
formal statement $S$ and a machine-checkable proof of $S$.
Its forward checker theorem states that $T$ implies $S$:
\[
T \Rightarrow S.
\]
\end{definition}

\subsection{Complete Shadow Sets}
\label{sec:complete-shadow-sets}

\begin{definition}[Complete shadow set]
Let $T$ be the intended formal statement, and let
$\mathcal{S}=\{S_1,\ldots,S_n\}$ be a finite set of shadow statements.
The set $\mathcal{S}$ is \emph{complete} if the shadows jointly characterize
$T$, i.e., the backward checker theorem states
\[
S_1 \wedge \cdots \wedge S_n \Rightarrow T.
\]
Together with the forward checker theorems $T \Rightarrow S_i$, this gives
$S_1 \wedge \cdots \wedge S_n \Leftrightarrow T$.
\end{definition}

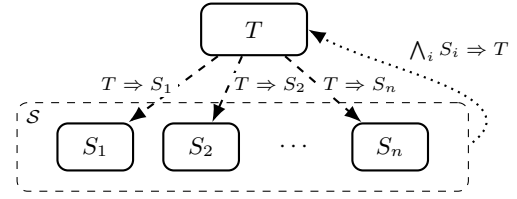
\begin{figure}[t]
\centering
\begin{tikzpicture}[
  every node/.style={font=\small},
  box/.style={draw, rounded corners, minimum width=1.4cm, minimum height=0.68cm, align=center},
  shadow/.style={draw, rounded corners, minimum width=1.0cm, minimum height=0.62cm, align=center},
  edgeLabel/.style={font=\scriptsize, fill=white, inner sep=1.5pt},
  >=Latex, thick
]
  \node[box] (T) at (0, 0) {$T$};
  \node[shadow] (S1) at (-2.1, -1.55) {$S_1$};
  \node[shadow] (S2) at (-0.7, -1.55) {$S_2$};
  \node[font=\small] (dots) at (0.55, -1.55) {$\cdots$};
  \node[shadow] (S3) at (1.8, -1.55) {$S_n$};
  \begin{scope}[on background layer]
    \node[draw, rounded corners, dashed, inner xsep=0.52cm, inner ysep=0.26cm,
          fit=(S1)(S2)(dots)(S3)] (group) {};
  \end{scope}
  \node[font=\scriptsize, fill=white, inner sep=1pt, anchor=north west]
      at ($(group.north west)+(0.08,-0.06)$) {$\mathcal{S}$};
  \draw[->, dashed] (T) -- node[edgeLabel, left, pos=0.44] {$T \Rightarrow S_1$} (S1);
  \draw[->, dashed] (T) -- node[edgeLabel, right, pos=0.43] {$T \Rightarrow S_2$} (S2);
  \draw[->, dashed] (T) -- node[edgeLabel, right, pos=0.43] {$T \Rightarrow S_n$} (S3);
  \draw[->, dotted, thick]
      ($(group.east)+(0.05,0.06)$) to[out=38, in=330]
      node[edgeLabel, right, pos=0.68, xshift=5pt, yshift=5pt] {$\bigwedge_i S_i \Rightarrow T$} (T.east);
\end{tikzpicture}
\caption{A complete shadow set.
Each shadow statement $S_i$ follows from the intended formal statement $T$.
The set is complete when the shadows jointly imply $T$.}
\label{fig:complete-shadow-set}
\end{figure}

The completeness of a shadow set is verified in Lean by compiling checker
theorems in both directions. 
For each shadow $S_i$, we compile a forward checker theorem establishing $T \Rightarrow S_i.$
A candidate shadow is accepted only if its forward checker compiles.
We then compile the backward checker theorem $S_1 \wedge \cdots \wedge S_n \Rightarrow T.$
The backward checker theorem compiles only when the shadow statements jointly
imply $T$, which is precisely the completeness condition illustrated in
\Cref{fig:complete-shadow-set}.

The annotation process of complete shadow sets and corresponding checker theorems can be mostly automated (\Cref{sec:curation}). 

The following example shows common ways to build complete shadow sets.
Additional checker patterns appear in \Cref{app:complete-shadow-checkers}.

\begin{figure*}[t]
{
\centering
    \includegraphics[width=\linewidth]{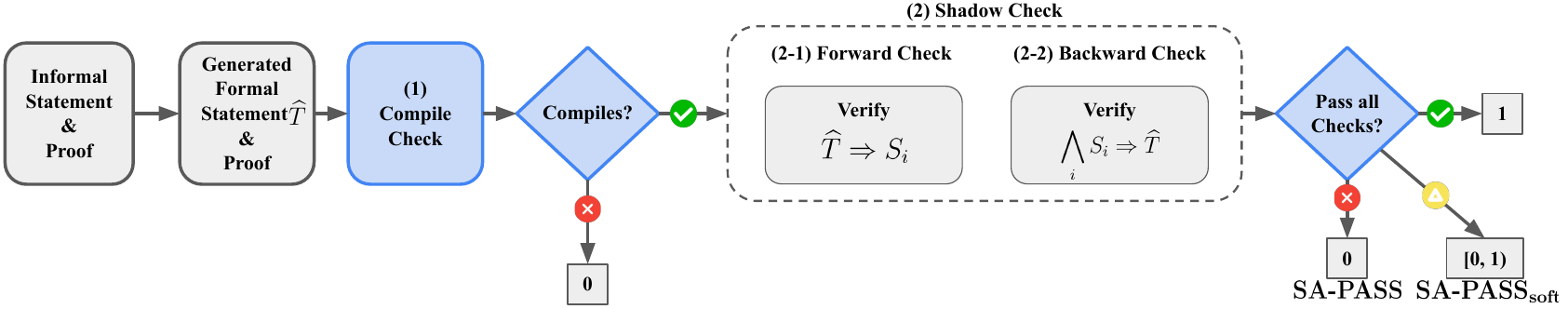}
    \caption{
    Evaluation procedure for \ourmetric and \ourrelaxed on one \ours problem.
    }
    \label{fig:pipeline}
}
\end{figure*}

\begin{example}[Bundled conclusions]
\label{ex:bundled-conclusions}
Suppose $A,B_1,\ldots,B_n$ are predicates and the intended formal statement has
the form
\[
A \Rightarrow B_1 \wedge \cdots \wedge B_n.
\]
For each $i$, define a shadow theorem
\[
S_i : A \Rightarrow B_i.
\]
Then $\mathcal{S}$ is complete, because the individual components can be
assembled back into the bundled conclusion.

\Cref{app:complete-shadow-bundled,app:complete-shadow-equality} show the corresponding Lean checkers.

\end{example}




\subsection{\ourmetric Scoring}
\label{sec:sa-pass-condition}

\begin{definition}[Passing condition]
\label{def:passing-condition}
For a problem whose intended formal statement is $T$, a generated formal theorem
with statement $\widehat T$ passes if and only if its declaration and proof type-check
in Lean and it passes the forward checks
\[
\widehat T \Rightarrow S_1,\quad
\widehat T \Rightarrow S_2,\quad \ldots,\quad
\widehat T \Rightarrow S_n
\]
and the backward check
\[
S_1 \wedge \cdots \wedge S_n \Rightarrow \widehat T,
\]
where $\mathcal{S}=\{S_1,\ldots,S_n\}$ is a complete shadow set for $T$.
\end{definition}

During evaluation, we append the forward and backward checkers to the generated Lean
code, and compile the resulting file.

Because $\mathcal{S}$ is complete, passing all forward checks establishes
$\widehat T \Rightarrow S_1 \wedge \cdots \wedge S_n \Rightarrow T$.
Passing the backward check certifies
$T \Rightarrow S_1 \wedge \cdots \wedge S_n \Rightarrow \widehat T$,
and it rejects generated formal statements that add constraints not required by the
intended formal statement.
For example, if the generated formal statement has the form
$\widehat T = T \wedge R$ for an arbitrary extra requirement $R$, then
$\widehat T \Rightarrow T$ but $T \nRightarrow \widehat T$.
Together, the forward and backward checks establish
$\widehat T \Leftrightarrow T$.

The passing condition in \Cref{def:passing-condition} induces the following
binary semantic-alignment metric.

\begin{definition}[\ourmetric]
\label{def:sa-pass-strict}
For a generated formal theorem with statement $\widehat T$ and a complete shadow
set $\mathcal{S}=\{S_1,\ldots,S_n\}$, define
\[
\text{\ourmetric}(\widehat T,\mathcal{S}) =
\begin{cases}
1,
& \begin{array}{l}
\text{if } \widehat T \text{ compiles, } \\
\widehat T \Rightarrow S_i \text{ for all } i, \\
\text{and } \bigwedge_{i=1}^n S_i \Rightarrow \widehat T;
\end{array}\\
0, & \text{otherwise.}
\end{cases}
\]
\end{definition}

Since $\mathcal{S}$ is complete for the intended formal statement $T$,
$\text{\ourmetric}(\widehat T,\mathcal{S}){=}1$ certifies
$\widehat T \Leftrightarrow T$.

\begin{definition}[\ourrelaxed]
\label{def:sa-pass}
Let $\widehat T$ be the statement of a generated formal theorem, and let
$\mathcal{S}=\{S_1,\ldots,S_n\}$ be a complete shadow set.
If the generated theorem does not type-check in Lean, set
$\text{\ourrelaxed}(\widehat T,\mathcal{S}){=}0$.
Otherwise, its \ourrelaxed score is
\begin{multline}
\text{\ourrelaxed}(\widehat T,\mathcal{S}) := 
\frac{1}{2}\cdot
\underbracket[0.5pt][2pt]{
\frac{|\{i : \widehat T \Rightarrow S_i\}|}{|\mathcal{S}|}}_{\text{forward checks}} \\
+
\frac{1}{2}\cdot
\underbracket[0.5pt][2pt]{
\mathbf{1}\{\bigwedge_{i=1}^n S_i \Rightarrow \widehat T\}}_{\text{backward check}}. \nonumber
\end{multline}
Here $\mathbf{1}\{\phi\}$ equals $1$ if the corresponding checker theorem
type-checks in Lean, and $0$ otherwise.
\end{definition}

A generated statement may become weaker than the intended theorem by adding
assumptions, or stronger by omitting them.
\ourrelaxed assigns partial credit in such cases by averaging the fraction of
successful forward checks and the backward check indicator.
By construction,
$\text{\ourrelaxed}(\widehat T,\mathcal{S})=1 
\Leftrightarrow
\text{\ourmetric}(\widehat T,\mathcal{S})=1.$

\Cref{fig:pipeline} summarizes the evaluation procedure.

\section{\ours}
To evaluate full autoformalization systems with \ourmetric, we construct \ours,
a Lean~4 benchmark equipped with checker theorems for semantic alignment
evaluation.
Its checker construction is LLM-assisted and compiler-verified, while the
resulting benchmark offers broad mathematical coverage and remains challenging
for current systems.




\paragraph{LLM-assisted checker construction.}
Qwen3-235B proposes candidate shadow sets and their checker theorems, Lean
verifies shadow set completeness, and experts review the results and guide
revisions when needed.
The checker construction takes about $10$ minutes per problem on average, where for $85\%$ of problems the initial LLM-generated shadow set is already complete (\Cref{sec:curation}).

\paragraph{Coverage.}
\ours contains \edit{178} problems ranging from postgraduate- to research-level across
eight mathematical areas and three difficulty levels
(\Cref{tab:benchmark-coverage}).
L1 and L2 contain postgraduate-level problems, separated by reference Lean
length as a proxy for formalization complexity, with L2 containing longer
formalizations.
L3 contains research-level problems.


\begin{table}[t]
\centering
\small
\resizebox{\columnwidth}{!}{%
\begin{tabular}{lcccc}
\toprule
Area & L1 & L2 & L3 & Total \\
\midrule
\texttt{geometry} (Geo)                 & 18 &  6 &  0 &  24 \\
\texttt{topology} (Top)                 & \edit{16} &  9 &  0 &  \edit{25} \\
\texttt{algebra} (Alg)                  & 19 &  4 &  3 &  26 \\
\texttt{analysis} (Anl)                 & 14 & 10 &  4 &  28 \\
\texttt{algebraic-geometry} (AG)        &  2 & 16 &  4 &  22 \\
\texttt{combinatorics} (Cmb)            &  8 &  6 &  0 &  14 \\
\texttt{number-theory} (NT)             & 17 &  0 &  2 &  19 \\
\texttt{probability} (Prob)             & 19 &  1 &  0 &  20 \\
\midrule
Total                                   & \edit{113} & 52 & 13 & \edit{178} \\
\bottomrule
\end{tabular}}
\caption{Number of problems per area and difficulty level in \ours. One topology problem was withdrawn after an error was found in its reference statement.}
\label{tab:benchmark-coverage}
\end{table}

\paragraph{Difficulty.}
\ours remains challenging for current systems.
\edit{Claude Code (Opus~4.8) with Numina-Lean-Agent, the strongest system we
evaluate under \ourmetric, achieves a $61.8\%$ compile rate but scores only $18.3\%$ on
\ourrelaxed and $11.2\%$ on \ourmetric (\Cref{tab:main-results}).}

\subsection{Benchmark Construction}
\label{sec:curation}

The curation interface for entering informal statements, reference Lean code,
metadata, and checker blocks is shown in Appendix (\Cref{fig:annotation-main,fig:annotation-example,fig:annotation-write}).
Five mathematicians, spanning graduate students, postdoctoral researchers, and
faculty, performed semantic review and guided revisions throughout the
construction process.

\paragraph{Informal theorem.}
Annotators first collect source problems from textbooks, lecture notes, and
research repositories, summarized in \Cref{tab:repository_detail}.
When only a Lean formalization is available, Qwen3-235B drafts the informal
theorem.
We exclude problems whose solutions are dominated by theorem retrieval, since
such cases test library search more than full autoformalization.

\paragraph{Reference formalization.}

For each informal theorem, we construct a reference Lean formalization, with
Qwen3-235B providing a draft when only the informal theorem is available.
Annotators edit the code until the theorem type-checks in Lean, verify that its
formal statement matches the informal theorem, and use
LeanSearch\footnote{\url{https://leansearch.net/}} to correct minor errors such
as hallucinated theorem names.
 

\paragraph{Checker theorems.}
\label{sec:shadow-construction}

Given the reference formalization, Qwen3-235B proposes a candidate shadow set
and its checker theorems (\Cref{fig:prompt-forward-checker}), and Lean verifies
the set's completeness by compiling all forward and backward checkers
(\Cref{sec:complete-shadow-sets}).
The initial draft is complete for $85\%$ of problems. For the remaining $15\%$,
an expert guides the model to revise or add shadows until all checkers compile.
For $77.1\%$ of checker proofs, \texttt{exact} alone suffices because the goal
follows directly from existing library facts or the statement structure.
Including LLM drafting, checker construction averages about $10$ minutes per
problem.
\Cref{tab:hidden-checker-statistics} summarizes the number of forward and backward checker theorems per problem.

While each checker is constructed using the reference formalization, valid solver outputs may use different binder styles, declaration names, or field notation. 
To reduce false negatives, we diversify the checkers during benchmark construction to cover common differences in binder style and field notation. 
At evaluation time, rule-based rewriting first addresses any remaining declaration-name mismatches. 
If rule-based rewriting fails, Qwen3-235B attempts to generate an adapter proof that establishes equivalence between the solver declaration and the corresponding reference declaration, allowing the checker to run in Lean.
\begin{table}[t]
\centering
\small
\setlength{\tabcolsep}{4pt}
\resizebox{0.7\columnwidth}{!}{%
\begin{tabular}{lrrrr}
\toprule
Type & Count & Mean & Min & Max \\
\midrule
Forward  & \edit{513} & \edit{2.88} & 1 & 13 \\
Backward & \edit{193} & \edit{1.08} & 1 & 12 \\
Total    & \edit{706} & \edit{3.97} & 1 & 25 \\
\bottomrule
\end{tabular}}
\caption{
Statistics of forward and backward checker theorems in \ours, counted over its 178 problems.
}
\label{tab:hidden-checker-statistics}
\end{table}

\subsection{Benchmark Release}
\label{sec:release}

We provide the informal theorems and their formalization rules through a hosted
portal such as CodaBench,\footnote{\url{https://www.codabench.org/}} while keeping the shadow and checker
theorems hidden to prevent direct optimization for the checks.
A version-pinned evaluator runs the hidden checkers and reports the \ourmetric
and \ourrelaxed scores.

\section{Experiment}
\label{sec:exp}

\subsection{Experimental Setup}
\label{sec:experimental-setup}

\paragraph{Evaluation metrics.}
\label{par:eval-metrics}\label{par:sa-setup}

The main results in \Cref{tab:main-results} use three metrics.
\textbf{Compile rate} \cite{azerbayev2023proofnet} is the fraction of outputs whose final \texttt{theorem}
block type-checks in the pinned Lean~4 and Mathlib
environment.\footnote{v4.29.0}
\textbf{\ourmetric} (\Cref{def:sa-pass-strict}) is $1$ if the generated theorem
type-checks and all forward and backward checks pass, and $0$ otherwise.
\textbf{\ourrelaxed} (\Cref{def:sa-pass}) averages the forward-check pass rate
and the backward-check indicator.

To analyze agreement with expert judgments
(\Cref{tab:human-correlation,tab:proofnet-agreement}), we include three
additional metrics.
\textbf{BLEU}~\citep{azerbayev2023proofnet,papineni2002bleu} measures lexical
$n$-gram overlap between the generated theorem block and the reference Lean
theorem.
\textbf{BEq+}~\citep{poiroux-etal-2025-reliable} uses Lean tactics to prove both directions
of implication between the generated and reference statements.
\textbf{LLM-as-judge}~\citep{guo2025autoformalizer} uses
Gemini Pro 3.1, Claude Opus 4.7, and GPT-5.4.
We queried each judge three times and took the majority over all nine votes, where the total API cost was approximately \$480.

\begin{table*}[t]
  \centering
  \scriptsize
  \setlength{\tabcolsep}{3pt}
  \providecommand{\ShadowAvgsep}{\rule[-0.65ex]{0.45pt}{2.7ex}}
  \resizebox{\textwidth}{!}{%
  \begin{tabular}{lccccccccc@{\hspace{3pt}}c@{\hspace{3pt}}ccc}
    \toprule
    & \multicolumn{3}{c}{L1} & \multicolumn{3}{c}{L2} & \multicolumn{3}{c}{L3} & \ShadowAvgsep & \multicolumn{3}{c}{Average} \\
    & \multicolumn{3}{c}{(n{=}\edit{113})} & \multicolumn{3}{c}{(n{=}52)} & \multicolumn{3}{c}{(n{=}13)} & \ShadowAvgsep & \multicolumn{3}{c}{(n{=}\edit{178})} \\
    \cmidrule(lr){2-4}\cmidrule(lr){5-7}\cmidrule(lr){8-10}\cmidrule(lr){12-14}
    Method & Compile & \ourrelaxed & \ourmetric & Compile & \ourrelaxed & \ourmetric & Compile & \ourrelaxed & \ourmetric & \ShadowAvgsep & Compile & \ourrelaxed & \ourmetric \\
    \midrule
    \multicolumn{14}{c}{\textit{\textbf{Agentic methods}}} \\
    \midrule
    Claude Code (Opus 4.6) & \edit{\phantom{0}6.2} & \phantom{0}1.8 & \phantom{0}0.0 & \phantom{0}1.9 & \phantom{0}1.0 & \phantom{0}0.0 & \phantom{0}0.0 & \phantom{0}0.0 & \phantom{0}0.0 & \ShadowAvgsep & \phantom{0}4.5 & \phantom{0}1.4 & \phantom{0}0.0 \\
    \quad + Numina & \edit{55.8} & \edit{\textbf{16.8}} & \edit{\textbf{\phantom{0}8.8}} & 36.5 & 17.3 & 11.5 & 15.4 & \phantom{0}0.0 & \phantom{0}0.0 & \ShadowAvgsep & \edit{47.2} & \edit{15.7} & \edit{\phantom{0}9.0} \\
    Claude Code (Opus 4.8) & \edit{24.8} & \edit{\phantom{0}4.9} & \edit{\phantom{0}2.7} & 13.5 & \phantom{0}6.7 & \phantom{0}3.8 & \phantom{0}7.7 & \phantom{0}0.0 & \phantom{0}0.0 & \ShadowAvgsep & \edit{20.2} & \edit{\phantom{0}5.1} & \phantom{0}2.8 \\
    \quad + Numina & \edit{61.1} & \edit{13.3} & \edit{\phantom{0}6.2} & \textbf{71.2} & \textbf{31.7} & \textbf{23.1} & \textbf{30.8} & \textbf{\phantom{0}7.7} & \textbf{\phantom{0}7.7} & \ShadowAvgsep & \edit{\textbf{61.8}} & \edit{\textbf{18.3}} & \edit{\textbf{11.2}} \\
    Claude Code (Qwen3 235B) & \phantom{0}4.4 & \phantom{0}1.3 & \phantom{0}0.9 & \phantom{0}5.8 & \phantom{0}0.0 & \phantom{0}0.0 & \phantom{0}7.7 & \phantom{0}0.0 & \phantom{0}0.0 & \ShadowAvgsep & \edit{\phantom{0}5.1} & \phantom{0}0.8 & \phantom{0}0.6 \\
    \quad + Numina & \phantom{0}8.8 & \phantom{0}1.3 & \phantom{0}0.9 & \phantom{0}3.8 & \phantom{0}0.0 & \phantom{0}0.0 & \phantom{0}0.0 & \phantom{0}0.0 & \phantom{0}0.0 & \ShadowAvgsep & \phantom{0}6.7 & \phantom{0}0.8 & \phantom{0}0.6 \\
    Codex (GPT-5.4) & \edit{20.4} & \phantom{0}4.4 & \edit{\phantom{0}2.7} & 11.5 & \phantom{0}1.9 & \phantom{0}1.9 & \phantom{0}7.7 & \phantom{0}0.0 & \phantom{0}0.0 & \ShadowAvgsep & \edit{16.9} & \phantom{0}3.4 & \phantom{0}2.2 \\
    \quad + Numina & \edit{\textbf{61.9}} & \edit{15.9} & \edit{\textbf{\phantom{0}8.8}} & 48.1 & 21.2 & 13.5 & \textbf{30.8} & \textbf{\phantom{0}7.7} & \textbf{\phantom{0}7.7} & \ShadowAvgsep & \edit{55.6} & \edit{16.9} & \edit{10.1} \\
    \midrule
    \multicolumn{14}{c}{\textit{\textbf{Closed-source LLMs}}} \\
    \midrule
    Claude Opus 4.6 & \phantom{0}1.8 & \phantom{0}0.0 & \phantom{0}0.0 & \phantom{0}0.0 & \phantom{0}0.0 & \phantom{0}0.0 & \phantom{0}0.0 & \phantom{0}0.0 & \phantom{0}0.0 & \ShadowAvgsep & \phantom{0}1.1 & \phantom{0}0.0 & \phantom{0}0.0 \\
    Claude Sonnet 4.6 & \phantom{0}0.9 & \phantom{0}0.0 & \phantom{0}0.0 & \phantom{0}0.0 & \phantom{0}0.0 & \phantom{0}0.0 & \phantom{0}0.0 & \phantom{0}0.0 & \phantom{0}0.0 & \ShadowAvgsep & \phantom{0}0.6 & \phantom{0}0.0 & \phantom{0}0.0 \\
    Claude Haiku 4.5 & \phantom{0}0.9 & \phantom{0}0.4 & \phantom{0}0.0 & \phantom{0}0.0 & \phantom{0}0.0 & \phantom{0}0.0 & \phantom{0}0.0 & \phantom{0}0.0 & \phantom{0}0.0 & \ShadowAvgsep & \phantom{0}0.6 & \phantom{0}0.3 & \phantom{0}0.0 \\
    GPT-5.4 & \phantom{0}1.8 & \phantom{0}0.4 & \phantom{0}0.0 & \phantom{0}5.8 & \phantom{0}0.0 & \phantom{0}0.0 & \phantom{0}7.7 & \phantom{0}0.0 & \phantom{0}0.0 & \ShadowAvgsep & \phantom{0}3.4 & \phantom{0}0.3 & \phantom{0}0.0 \\
    GPT-5.4 mini & \phantom{0}8.8 & \phantom{0}0.4 & \phantom{0}0.0 & \phantom{0}3.8 & \phantom{0}0.0 & \phantom{0}0.0 & \textbf{30.8} & \phantom{0}0.0 & \phantom{0}0.0 & \ShadowAvgsep & \phantom{0}8.9 & \phantom{0}0.3 & \phantom{0}0.0 \\
    GPT-5.4 nano & \phantom{0}6.1 & \phantom{0}0.0 & \phantom{0}0.0 & \phantom{0}5.8 & \phantom{0}0.0 & \phantom{0}0.0 & 15.4 & \phantom{0}0.0 & \phantom{0}0.0 & \ShadowAvgsep & \phantom{0}6.7 & \phantom{0}0.0 & \phantom{0}0.0 \\
    Gemini 3.1 Pro & \phantom{0}0.9 & \phantom{0}0.0 & \phantom{0}0.0 & \phantom{0}0.0 & \phantom{0}0.0 & \phantom{0}0.0 & 15.4 & \phantom{0}0.0 & \phantom{0}0.0 & \ShadowAvgsep & \phantom{0}1.7 & \phantom{0}0.0 & \phantom{0}0.0 \\
    Gemini 2.5 Flash & \phantom{0}0.9 & \phantom{0}0.4 & \phantom{0}0.0 & \phantom{0}0.0 & \phantom{0}0.0 & \phantom{0}0.0 & \phantom{0}0.0 & \phantom{0}0.0 & \phantom{0}0.0 & \ShadowAvgsep & \phantom{0}0.6 & \phantom{0}0.3 & \phantom{0}0.0 \\
    \midrule
    \multicolumn{14}{c}{\textit{\textbf{Open-source LLMs}}} \\
    \midrule
    GPT-OSS-120B & \phantom{0}0.9 & \phantom{0}0.4 & \phantom{0}0.0 & \phantom{0}0.0 & \phantom{0}0.0 & \phantom{0}0.0 & \phantom{0}0.0 & \phantom{0}0.0 & \phantom{0}0.0 & \ShadowAvgsep & \phantom{0}0.6 & \phantom{0}0.3 & \phantom{0}0.0 \\
    GPT-OSS-20B & \phantom{0}4.4 & \phantom{0}0.4 & \phantom{0}0.0 & \phantom{0}1.9 & \phantom{0}0.0 & \phantom{0}0.0 & \phantom{0}0.0 & \phantom{0}0.0 & \phantom{0}0.0 & \ShadowAvgsep & \phantom{0}3.4 & \phantom{0}0.3 & \phantom{0}0.0 \\
    DeepSeek R1 0528 & \phantom{0}0.9 & \phantom{0}0.0 & \phantom{0}0.0 & \phantom{0}0.0 & \phantom{0}0.0 & \phantom{0}0.0 & \phantom{0}0.0 & \phantom{0}0.0 & \phantom{0}0.0 & \ShadowAvgsep & \phantom{0}0.6 & \phantom{0}0.0 & \phantom{0}0.0 \\
    DeepSeek V3.2 & \phantom{0}0.0 & \phantom{0}0.0 & \phantom{0}0.0 & \phantom{0}0.0 & \phantom{0}0.0 & \phantom{0}0.0 & \phantom{0}0.0 & \phantom{0}0.0 & \phantom{0}0.0 & \ShadowAvgsep & \phantom{0}0.0 & \phantom{0}0.0 & \phantom{0}0.0 \\
    Qwen3 235B & \phantom{0}0.0 & \phantom{0}0.0 & \phantom{0}0.0 & \phantom{0}0.0 & \phantom{0}0.0 & \phantom{0}0.0 & \phantom{0}0.0 & \phantom{0}0.0 & \phantom{0}0.0 & \ShadowAvgsep & \phantom{0}0.0 & \phantom{0}0.0 & \phantom{0}0.0 \\
    Qwen3-coder & \phantom{0}0.0 & \phantom{0}0.0 & \phantom{0}0.0 & \phantom{0}0.0 & \phantom{0}0.0 & \phantom{0}0.0 & \phantom{0}0.0 & \phantom{0}0.0 & \phantom{0}0.0 & \ShadowAvgsep & \phantom{0}0.0 & \phantom{0}0.0 & \phantom{0}0.0 \\
    Llama3.1 70b & \phantom{0}0.0 & \phantom{0}0.0 & \phantom{0}0.0 & \phantom{0}0.0 & \phantom{0}0.0 & \phantom{0}0.0 & \phantom{0}0.0 & \phantom{0}0.0 & \phantom{0}0.0 & \ShadowAvgsep & \phantom{0}0.0 & \phantom{0}0.0 & \phantom{0}0.0 \\
    Llama3.1 8b & \phantom{0}0.0 & \phantom{0}0.0 & \phantom{0}0.0 & \phantom{0}0.0 & \phantom{0}0.0 & \phantom{0}0.0 & \phantom{0}0.0 & \phantom{0}0.0 & \phantom{0}0.0 & \ShadowAvgsep & \phantom{0}0.0 & \phantom{0}0.0 & \phantom{0}0.0 \\
    \midrule
    \multicolumn{14}{c}{\textit{\textbf{Lean-specialized methods}}} \\
    \midrule
    Kimina 7B$\to$8B & \phantom{0}0.0 & \phantom{0}0.0 & \phantom{0}0.0 & \phantom{0}1.9 & \phantom{0}0.0 & \phantom{0}0.0 & \phantom{0}0.0 & \phantom{0}0.0 & \phantom{0}0.0 & \ShadowAvgsep & \phantom{0}0.5 & \phantom{0}0.0 & \phantom{0}0.0 \\
    Goedel 8B$\to$32B & \phantom{0}0.0 & \phantom{0}0.0 & \phantom{0}0.0 & \phantom{0}0.0 & \phantom{0}0.0 & \phantom{0}0.0 & \phantom{0}0.0 & \phantom{0}0.0 & \phantom{0}0.0 & \ShadowAvgsep & \phantom{0}0.0 & \phantom{0}0.0 & \phantom{0}0.0 \\
    Goedel 8B$\to$8B & \phantom{0}0.0 & \phantom{0}0.0 & \phantom{0}0.0 & \phantom{0}0.0 & \phantom{0}0.0 & \phantom{0}0.0 & \phantom{0}0.0 & \phantom{0}0.0 & \phantom{0}0.0 & \ShadowAvgsep & \phantom{0}0.0 & \phantom{0}0.0 & \phantom{0}0.0 \\
    \bottomrule
  \end{tabular}}
  \caption{
    Per-difficulty pass rates (\%) on \ours.
    \emph{Compile} is the standalone Lean compile rate.
    \ourrelaxed measures the fraction of forward and backward checkers that pass.
    \ourmetric requires all forward and backward checkers to pass.
  }
  \label{tab:main-results}
\end{table*}






\paragraph{Methods.}
\label{par:models}\label{par:agentic}

The main experimental results in \Cref{tab:main-results} cover four method
groups.
For \textbf{agentic methods}, we run Claude Code with Claude Opus
4.8~\citep{claude48opus}, Claude Opus 4.6~\citep{claude46opus}, or
Qwen3-235B~\citep{yang2025qwen3}, and Codex with GPT-5.4~\citep{gpt54}.
Each configuration is evaluated with and without
Numina-Lean-Agent~\citep{numina2026leanagent}, which provides search and
compilation tools during proof construction (\Cref{tab:agentic-configs}).
For \textbf{closed-source LLMs}, we use Claude Haiku 4.5, Claude Sonnet 4.6, and
Claude Opus 4.6~\citep{claude45haiku,claude46sonnet,claude46opus}, Gemini Flash
2.5 and Gemini Pro 3.1~\citep{gemini25flash,gemini31pro}, and GPT-5.4 nano,
GPT-5.4 mini, and GPT-5.4~\citep{gpt54,gpt54mininano}.
For \textbf{open-source LLMs}, we run GPT-OSS-20B and
GPT-OSS-120B~\citep{gptoss120b}, DeepSeek V3.2 and DeepSeek R1
0528~\citep{deepseekv32,deepseekr1}, qwen3-coder and Qwen3-235B~\citep{yang2025qwen3}, and
Llama 3.1 8B and 70B~\citep{grattafiori2024llama}.
For the \textbf{Lean-specialized methods}, we use two-stage formalizer-to-prover
configurations: Kimina 7B$\to$8B~\citep{wang2025kiminaprover,kiminaprover2025full}, Goedel
8B$\to$8B, and Goedel 8B$\to$32B~\citep{goedelproverv2}.
Further model and configuration details are provided in
\Cref{sec:appendix:model_details}.

\paragraph{Method Configurations.}
\label{par:settings}
Agentic methods use the default configuration of each agent, with prompts shown
in \Cref{app:prompt-claude-code,app:prompt-agentic}.
Closed-source and open-source LLMs use zero-shot prompting with
\texttt{max\_tokens}${=}8192$, temperature $0.6$, and \texttt{top\_p}${=}0.95$.
Lean-specialized methods use \texttt{max\_tokens}${=}2048$ for the formalizer
and \texttt{max\_tokens}${=}16384$ for the prover, both with temperature $0.6$
and \texttt{top\_p}${=}0.95$.


\subsection{Results on \ours}
\label{sec:overview-results}
\Cref{tab:main-results} reports compile rate, \ourrelaxed, and \ourmetric
across all evaluated methods and difficulty levels.

\paragraph{Compilation substantially overestimates semantic alignment on \ours.}
\edit{The strongest system under \ourmetric, Claude Code (Opus~4.8) with
Numina-Lean-Agent, reaches a $61.8\%$ compile rate but only $18.3\%$
\ourrelaxed and $11.2\%$ \ourmetric.
Thus, even the relaxed score remains $43.5$ points below compile rate, while the gap to full semantic alignment is $50.6$ points.}
This discrepancy primarily reflects false positives from compile rate, 
\edit{which achieves a precision of only 0.178 in our expert analysis (\Cref{sec:metric-correlation}), compared with 1.000 for \ourmetric.}

\paragraph{Numina-Lean-Agent improves compilability more than semantic alignment.}
Employing a concurrent agentic method Numina-Lean-Agent raises the compile rate of Claude Code (Opus~4.8)
from \edit{$20.2\%$} to \edit{$61.8\%$}, a gain of \edit{$41.6$} points.
Over the same comparison, \ourrelaxed increases from \edit{$5.1\%$} to \edit{$18.3\%$}
and \ourmetric from \edit{$2.8\%$} to \edit{$11.2\%$}, gains of \edit{$13.2$} and \edit{$8.4$} points,
respectively.
For Codex, the corresponding gains are \edit{$38.7$} points in compile rate,
\edit{$13.5$} points in \ourrelaxed, and \edit{$7.9$} points in \ourmetric.
The same qualitative pattern holds for the other agentic configurations.
These results indicate that search and compiler feedback are substantially
more effective at helping agents produce type-correct Lean theorem--proof
pairs than at ensuring that the generated statements express the intended
mathematics.


\begin{table}[t]
  \centering
  \footnotesize
  \setlength{\tabcolsep}{4pt}
  \begin{tabular}{lcccc}
    \toprule
    Metric & Precision & Recall & F1 & \edit{Agreement} \\
    \midrule
    Compile      & \edit{0.178} & \edit{\textbf{1.000}} & \edit{0.302} & \edit{0.178} \\
    BLEU         & \edit{0.234} & \edit{0.674} & \edit{0.347} & \edit{0.550} \\
    LLM-as-judge & \edit{0.235} & \edit{0.093} & \edit{0.133} & \edit{0.785} \\
    BEq+         & \edit{0.400} & \edit{0.186} & \edit{0.254} & \edit{\underline{0.806}} \\
    \ourrelaxed  & \edit{\underline{0.414}} & \edit{\underline{0.953}} & \edit{\underline{0.577}} & \edit{0.752} \\
    \ourmetric   & \edit{\textbf{1.000}} & \edit{0.930} & \edit{\textbf{0.964}} & \edit{\textbf{0.988}} \\
    \bottomrule
  \end{tabular}
  \caption{\edit{Binary agreement of automatic metrics with expert judgment, averaged measurement over all six agentic configurations on \ours.
  Per-configuration results are in \Cref{sec:appendix:agreement}.}
  The best value per column is bold, and the second best is underlined.}
  \label{tab:human-correlation}
\end{table}

\paragraph{Semantic alignment on \ours remains largely unsolved.}
Closed-source LLMs, open-source LLMs, and Lean-specialized methods all score
$0.0\%$ on \ourmetric, and none exceeds $0.3\%$ on \ourrelaxed.
Even among agentic methods, the best scores are only \edit{$11.2\%$} on \ourmetric
and \edit{$18.3\%$} on \ourrelaxed.
Performance is particularly limited on L3, \edit{where the best \ourmetric scores $7.7\%$}.
For comparison, \Cref{sec:proofnet-prelim} finds that compile rate and
\ourmetric differ by only \edit{$2.1$} points on average and at most \edit{$7.8$} points
across the 16 non-agentic models evaluated on ProofNet.
This contrast suggests that the longer and more compositional targets in
\ours expose semantic failures that are less visible on shorter,
mostly single-conclusion formalizations.

\section{Analysis}
\label{sec:analysis}

\subsection{Agreement with Expert Judgment}
\label{sec:metric-correlation}

The large gap between compile rate and \ourmetric in \Cref{tab:main-results} admits two possible explanations: compilation may accept semantically misaligned formalizations, or \ourmetric may reject valid alternative formulations. We distinguish these explanations by comparing each automatic metric with expert judgments.

We collect expert judgments for compile-passed generations on \ours, where compilation and semantic alignment can disagree. Two mathematics experts independently assess whether each generated theorem is semantically aligned with the source problem and resolve disagreements through discussion. 
We analyze six configurations with sufficient compile-passed outputs: Claude Code (Opus~4.6), Claude Code (Qwen3-235B), and Codex (GPT-5.4), each with and without Numina-Lean-Agent. 
Against the resulting binary judgments, we measure precision, recall, F1, and agreement for compile rate, BLEU, BEq+, LLM-as-judge, \ourrelaxed, and \ourmetric. For the non-binary metrics, we use thresholds of $0$ for \ourrelaxed and \edit{$0.1$} for BLEU, as determined in \Cref{sec:appendix:threshold}.
Agreement measures the proportion of an automatic metric's binary decisions that match expert judgments.

\Cref{tab:human-correlation} shows that 
\ourmetric achieves the highest F1 ($0.964$) and agreement ($0.988$).
Compile rate treats compilation success as semantically aligned, yielding
$1.000$ recall but only $0.178$ precision due to false positives.
BEq+ applies Lean's built-in equivalence check, achieving higher precision
than Compile rate ($0.400$ vs.\ $0.178$) but substantially lower recall
($0.186$).
Its precision remains limited because the check covers only the final theorem
statement, ignoring the auxiliary declarations required for a complete
formalization of the informal theorem.
In contrast, \ourmetric achieves \(1.000\) precision and \(0.930\) recall. Its few false negatives stem from namespace mismatches that the LLM fallback fails to resolve (\Cref{sec:appendix:false-negatives}).

Taken together, these results indicate that the large gap between Compile rate
and \ourmetric in \Cref{tab:main-results} primarily reflects false positives
from compilation rather than false negatives from \ourmetric.

\begin{table}[t]
  \centering
  \footnotesize
  \setlength{\tabcolsep}{5pt}
  \begin{tabular}{lccc}
    \toprule
    Model & Compile & \ourrelaxed & \ourmetric \\
    \midrule
    \multicolumn{4}{c}{\textit{\textbf{Closed-source LLMs}}} \\
    \midrule
    Claude Opus 4.6   & \textbf{96.5} & 92.9 & 90.6 \\
    Claude Sonnet 4.6 & 85.9 & 85.1 & 85.1 \\
    Claude Haiku 4.5  & \textbf{96.5} & \textbf{95.3} & \textbf{95.3} \\
    GPT 5.4           & 94.9 & 93.7 & 93.7 \\
    GPT 5.4 Mini      & 95.3 & 94.1 & 94.1 \\
    GPT 5.4 Nano      & 77.6 & 76.5 & 76.5 \\
    Gemini 3.1 Pro    & 35.7 & 34.9 & 34.9 \\
    Gemini 2.5 Flash  & 82.4 & 76.1 & 76.1 \\
    \midrule
    \multicolumn{4}{c}{\textit{\textbf{Open-source LLMs}}} \\
    \midrule
    GPT-OSS-120B & 60.8 & 60.8 & 60.8 \\
    GPT-OSS-20B & 60.4 & 60.4 & 60.4 \\
    DeepSeek R1 0528 & 25.5 & 25.5 & 25.5 \\
    DeepSeek V3.2 & 64.3 & 64.3 & 64.3 \\
    Qwen3 235B & 91.0 & 91.0 & 91.0 \\
    Qwen3-coder & 89.8 & 89.8 & 89.8 \\
    Llama 3.1 70B & 32.9 & 25.1 & 25.1 \\
    Llama 3.1 8B & 35.3 & 28.2 & 28.2 \\
    \bottomrule
  \end{tabular}
  \caption{
  Non-agentic pass rates (\%) on the Lean~4 version of ProofNet, using automatically generated shadow sets. 
  }
  \label{tab:proofnet-results}
\end{table}

\subsection{\ourmetric on ProofNet}
\label{sec:proofnet-prelim}

\begin{table}[t]
  \centering
  \footnotesize
  \setlength{\tabcolsep}{4pt}
  \begin{tabular}{lcccc}
    \toprule
    Metric & Precision & Recall & F1 & \edit{Agreement} \\
    \midrule
    Compile      & 0.800 & \textbf{1.000} & \underline{0.889} & \edit{\underline{0.800}} \\
    BLEU         & 0.800 & \textbf{1.000} & \underline{0.889} & \edit{\underline{0.800}} \\
    LLM-as-judge & 0.865 & 0.889 & 0.877 & \edit{\underline{0.800}} \\
    BEq+         & \underline{0.887} & 0.653 & 0.752 & \edit{0.656} \\
    \ourrelaxed  & \textbf{1.000} & \textbf{1.000} & \textbf{1.000} & \edit{\textbf{1.000}} \\
    \ourmetric   & \textbf{1.000} & \textbf{1.000} & \textbf{1.000} & \edit{\textbf{1.000}} \\
    \bottomrule
  \end{tabular}
  \caption{
  Binary agreement of automatic metrics with expert judgment for Llama 3.1 8B
  outputs on ProofNet.
  The best value per column is bold, and the second best is underlined.
  }
  \label{tab:proofnet-agreement}
\end{table}

To test whether \ourmetric transfers beyond \ours, we apply it to the
Lean~4 version of ProofNet~\citep{azerbayev2023proofnet}, an
undergraduate-level benchmark with shorter formalizations.\footnote{We exclude
the 31\% of problems with known faulty reference
formalizations~\citep{poiroux-etal-2025-reliable}.}
We construct a complete shadow set and checker theorems for each problem using
the same LLM-assisted pipeline as for \ours (\Cref{sec:curation}).
All initial drafts pass the Lean completeness checks, so no expert revision is
required.

We first compare the resulting metrics with expert judgments on
Llama~3.1~8B outputs, which include both positive and negative expert labels.
As shown in \Cref{tab:proofnet-agreement}, \ourmetric and
\ourrelaxed both match every expert judgment, while compile rate reaches an
F1 of $0.889$.
This result provides evidence that the shadow checks remain reliable when
applied to ProofNet.

\Cref{tab:proofnet-results} shows that compile rate and \ourmetric are much
closer on ProofNet than on \ours.
Across the 16 models, the gap is $2.1$ points on average and at most $7.8$ points.
The smaller gap is consistent with the shorter and less compositional targets
in ProofNet.
Reference statements in \ours are about \edit{$1.6\times$ longer by line
count}, and its reference solutions contain $4.5$ auxiliary declarations per
problem on average, compared with none in ProofNet
(\Cref{tab:dataset-statistics}).
Most ProofNet targets also have a single conclusion, so their shadow sets
typically require only one or two short checks.
Compile rate is therefore a closer proxy for semantic alignment on ProofNet,
whereas \ours exposes failures that are less visible in shorter
formalizations.

\begin{table*}[t]
  \centering
  \scriptsize
  \setlength{\tabcolsep}{3pt}
  \providecommand{\ShadowAvgsep}{\rule[-0.65ex]{0.45pt}{2.7ex}}
  \resizebox{\textwidth}{!}{%
  \begin{tabular}{lcccccccccccc@{\hspace{3pt}}c@{\hspace{3pt}}cccc}
    \toprule
    & \multicolumn{4}{c}{L1 (n{=}\edit{113})} & \multicolumn{4}{c}{L2 (n{=}52)} & \multicolumn{4}{c}{L3 (n{=}13)} & \ShadowAvgsep & \multicolumn{4}{c}{Average (n{=}\edit{178})} \\
    \cmidrule(lr){2-5}\cmidrule(lr){6-9}\cmidrule(lr){10-13}\cmidrule(lr){15-18}
    Method & Fwd & Fwd\textsubscript{all} & Bwd & \ourmetric & Fwd & Fwd\textsubscript{all} & Bwd & \ourmetric & Fwd & Fwd\textsubscript{all} & Bwd & \ourmetric & \ShadowAvgsep & Fwd & Fwd\textsubscript{all} & Bwd & \ourmetric \\
    \midrule
    Claude Code (Opus 4.6) & \phantom{0}0.9 & \phantom{0}0.9 & \edit{\phantom{0}2.7} & \phantom{0}0.0 & \phantom{0}1.9 & \phantom{0}1.9 & \phantom{0}0.0 & \phantom{0}0.0 & \phantom{0}0.0 & \phantom{0}0.0 & \phantom{0}0.0 & \phantom{0}0.0 & \ShadowAvgsep & \phantom{0}1.1 & \phantom{0}1.1 & \phantom{0}1.7 & \phantom{0}0.0 \\
    \quad + Numina & \edit{15.0} & \edit{15.0} & \edit{\textbf{18.6}} & \edit{\textbf{\phantom{0}8.8}} & 17.3 & 17.3 & 17.3 & 11.5 & \phantom{0}0.0 & \phantom{0}0.0 & \phantom{0}0.0 & \phantom{0}0.0 & \ShadowAvgsep & \edit{14.6} & \edit{14.6} & \edit{16.9} & \edit{\phantom{0}9.0} \\
    Claude Code (Opus 4.8) & \phantom{0}4.4 & \phantom{0}4.4 & \phantom{0}5.3 & \edit{\phantom{0}2.7} & \phantom{0}5.8 & \phantom{0}5.8 & \phantom{0}7.7 & \phantom{0}3.8 & \phantom{0}0.0 & \phantom{0}0.0 & \phantom{0}0.0 & \phantom{0}0.0 & \ShadowAvgsep & \phantom{0}4.5 & \phantom{0}4.5 & \phantom{0}5.6 & \phantom{0}2.8 \\
    \quad + Numina & \edit{\phantom{0}9.7} & \edit{\phantom{0}9.7} & \edit{16.8} & \edit{\phantom{0}6.2} & \textbf{25.0} & \textbf{25.0} & \textbf{38.5} & \textbf{23.1} & \textbf{\phantom{0}7.7} & \textbf{\phantom{0}7.7} & \textbf{\phantom{0}7.7} & \textbf{\phantom{0}7.7} & \ShadowAvgsep & \edit{14.0} & \edit{14.0} & \edit{\textbf{22.5}} & \edit{\textbf{11.2}} \\
    Claude Code (Qwen3 235B) & \phantom{0}0.9 & \phantom{0}0.9 & \phantom{0}1.8 & \phantom{0}0.9 & \phantom{0}0.0 & \phantom{0}0.0 & \phantom{0}0.0 & \phantom{0}0.0 & \phantom{0}0.0 & \phantom{0}0.0 & \phantom{0}0.0 & \phantom{0}0.0 & \ShadowAvgsep & \phantom{0}0.6 & \phantom{0}0.6 & \phantom{0}1.1 & \phantom{0}0.6 \\
    \quad + Numina & \phantom{0}0.9 & \phantom{0}0.9 & \phantom{0}1.8 & \phantom{0}0.9 & \phantom{0}0.0 & \phantom{0}0.0 & \phantom{0}0.0 & \phantom{0}0.0 & \phantom{0}0.0 & \phantom{0}0.0 & \phantom{0}0.0 & \phantom{0}0.0 & \ShadowAvgsep & \phantom{0}0.6 & \phantom{0}0.6 & \phantom{0}1.1 & \phantom{0}0.6 \\
    Codex (GPT-5.4) & \phantom{0}3.5 & \phantom{0}3.5 & \phantom{0}5.3 & \edit{\phantom{0}2.7} & \phantom{0}1.9 & \phantom{0}1.9 & \phantom{0}1.9 & \phantom{0}1.9 & \phantom{0}0.0 & \phantom{0}0.0 & \phantom{0}0.0 & \phantom{0}0.0 & \ShadowAvgsep & \phantom{0}2.8 & \phantom{0}2.8 & \phantom{0}3.9 & \phantom{0}2.2 \\
    \quad + Numina & \edit{\textbf{17.7}} & \edit{\textbf{17.7}} & \edit{14.2} & \edit{\textbf{\phantom{0}8.8}} & 19.2 & 19.2 & 23.1 & 13.5 & \textbf{\phantom{0}7.7} & \textbf{\phantom{0}7.7} & \textbf{\phantom{0}7.7} & \textbf{\phantom{0}7.7} & \ShadowAvgsep & \edit{\textbf{17.4}} & \edit{\textbf{17.4}} & \edit{16.3} & \edit{10.1} \\
    \bottomrule
  \end{tabular}}
  \caption{
    Per-difficulty checker metrics (\%) for agentic systems on \ours.
    Fwd and Bwd average the per-problem fractions of passed forward and backward
    checkers, Fwd\textsubscript{all} is the fraction of problems passing every
    forward checker.
  }
  \label{tab:shadow-forward-backward}
\end{table*}

\subsection{Forward and Backward Checker Analysis}
\label{sec:forward-backward-analysis}

The forward and backward checks used by \ourmetric detect different forms of
semantic mismatch.
A successful forward check certifies that the generated statement implies a
required consequence of the intended theorem, while a successful backward
check certifies that the intended theorem implies the generated statement.
Passing the backward checks but not all forward checks can arise from omitted
conclusions or added assumptions (\Cref{app:cs-weakening}).
Conversely, passing all forward checks but failing the backward checks can
arise from added conclusions or omitted assumptions. 
When both directions fail, neither implication is certified (\Cref{app:cs-brahmagupta}).

\Cref{tab:shadow-forward-backward} shows that the two implication directions
can differ substantially even when \ourmetric is low.
\edit{Claude Code (Opus~4.8) with Numina-Lean-Agent reaches a
\ourmetric score of $11.2\%$}, with a \edit{$14.0\%$} all forward pass rate but
\edit{$22.5\%$} average backward pass rate.
Among its \edit{110} compile-passed outputs, \edit{20} pass both directions, \edit{20} pass only
the backward direction, \edit{5} pass only the forward direction, and \edit{65} pass
neither.

\section{Conclusion}
\label{sec:conclusion}

We introduced \ourmetric, 
an automatic metric that evaluates semantic alignment
between generated formal statements and informal theorem statements.
For each problem, \ourmetric uses a set of auxiliary formal statements,
called shadows, whose conjunction is equivalent to the intended theorem statement,
and checks in Lean whether the generated statement implies each shadow
and is implied by their conjunction.
We instantiated \ourmetric in \ours, 
a Lean~4 full autoformalization benchmark of
\edit{178} postgraduate- to research-level problems spanning eight mathematical areas.
On \ours, Claude Code (Opus~4.8) with Numina-Lean-Agent reaches 
\edit{$61.8\%$} compile rate, 
\edit{$18.3\%$} \ourrelaxed, 
and \edit{$11.2\%$} \ourmetric, 
showing that compile success alone can overstate semantic alignment.
\edit{Across six agentic configurations, \ourmetric achieves
$98.8\%$ binary agreement with expert judgments.}

%
%

\section*{Limitations}

\ours covers \edit{178} postgraduate- to research-level problems across eight
mathematical areas, and future versions can extend this coverage to more
domains and proof styles.
\ourmetric requires one-time construction of shadow statements and checker proofs for each benchmark problem.

\section*{Potential Risks}

The main risk is benchmark leakage.
If checker artifacts are exposed, systems can optimize for the checks rather
than the informal theorem.
To reduce this risk, public problem inputs are separated from the checking
artifacts used for evaluation.
We give the full release and evaluation protocol in \Cref{sec:release}.

\section*{Acknowledgments}
This work was supported by
the National Research Foundation of Korea (NRF) grant funded by the Korea government (MSIT) (No. RS-2025-00520280),
Institute of Information \& communications Technology Planning \& Evaluation (IITP) grant funded by the Korea government(MSIT) [NO.RS-2021-II211343, Artificial Intelligence Graduate School Program (Seoul National University)],
Electronics and Telecommunications Research Institute (ETRI) grant funded by ICT R\&D program of MSIT/IITP (2022-0-00995, Automated reliable source code generation from natural language descriptions).
We thank Soonho Kong for supporting our experiments with Claude and Claude Code on Amazon Web Services.


\newpage
\bibliography{references}

@inproceedings{poiroux-etal-2025-reliable,
    title = "Reliable Evaluation and Benchmarks for Statement Autoformalization",
    author = "Poiroux, Auguste  and
      Weiss, Gail  and
      Kun{\v{c}}ak, Viktor  and
      Bosselut, Antoine",
    editor = "Christodoulopoulos, Christos  and
      Chakraborty, Tanmoy  and
      Rose, Carolyn  and
      Peng, Violet",
    booktitle = "Proceedings of the 2025 Conference on Empirical Methods in Natural Language Processing",
    month = nov,
    year = "2025",
    address = "Suzhou, China",
    publisher = "Association for Computational Linguistics",
    url = "https://aclanthology.org/2025.emnlp-main.907/",
    doi = "10.18653/v1/2025.emnlp-main.907",
    pages = "17947--17969",
    ISBN = "979-8-89176-332-6"
}

@inproceedings{kwon2023efficient.vllm,
  title={Efficient Memory Management for Large Language Model Serving with {PagedAttention}},
  author={Woosuk Kwon and Zhuohan Li and Siyuan Zhuang and Ying Sheng and Lianmin Zheng and Cody Hao Yu and Joseph E. Gonzalez and Hao Zhang and Ion Stoica},
  booktitle={Proceedings of the ACM SIGOPS 29th Symposium on Operating Systems Principles},
  year={2023}
}

@inproceedings{mathlib2020,
author    = {{The mathlib Community}},
title     = {The {L}ean {M}athematical {L}ibrary},
booktitle = {Proceedings of the 9th {ACM} {SIGPLAN} International Conference
       on Certified Programs and Proofs},
series    = {CPP 2020},
publisher = {ACM},
address   = {New Orleans, LA, USA},
year      = {2020},
month     = jan,
doi       = {10.1145/3372885.3373824},
url       = {https://doi.org/10.1145/3372885.3373824}
}

@misc{shapiro_mat520_functional_analysis,
author       = {Shapiro, Jacob},
title        = {Functional Analysis: Princeton University MAT520 Lecture Notes},
institution  = {Princeton University},
howpublished = {\url{https://web.math.princeton.edu/~js129/PDFs/teaching/MAT520_fall_2023/MAT520_Lecture_Notes.pdf}},
note         = {Created August 18, 2023; last typeset September 5, 2024; accessed May 26, 2026},
year         = {2024}
}

@book{boyd2004convex,
title={Convex optimization},
author={Boyd, Stephen and Vandenberghe, Lieven},
year={2004},
publisher={Cambridge university press}
}

@book{cox2008ideals,
title={Ideals, Varieties, and Algorithms: An Introduction to Computational Algebraic Geometry and Commutative Algebra},
author={Cox, D.A. and Little, J. and O'Shea, D.},
isbn={9780387514857},
lccn={89021569},
series={Undergraduate Texts in Mathematics},
url={https://books.google.co.kr/books?id=qs9fAQAACAAJ},
year={2008},
publisher={Springer New York}
}

@book{dieudonne1971elements,
title={{\'E}l{\'e}ments de g{\'e}om{\'e}trie alg{\'e}brique},
author={Dieudonne, Jean Alexandre and Grothendieck, Alexandre},
volume={166},
year={1971},
publisher={Springer Berlin Heidelberg New York}
}

@incollection{lee2003smooth,
title={Smooth manifolds},
author={Lee, John M},
booktitle={Introduction to smooth manifolds},
pages={1--29},
year={2003},
publisher={Springer}
}

@misc{stacks-project,
author       = {The {Stacks project authors}},
title        = {The Stacks project},
howpublished = {\url{https://stacks.math.columbia.edu}},
year         = {2026},
}

@book{stein2009real,
title={Real analysis: measure theory, integration, and Hilbert spaces},
author={Stein, Elias M and Shakarchi, Rami},
year={2009},
publisher={Princeton University Press}
}

@book{wunsch2005complex,
title={Complex variables with applications},
author={Wunsch, A David},
year={2005},
publisher={Pearson Education India}
}

@article{agrawal2004primes,
title={PRIMES is in P},
author={Agrawal, Manindra and Kayal, Neeraj and Saxena, Nitin},
journal={Annals of mathematics},
pages={781--793},
year={2004},
publisher={JSTOR}
}

@article{aigner1999proofs,
title={Proofs from the Book},
author={Aigner, Martin and Ziegler, G{\"u}nter M},
journal={Berlin. Germany},
volume={1},
number={2},
pages={7},
year={1999},
publisher={Springer}
}

@misc{theorems1000plus_euler_quadrilateral,
author       = {{1000+ Theorems contributors}},
title        = {{1000+ Theorems}: Euler's quadrilateral theorem},
howpublished = {\url{https://1000-plus.github.io/all}},
note         = {Entry ``Euler's quadrilateral theorem''; accessed 2026-05-26},
year         = {2026}
}

@article{claude48opus,
  title   = {{System Card: Claude Opus 4.8}},
  author  = {{Anthropic}},
  journal = {Claude Opus 4.8 Model Card},
  year    = {2026},
  url     = {https://www-cdn.anthropic.com/0b4915911bb0d19eca5b5ee635c80fef830a37ea.pdf}
}

@article{claude46opus,
  title   = {{System Card: Claude Opus 4.6}},
  author  = {{Anthropic}},
  journal = {Claude Opus 4.6 Model Card},
  year    = {2026},
  url     = {https://www-cdn.anthropic.com/14e4fb01875d2a69f646fa5e574dea2b1c0ff7b5.pdf}
}

@article{claude46sonnet,
  title   = {{System Card: Claude Sonnet 4.6}},
  author  = {{Anthropic}},
  journal = {Claude Sonnet 4.6 Model Card},
  year    = {2026},
  url     = {https://www-cdn.anthropic.com/78073f739564e986ff3e28522761a7a0b4484f84.pdf}
}

@misc{claude45haiku,
  title  = {{System Card: Claude Haiku 4.5}},
  author = {{Anthropic}},
  year   = {2025},
  url    = {https://assets.anthropic.com/m/99128ddd009bdcb/original/Claude-Haiku-4-5-System-Card.pdf}
}

@misc{gpt54,
  title  = {{GPT-5.4 Thinking System Card}},
  author = {{OpenAI}},
  year   = {2026},
  url    = {https://deploymentsafety.openai.com/gpt-5-4-thinking/gpt-5-4-thinking.pdf}
}

@misc{gpt54mininano,
  title  = {{Introducing GPT-5.4 mini and nano}},
  author = {{OpenAI}},
  year   = {2026},
  url    = {https://openai.com/index/introducing-gpt-5-4-mini-and-nano/}
}

@article{gemini31pro,
  title   = {{Gemini 3.1 Pro Model Card}},
  author  = {{Google DeepMind}},
  journal = {Gemini 3.1 Pro Model Card},
  year    = {2026},
  url     = {https://storage.googleapis.com/deepmind-media/Model-Cards/Gemini-3-1-Pro-Model-Card.pdf}
}

@misc{gemini25flash,
  title  = {{Gemini 2.5 Flash Model Card}},
  author = {{Google DeepMind}},
  year   = {2025},
  url    = {https://storage.googleapis.com/deepmind-media/Model-Cards/Gemini-2-5-Flash-Model-Card.pdf}
}

@article{gptoss120b,
  title   = {gpt-oss-120b \& gpt-oss-20b model card},
  author  = {Agarwal, Sandhini and Ahmad, Lama and Ai, Jason and Altman, Sam and Applebaum, Andy and Arbus, Edwin and Arora, Rahul K. and Bai, Yu and Baker, Bowen and Bao, Haiming and others},
  journal = {arXiv preprint arXiv:2508.10925},
  year    = {2025},
  url     = {https://arxiv.org/pdf/2508.10925}
}

@article{deepseekr1,
  title   = {{DeepSeek-R1} incentivizes reasoning in {LLMs} through reinforcement learning},
  author  = {Guo, Daya and Yang, Dejian and Zhang, Haowei and Song, Junxiao and Wang, Peiyi and Zhu, Qihao and Xu, Runxin and Zhang, Ruoyu and Ma, Shirong and Bi, Xiao and others},
  journal = {Nature},
  volume  = {645},
  pages   = {633--638},
  year    = {2025},
  doi     = {10.1038/s41586-025-09422-z},
  url     = {https://www.nature.com/articles/s41586-025-09422-z}
}

@article{deepseekv32,
  title   = {{DeepSeek-V3.2}: Pushing the Frontier of Open Large Language Models},
  author  = {{DeepSeek-AI}},
  journal = {arXiv preprint arXiv:2512.02556},
  year    = {2025},
  url     = {https://arxiv.org/abs/2512.02556}
}

@article{goedelproverv2,
  title   = {{Goedel-Prover-V2}: Scaling Formal Theorem Proving with Scaffolded Data Synthesis and Self-Correction},
  author  = {Lin, Yong and Tang, Shange and Lyu, Bohan and Yang, Ziran and Chung, Jui-Hui and Zhao, Haoyu and Jiang, Lai and Geng, Yihan and Ge, Jiawei and Sun, Jingruo and Wu, Jiayun and Gesi, Jiri and Lu, Ximing and Acuna, David and Yang, Kaiyu and Lin, Hongzhou and Choi, Yejin and Chen, Danqi and Arora, Sanjeev and Jin, Chi},
  journal = {arXiv preprint arXiv:2508.03613},
  year    = {2025},
  url     = {https://arxiv.org/pdf/2508.03613}
}

@article{yang2025qwen3,
  title={{Qwen3} Technical Report},
  author={Yang, An and Li, Anfeng and Yang, Baosong and Zhang, Beichen and Hui, Binyuan and Zheng, Bo and Yu, Bowen and Gao, Chang and Huang, Chengen and Lv, Chenxu and Zheng, Chujie and Liu, Dayiheng and Zhou, Fan and Huang, Fei and Hu, Feng and Ge, Hao and Wei, Haoran and Lin, Huan and Tang, Jialong and Yang, Jian and Tu, Jianhong and Zhang, Jianwei and Yang, Jianxin and Yang, Jiaxi and Zhou, Jing and Zhou, Jingren and Lin, Junyang and Dang, Kai and Bao, Keqin and Yang, Kexin and Yu, Le and Deng, Lianghao and Li, Mei and Xue, Mingfeng and Li, Mingze and Zhang, Pei and Wang, Peng and Zhu, Qin and Men, Rui and Gao, Ruize and Liu, Shixuan and Luo, Shuang and Li, Tianhao and Tang, Tianyi and Yin, Wenbiao and Ren, Xingzhang and Wang, Xinyu and Zhang, Xinyu and Ren, Xuancheng and Fan, Yang and Su, Yang and Zhang, Yichang and Zhang, Yinger and Wan, Yu and Liu, Yuqiong and Wang, Zekun and Cui, Zeyu and Zhang, Zhenru and Zhou, Zhipeng and Qiu, Zihan},
  journal={arXiv preprint arXiv:2505.09388},
  year={2025}
}

@article{grattafiori2024llama,
  title={The llama 3 herd of models},
  author={Grattafiori, Aaron and Dubey, Abhimanyu and Jauhri, Abhinav and Pandey, Abhinav and Kadian, Abhishek and Al-Dahle, Ahmad and Letman, Aiesha and Mathur, Akhil and Schelten, Alan and Vaughan, Alex and others},
  journal={arXiv preprint arXiv:2407.21783},
  year={2024}
}

@article{guo2025autoformalizer,
  title={Autoformalizer with Tool Feedback},
  author={Guo, Qi and Wang, Jianing and Zhang, Jianfei and Kong, Deyang and Huang, Xiangzhou and Xi, Xiangyu and Wang, Wei and Wang, Jingang and Cai, Xunliang and Zhang, Shikun and Ye, Wei},
  journal={arXiv preprint arXiv:2510.06857},
  year={2025}
}

@inproceedings{papineni2002bleu,
  title = {{BLEU}: a Method for Automatic Evaluation of Machine Translation},
  author = {Papineni, Kishore and Roukos, Salim and Ward, Todd and Zhu, Wei-Jing},
  booktitle = {Proceedings of the 40th Annual Meeting of the Association for Computational Linguistics},
  pages = {311--318},
  year = {2002},
  doi = {10.3115/1073083.1073135}
}

@article{chen2021codex,
  title = {Evaluating Large Language Models Trained on Code},
  author = {Chen, Mark and Tworek, Jerry and Jun, Heewoo and Yuan, Qiming and Pinto, Henrique Ponde de Oliveira and Kaplan, Jared and Edwards, Harri and Burda, Yuri and Joseph, Nicholas and Brockman, Greg and Ray, Alex and Puri, Raul and Krueger, Gretchen and Petrov, Michael and Khlaaf, Heidy and Sastry, Girish and Mishkin, Pamela and Chan, Brooke and Gray, Scott and Ryder, Nick and Pavlov, Mikhail and Power, Alethea and Kaiser, Lukasz and Bavarian, Mohammad and Winter, Clemens and Tillet, Philippe and Such, Felipe Petroski and Cummings, Dave and Plappert, Matthias and Chantzis, Fotios and Barnes, Elizabeth and Herbert-Voss, Ariel and Guss, William Hebgen and Nichol, Alex and Paino, Alex and Tezak, Nikolas and Tang, Jie and Babuschkin, Igor and Balaji, Suchir and Jain, Shantanu and Saunders, William and Hesse, Christopher and Carr, Andrew N. and Leike, Jan and Achiam, Josh and Misra, Vedant and Morikawa, Evan and Radford, Alec and Knight, Matthew and Brundage, Miles and Murati, Mira and Mayer, Katie and Welinder, Peter and McGrew, Bob and Amodei, Dario and McCandlish, Sam and Sutskever, Ilya and Zaremba, Wojciech},
  journal = {arXiv preprint arXiv:2107.03374},
  year = {2021},
  url = {https://arxiv.org/abs/2107.03374}
}

@inproceedings{kim2026benchmarking,
  title = {Benchmarking Testing in Automated Theorem Proving},
  author = {Kim, Jongyoon and Han, Hojae and Hwang, Seung-Won},
  booktitle = {Proceedings of the 64th Annual Meeting of the Association for Computational Linguistics (Volume 6: Industry Track)},
  month = jul,
  year = {2026},
  address = {San Diego, California, USA},
  publisher = {Association for Computational Linguistics},
  pages = {2241--2260},
  doi = {10.18653/v1/2026.acl-industry.150},
  url = {https://aclanthology.org/2026.acl-industry.150/}
}

@article{azerbayev2023proofnet,
  title = {{ProofNet}: Autoformalizing and Formally Proving Undergraduate-Level Mathematics},
  author = {Azerbayev, Zhangir and Piotrowski, Bartosz and Schoelkopf, Hailey and Ayers, Edward W. and Radev, Dragomir and Avigad, Jeremy},
  journal = {arXiv preprint arXiv:2302.12433},
  year = {2023},
  url = {https://arxiv.org/abs/2302.12433}
}

@article{lu2024processdriven,
  title = {Process-Driven Autoformalization in {Lean} 4},
  author = {Lu, Jianqiao and Wan, Yingjia and Liu, Zhengying and Huang, Yinya and Xiong, Jing and Liu, Chengwu and Shen, Jianhao and Jin, Hui and Zhang, Jipeng and Wang, Haiming and Yang, Zhicheng and Tang, Jing and Guo, Zhijiang},
  journal = {arXiv preprint arXiv:2406.01940},
  year = {2024},
  url = {https://arxiv.org/abs/2406.01940}
}

@inproceedings{li2026proofbridge,
  title = {{ProofBridge}: Auto-Formalization of Natural Language Proofs in {Lean} via Joint Embeddings},
  author = {Jana, Prithwish and Kale, Kaan and Tanriverdi, Ahmet Ege and Song, Cruise and Vishwanath, Sriram and Ganesh, Vijay},
  booktitle = {International Conference on Learning Representations},
  year = {2026},
  note = {Poster},
  url = {https://openreview.net/forum?id=U2jxHXuOX9}
}

@inproceedings{poiroux2025rlmeval,
  title = {{RLME}val: Evaluating Research-Level Neural Theorem Proving},
  author = {Poiroux, Auguste and Bosselut, Antoine and Kun{\v{c}}ak, Viktor},
  editor = {Christodoulopoulos, Christos and Chakraborty, Tanmoy and Rose, Carolyn and Peng, Violet},
  booktitle = {Findings of the Association for Computational Linguistics: EMNLP 2025},
  address = {Suzhou, China},
  publisher = {Association for Computational Linguistics},
  month = nov,
  year = {2025},
  pages = {10946--10957},
  doi = {10.18653/v1/2025.findings-emnlp.581},
  url = {https://aclanthology.org/2025.findings-emnlp.581/}
}

@inproceedings{wu2022autoformalization,
  title = {Autoformalization with Large Language Models},
  author = {Wu, Yuhuai and Jiang, Albert Q. and Li, Wenda and Rabe, Markus N. and Staats, Charles and Jamnik, Mateja and Szegedy, Christian},
  booktitle = {Advances in Neural Information Processing Systems},
  year = {2022},
  url = {https://arxiv.org/abs/2205.12615}
}

@inproceedings{jiang2023herald,
  title = {{Herald}: A Natural Language Annotated {Lean} 4 Dataset},
  author = {Gao, Guoxiong and Wang, Yutong and Jiang, Jiedong and Gao, Qi and Qin, Zihan and Xu, Tianyi and Dong, Bin},
  booktitle = {International Conference on Learning Representations},
  year = {2025},
  note = {Poster},
  url = {https://openreview.net/forum?id=Se6MgCtRhz}
}

@inproceedings{wu2024leanworkbook,
  title = {{Lean Workbook}: A Large-Scale {Lean} Problem Set Formalized from Natural Language Math Problems},
  author = {Ying, Huaiyuan and Wu, Zijian and Geng, Yihan and Wang, Jiayu and Lin, Dahua and Chen, Kai},
  booktitle = {Advances in Neural Information Processing Systems},
  year = {2024},
  note = {Datasets and Benchmarks Track Poster},
  url = {https://openreview.net/forum?id=Vcw3vzjHDb}
}

@inproceedings{murphy2024leaneuclid,
  title = {Autoformalizing {Euclidean} Geometry},
  author = {Murphy, Logan and Yang, Kaiyu and Sun, Jialiang and Li, Zhaoyu and Anandkumar, Anima and Si, Xujie},
  booktitle = {Proceedings of the 41st International Conference on Machine Learning},
  volume = {235},
  pages = {36847--36893},
  year = {2024},
  publisher = {PMLR},
  url = {https://proceedings.mlr.press/v235/murphy24a.html}
}

@inproceedings{wang2025nl2lean,
  title = {{NL}2{L}ean: Translating Natural Language into {Lean} 4 through Multi-Aspect Reinforcement Learning},
  author = {Fang, Yue and Huang, Shaohan and Yu, Xin and Huang, Haizhen and Zhang, Zihan and Deng, Weiwei and Wei, Furu and Sun, Feng and Zhang, Qi and Jin, Zhi},
  editor = {Christodoulopoulos, Christos and Chakraborty, Tanmoy and Rose, Carolyn and Peng, Violet},
  booktitle = {Proceedings of the 2025 Conference on Empirical Methods in Natural Language Processing},
  address = {Suzhou, China},
  publisher = {Association for Computational Linguistics},
  month = nov,
  year = {2025},
  pages = {31148--31158},
  doi = {10.18653/v1/2025.emnlp-main.1586},
  url = {https://aclanthology.org/2025.emnlp-main.1586/}
}

@inproceedings{li2026reform,
  title = {{ReForm}: Reflective Autoformalization with Prospective Bounded Sequence Optimization},
  author = {Chen, Guoxin and Wu, Jing and Chen, Xinjie and Zhao, Wayne Xin and Song, Ruihua and Li, Chengxi and Fan, Kai and Liu, Dayiheng and Liao, Minpeng},
  booktitle = {International Conference on Learning Representations},
  year = {2026},
  note = {Poster},
  url = {https://openreview.net/forum?id=KfxRzCmRSX}
}

@inproceedings{li2025rethinking,
  title = {Rethinking and Improving Autoformalization: Towards a Faithful Metric and a Dependency Retrieval-based Approach},
  author = {Liu, Qi and Zheng, Xinhao and Lu, Xudong and Cao, Qinxiang and Yan, Junchi},
  booktitle = {International Conference on Learning Representations},
  year = {2025},
  note = {Spotlight},
  url = {https://openreview.net/forum?id=hUb2At2DsQ}
}

@inproceedings{ying2026assess,
  title = {{ASSESS}: A Semantic and Structural Evaluation Framework for Statement Similarity},
  author = {Liu, Xiaoyang and Zhu, Tao and Dong, Zineng and Liu, Yuntian and Guo, Qingfeng and Liu, Zhaoxuan and Chen, Yu and Luo, Tao},
  booktitle = {International Conference on Learning Representations},
  year = {2026},
  url = {https://openreview.net/forum?id=avwNGWtiHF}
}

@inproceedings{cheng2025fmc,
  title = {{FMC}: Formalization of Natural Language Mathematical Competition Problems},
  author = {Xie, Jiaxuan and Liu, Chengwu and Yuan, Ye and Li, Siqi and Xiao, Zhiping and Zhang, Ming},
  booktitle = {2nd AI for Math Workshop at ICML 2025},
  year = {2025},
  url = {https://icml.cc/virtual/2025/52460}
}

@inproceedings{li2025atlas,
  title = {{ATLAS}: Autoformalizing Theorems through Lifting, Augmentation, and Synthesis of Data},
  author = {Liu, Xiaoyang and Bao, Kangjie and Zhang, Jiashuo and Liu, Yunqi and Chen, Yu and Liu, Yuntian and Jiao, Yang and Luo, Tao},
  booktitle = {Advances in Neural Information Processing Systems},
  year = {2025},
  url = {https://openreview.net/forum?id=MlJyAvQaxp}
}

@article{lu2025conceptretrieval,
  title = {Automated Formalization via Conceptual Retrieval-Augmented {LLM}s},
  author = {Lu, Wangyue and Du, Lun and Li, Sirui and Weng, Ke and Sun, Haozhe and Liu, Hengyu and Yu, Minghe and Zhang, Tiancheng and Yu, Ge},
  journal = {arXiv preprint arXiv:2508.06931},
  year = {2025},
  url = {https://arxiv.org/abs/2508.06931}
}

@inproceedings{zheng2022minif2f,
  title = {{miniF2F}: A Cross-System Benchmark for Formal {Olympiad}-Level Mathematics},
  author = {Zheng, Kunhao and Han, Jesse Michael and Polu, Stanislas},
  booktitle = {International Conference on Learning Representations},
  year = {2022},
  url = {https://openreview.net/forum?id=9ZPegFuFTFv}
}

@inproceedings{yang2023leandojo,
  title = {{LeanDojo}: Theorem Proving with Retrieval-Augmented Language Models},
  author = {Yang, Kaiyu and Swope, Aidan M. and Gu, Alex and Chalamala, Rahul and Song, Peiyang and Yu, Shixing and Godil, Saad and Prenger, Ryan and Anandkumar, Anima},
  booktitle = {Advances in Neural Information Processing Systems},
  year = {2023},
  url = {https://openreview.net/forum?id=g7OX2sOJtn}
}

@article{xin2024deepseekprover,
  title = {{DeepSeek-Prover}: Advancing Theorem Proving in {LLM}s through Large-Scale Synthetic Data},
  author = {Xin, Huajian and Guo, Daya and Shao, Zhihong and Ren, Z. Z. and Zhu, Qihao and Liu, Bo and Ruan, Chong and Li, Wenda and Liang, Xiaodan},
  journal = {arXiv preprint arXiv:2405.14333},
  year = {2024},
  url = {https://arxiv.org/abs/2405.14333}
}

@inproceedings{lin2025goedelprover,
  title = {{Goedel-Prover}: A Frontier Model for Open-Source Automated Theorem Proving},
  author = {Lin, Yong and Tang, Shange and Lyu, Bohan and Wu, Jiayun and Lin, Hongzhou and Yang, Kaiyu and Li, Jia and Xia, Mengzhou and Chen, Danqi and Arora, Sanjeev and Jin, Chi},
  booktitle = {Conference on Language Modeling},
  year = {2025},
  url = {https://openreview.net/forum?id=x2y9i2HDjD}
}

@article{wang2025kiminaprover,
  title = {{Kimina-Prover} Preview: Towards Large Formal Reasoning Models with Reinforcement Learning},
  author = {Wang, Haiming and Unsal, Mert and Lin, Xiaohan and Baksys, Mantas and Liu, Junqi and Dos Santos, Marco and Sung, Flood and Vinyes, Marina and Ying, Zhenzhe and Zhu, Zekai and Lu, Jianqiao and de Saxc{\'e}, Hugues and Bailey, Bolton and Song, Chendong and Xiao, Chenjun and Zhang, Dehao and Zhang, Ebony and Pu, Frederick and Zhu, Han and Liu, Jiawei and Bayer, Jonas and Michel, Julien and Yu, Longhui and Dreyfus-Schmidt, L{\'e}o and Tunstall, Lewis and Pagani, Luigi and Machado, Moreira and Bourigault, Pauline and Wang, Ran and Polu, Stanislas and Barroyer, Thibaut and Li, Wen-Ding and Niu, Yazhe and Fleureau, Yann and Hu, Yangyang and Yu, Zhouliang and Wang, Zihan and Yang, Zhilin and Liu, Zhengying and Li, Jia},
  journal = {arXiv preprint arXiv:2504.11354},
  year = {2025},
  url = {https://arxiv.org/abs/2504.11354}
}

@misc{kiminaprover2025full,
  title  = {{Kimina-Prover}: Applying Test-time {RL} Search on Large Formal Reasoning Models},
  author = {{Numina and Kimi Team}},
  year   = {2025},
  url    = {https://huggingface.co/blog/AI-MO/kimina-prover}
}

@article{numina2026leanagent,
  title = {{Numina-Lean-Agent}: An Open and General Agentic Reasoning System for Formal Mathematics},
  author = {Liu, Junqi and Zhou, Zihao and Zhu, Zekai and Dos Santos, Marco and He, Weikun and Liu, Jiawei and Wang, Ran and Xie, Yunzhou and Zhao, Junqiao and Wang, Qiufeng and Zhi, Lihong and Li, Jia and Li, Wenda},
  journal = {arXiv preprint arXiv:2601.14027},
  year = {2026},
  url = {https://arxiv.org/abs/2601.14027}
}

@inproceedings{liu2025formalalign,
  title = {{FormalAlign}: Automated Alignment Evaluation for Autoformalization},
  author = {Lu, Jianqiao and Wan, Yingjia and Huang, Yinya and Xiong, Jing and Liu, Zhengying and Guo, Zhijiang},
  booktitle = {International Conference on Learning Representations},
  year = {2025},
  url = {https://openreview.net/forum?id=B5RrIFMqbe}
}

@inproceedings{chen2025minif2fv2,
  title = {{miniF2F-Lean} Revisited: Reviewing Limitations and Charting a Path Forward},
  author = {Ospanov, Azim and Farnia, Farzan and Yousefzadeh, Roozbeh},
  booktitle = {Advances in Neural Information Processing Systems},
  year = {2025},
  note = {Poster},
  url = {https://openreview.net/forum?id=KtaHv0YUyh}
}

@inproceedings{ammanamanchi2026faults,
  title = {Faults in Our Formal Benchmarking: Dataset Defects and Evaluation Failures in {Lean} Theorem Proving},
  author = {Ammanamanchi, Pawan Sasanka and Bhat, Siddharth and Biderman, Stella},
  booktitle = {Forty-third International Conference on Machine Learning},
  year = {2026},
  url = {https://openreview.net/forum?id=bHYAWawd4A}
}

\clearpage
\newpage

\appendix
\section{Dataset Details}
\label{sec:appendix:dataset_details}
\label{app:curator}

\begin{table*}[b]
  \centering
  \scriptsize
  \setlength{\tabcolsep}{3pt}
  \renewcommand{\arraystretch}{1.12}
  \begin{tabular}{p{0.42\textwidth}p{0.34\textwidth}cc}
    \toprule
    \textbf{Source} & \textbf{Reference} & \textbf{Informal} & \textbf{Formal} \\
    \midrule
    Analysis problem & \citep{shapiro_mat520_functional_analysis} & O & X \\
    Complex Variables with Applications & \citep{wunsch2005complex} & O & X \\
    Convex Optimization & \citep{boyd2004convex} & O & X \\
    Elements de geometrie algebrique (EGA) & \citep{dieudonne1971elements} & O & X \\
    Ideals, Varieties, and Algorithms (4th ed.) & \citep{cox2008ideals} & O & X \\
    Introduction to Smooth Manifolds & \citep{lee2003smooth} & O & X \\
    Mathlib 4 & \citep{mathlib2020} & X & O \\
    Missing theorems from Wiedijk 1000+ & \citep{theorems1000plus_euler_quadrilateral} & O & X \\
    PRIMES is in P & \citep{agrawal2004primes} & O & X \\
    Proofs from THE BOOK & \citep{aigner1999proofs} & O & X \\
    Real Analysis: Measure Theory, Integration, and Hilbert Spaces & \citep{stein2009real} & O & X \\
    The Stacks Project & \citep{stacks-project} & $\triangle$ & $\triangle$ \\
    \bottomrule
  \end{tabular}
  \caption{
  List of reference sources used to construct \ours.
  \emph{Informal} and \emph{Formal} mark the statement type available before annotation. 
  A triangle denotes mixed availability across problems.
  }
  \label{tab:repository_detail}
\end{table*}

\begin{table}[H]
  \centering
  \footnotesize
  \setlength{\tabcolsep}{4pt}
  \begin{tabular}{llrr}
    \toprule
    \multicolumn{2}{l}{Dataset} & \ours & ProofNet \\
    \midrule
    \multicolumn{2}{l}{$n$} & \edit{178} & 371 \\
    \midrule
    \multirow{2}{*}{\shortstack[l]{Target\\statement}}
      & Lines       & 5.1  & 3.2 \\
      & Chars       & 226  & 149 \\
    \midrule
    \multirow{3}{*}{\shortstack[l]{Reference\\proof}}
      & Lines       & 71.6 & 3.4 \\
      & Chars       & 3242 & 157 \\
      & Aux.\ decl. & 4.5  & 0.0 \\
    \bottomrule
  \end{tabular}
  \caption{
Statistics of the reference statement and proof for both \ours and ProofNet (Lean~4).
  }
  \label{tab:dataset-statistics}
\end{table}

\subsection{Problem Sources}
\label{sec:appendix:sources}

\Cref{tab:repository_detail} lists the sources used to construct \ours.
Annotators collect problems from textbooks, lecture notes, research papers, and
Lean~4 library material.
The Informal and Formal columns mark the statement type available in each source
before annotation.
When a source provides only one side, Qwen3-235B drafts the other, as described
in \Cref{sec:curation}.


\subsection{Dataset Statistics}
\label{sec:appendix:stats}

\Cref{tab:dataset-statistics} compares reference Lean statistics of \ours
with the Lean 4 version of ProofNet.
Target statements in \ours are about $1.6\times$ longer by line count than those in ProofNet.
The larger difference lies in the reference proofs.
A reference proof in \ours averages $72$ lines and $4.5$ auxiliary declarations, while ProofNet averages about $3.4$ lines and none of auxiliary declarations.
\Cref{tab:hidden-checker-statistics} reports the number of checker theorems per problem.

\section{Experiment Details}
\label{sec:appendix:experiment_details}

\subsection{Generation Setup}
\label{sec:appendix:generation_setup}

Closed-source models and general-purpose open-source models are accessed through OpenRouter.
Lean-specialized models are self-hosted with vLLM \cite{kwon2023efficient.vllm}.
These models are executed
on two NVIDIA Titan RTX (24GB) for model sizes up to 8B,
on two NVIDIA RTX 6000 PRO Ada Generation (48GB) for the 32B model,
and on eight NVIDIA RTX 6000 PRO Black Edition (96GB) for the 235B model.
All non-agentic runs use zero-shot prompting with temperature $0.6$.
For each non-agentic problem run, we generate a single completion.
For the agentic setting, each harness is evaluated with and without Numina-Lean-Agent \cite{numina2026leanagent}.

\subsection{Model Details}
\label{sec:appendix:model_details}

\begin{table}[!h]
  \centering
  \setlength{\tabcolsep}{4pt}
  \renewcommand{\arraystretch}{1.1}
  \scalebox{0.7}{
  \begin{tabular}{lll}
    \toprule
    \textbf{Base system} & \textbf{Backbone} & \textbf{Thinking} \\
    \midrule
    Claude Code & Claude Opus 4.6~\citep{claude46opus} & Medium \\
    Claude Code & Claude Opus 4.8~\citep{claude48opus} & Medium \\
    Claude Code & Qwen3 235B~\citep{yang2025qwen3} & - \\
    Codex & GPT-5.4~\citep{gpt54} & Medium \\
    \bottomrule
  \end{tabular}
  }
  \caption{Configuration for agentic harnesses.}
  \label{tab:agentic-configs}
\end{table}

\begin{table*}[!t]
  \centering
  \setlength{\tabcolsep}{5pt}
  \renewcommand{\arraystretch}{1.1}
  \scalebox{0.7}{
      \begin{tabular}{lll}
        \toprule
        \textbf{Setting} & \textbf{Family} & \textbf{Models} \\
        \midrule
        \multirow{3}{*}{Closed-source}
          & Claude~\citep{claude45haiku,claude46sonnet,claude46opus}
          & Claude Haiku 4.5, Claude Sonnet 4.6, Claude Opus 4.6 \\
          & Gemini~\citep{gemini25flash,gemini31pro}
          & Gemini Flash 2.5, Gemini Pro 3.1 \\
          & GPT~\citep{gpt54}
          & GPT-5.4 nano, GPT-5.4 mini, GPT-5.4 \\
        \midrule
        \multirow{4}{*}{Open-source general}
          & GPT-OSS~\citep{gptoss120b}
          & GPT-OSS-20B, GPT-OSS-120B \\
          & DeepSeek~\citep{deepseekr1}
          & DeepSeek V3.2, DeepSeek R1 0528 \\
          & Qwen~\citep{yang2025qwen3}
          & qwen3-coder, Qwen3 235B \\
          & Llama~\citep{grattafiori2024llama}
          & Llama 3.1 8B, Llama 3.1 70B \\
        \midrule
        \multirow{2}{*}{Open-source Lean-specialized}
          & Kimina~\citep{wang2025kiminaprover}
          & Kimina 7B$\to$8B \\
          & Goedel~\citep{goedelproverv2}
          & Goedel 8B$\to$8B, Goedel 8B$\to$32B \\
        \bottomrule
      \end{tabular}
  }
  \caption{Non-agentic model configurations evaluated in our experiments.}
  \label{tab:models}
\end{table*}

We evaluate 19 non-agentic model configurations across 9 model families, as listed in \Cref{tab:models}.
Parameter counts for closed-source models are not publicly available.
General-purpose open-source models are evaluated directly, while Lean-specialized rows use the two-stage formalization-to-proving setup shown in \Cref{tab:models}.
The four agentic settings are Claude Code with Claude Opus~4.6, or Claude Opus 4.8, Claude Code with Qwen3~235B, and Codex with GPT-5.4.
Each setting is evaluated both with and without Numina-Lean-Agent, as listed in \Cref{tab:agentic-configs}.

\subsection{Per-Metric Rates}
\label{app:permetric}

\begin{table*}[!t]
  \centering
  \scriptsize
  \setlength{\tabcolsep}{3pt}
  \providecommand{\ShadowAvgsep}{\rule[-0.65ex]{0.45pt}{2.7ex}}
  \resizebox{\textwidth}{!}{%
  \begin{tabular}{lccccccccc@{\hspace{3pt}}c@{\hspace{3pt}}ccc}
    \toprule
    & \multicolumn{3}{c}{L1} & \multicolumn{3}{c}{L2} & \multicolumn{3}{c}{L3} & \ShadowAvgsep & \multicolumn{3}{c}{Average} \\
    & \multicolumn{3}{c}{(n{=}\edit{113})} & \multicolumn{3}{c}{(n{=}52)} & \multicolumn{3}{c}{(n{=}13)} & \ShadowAvgsep & \multicolumn{3}{c}{(n{=}\edit{178})} \\
    \cmidrule(lr){2-4}\cmidrule(lr){5-7}\cmidrule(lr){8-10}\cmidrule(lr){12-14}
    Model & BEq+ & BLEU & Judge & BEq+ & BLEU & Judge & BEq+ & BLEU & Judge & \ShadowAvgsep & BEq+ & BLEU & Judge \\
    \midrule
    \multicolumn{14}{c}{\textit{\textbf{Agentic methods}}} \\
    \midrule
    Claude Code (Opus 4.6) & \phantom{0}1.8 & \phantom{0}6.6 & 4.4 & \phantom{0}0.0 & \phantom{0}1.6 & 1.9 & \phantom{0}0.0 & \phantom{0}0.8 & 0.0 & \ShadowAvgsep & \phantom{0}1.1 & \phantom{0}4.7 & \edit{3.4} \\
    \quad + Numina & \edit{12.4} & \edit{18.9} & \edit{8.0} & 15.4 & 13.8 & 1.9 & \phantom{0}7.7 & \phantom{0}4.5 & 0.0 & \ShadowAvgsep & \edit{12.9} & 16.4 & \edit{5.6} \\
    Claude Code (Qwen3 235B) & \phantom{0}0.0 & \edit{11.6} & 0.0 & \phantom{0}0.0 & \phantom{0}7.5 & 0.0 & \phantom{0}0.0 & \phantom{0}1.2 & 0.0 & \ShadowAvgsep & \phantom{0}0.0 & \phantom{0}9.6 & 0.0 \\
    \quad + Numina & \phantom{0}0.0 & \edit{\phantom{0}9.6} & 0.0 & \phantom{0}0.0 & \phantom{0}7.4 & 0.0 & \phantom{0}0.0 & \phantom{0}2.8 & 0.0 & \ShadowAvgsep & \phantom{0}0.0 & \phantom{0}8.5 & 0.0 \\
    Codex (GPT-5.4) & \phantom{0}1.8 & \edit{12.4} & \edit{5.3} & \phantom{0}0.0 & 10.2 & 1.9 & \phantom{0}7.7 & \phantom{0}3.8 & 0.0 & \ShadowAvgsep & \phantom{0}1.7 & 11.1 & \edit{3.9} \\
    \quad + Numina & \edit{\phantom{0}9.7} & 15.6 & 5.3 & \phantom{0}7.7 & 11.8 & 1.9 & \phantom{0}7.7 & \phantom{0}2.7 & 0.0 & \ShadowAvgsep & \edit{\phantom{0}9.0} & \edit{13.5} & 3.9 \\
    \midrule
    \multicolumn{14}{c}{\textit{\textbf{Closed-source LLMs}}} \\
    \midrule
    GPT-5.4 & \edit{\phantom{0}2.7} & \edit{15.1} & \edit{1.8} & \phantom{0}0.0 & 12.7 & 0.0 & \phantom{0}7.7 & 12.8 & 0.0 & \ShadowAvgsep & \edit{\phantom{0}2.2} & 14.2 & 1.1 \\
    GPT-5.4 mini & \edit{\phantom{0}2.7} & 11.8 & 0.0 & \phantom{0}0.0 & 11.5 & 1.9 & \phantom{0}7.7 & 12.5 & 0.0 & \ShadowAvgsep & \edit{\phantom{0}2.2} & 11.8 & \edit{0.6} \\
    GPT-5.4 nano & \phantom{0}0.9 & 10.5 & 0.9 & \phantom{0}0.0 & \phantom{0}8.3 & 0.0 & \phantom{0}0.0 & \phantom{0}6.6 & 0.0 & \ShadowAvgsep & \edit{\phantom{0}0.6} & \phantom{0}9.6 & \edit{0.6} \\
    Gemini 3.1 Pro & \phantom{0}4.4 & 13.8 & \edit{5.3} & \phantom{0}3.8 & 11.7 & 0.0 & \phantom{0}7.7 & 14.4 & 0.0 & \ShadowAvgsep & \edit{\phantom{0}4.5} & 13.2 & \edit{3.4} \\
    Gemini 2.5 Flash & \phantom{0}3.5 & \edit{17.7} & 1.8 & \phantom{0}5.8 & 13.8 & 0.0 & 15.4 & 13.4 & 0.0 & \ShadowAvgsep & \phantom{0}5.1 & \edit{16.2} & 1.1 \\
    Claude Opus 4.6 & \edit{\phantom{0}6.2} & \edit{19.8} & \edit{1.8} & \phantom{0}1.9 & 18.0 & 1.9 & \phantom{0}7.7 & 14.7 & 0.0 & \ShadowAvgsep & \phantom{0}5.1 & \edit{18.9} & 1.7 \\
    Claude Sonnet 4.6 & \phantom{0}5.3 & \edit{16.8} & \edit{2.7} & \phantom{0}3.8 & 13.8 & 0.0 & 15.4 & 12.7 & 0.0 & \ShadowAvgsep & \phantom{0}5.6 & \edit{15.6} & 1.7 \\
    Claude Haiku 4.5 & \edit{\phantom{0}2.7} & 16.1 & 0.0 & \phantom{0}1.9 & 14.0 & 0.0 & 15.4 & 12.6 & 0.0 & \ShadowAvgsep & \edit{\phantom{0}3.4} & 15.2 & 0.0 \\
    \midrule
    \multicolumn{14}{c}{\textit{\textbf{Open-source LLMs}}} \\
    \midrule
    GPT-OSS-120B & \phantom{0}1.8 & \edit{13.9} & 2.7 & \phantom{0}1.9 & 13.1 & 0.0 & \phantom{0}7.7 & 14.2 & 0.0 & \ShadowAvgsep & \edit{\phantom{0}2.2} & \edit{13.7} & 1.7 \\
    GPT-OSS-20B & \phantom{0}0.9 & \edit{\phantom{0}5.7} & 2.7 & \phantom{0}1.9 & \phantom{0}6.7 & 0.0 & \phantom{0}0.0 & \phantom{0}8.2 & 0.0 & \ShadowAvgsep & \phantom{0}1.1 & \edit{\phantom{0}6.2} & 1.7 \\
    DeepSeek R1 0528 & \phantom{0}0.0 & \phantom{0}3.8 & 0.9 & \phantom{0}0.0 & \phantom{0}2.8 & 0.0 & \phantom{0}0.0 & \phantom{0}2.9 & 0.0 & \ShadowAvgsep & \phantom{0}0.0 & \phantom{0}3.5 & \edit{0.6} \\
    DeepSeek V3.2 & \phantom{0}3.5 & \edit{16.2} & 0.0 & \phantom{0}0.0 & 11.9 & 0.0 & 15.4 & 11.6 & 0.0 & \ShadowAvgsep & \edit{\phantom{0}3.4} & \edit{14.6} & 0.0 \\
    Qwen3 235B & \phantom{0}$-$ & 15.3 & $-$ & \phantom{0}$-$ & \phantom{0}8.7 & $-$ & \phantom{0}$-$ & \phantom{0}2.4 & $-$ & \ShadowAvgsep & \phantom{0}$-$ & \edit{12.4} & $-$ \\
    qwen3-coder & \phantom{0}0.9 & \edit{16.4} & 0.0 & \phantom{0}1.9 & 10.8 & 0.0 & 15.4 & \phantom{0}7.0 & 0.0 & \ShadowAvgsep & \edit{\phantom{0}2.2} & 14.1 & 0.0 \\
    llama3.1 70b & \phantom{0}0.0 & \edit{10.2} & 0.0 & \phantom{0}1.9 & \phantom{0}8.4 & 0.0 & \phantom{0}7.7 & \phantom{0}6.6 & 0.0 & \ShadowAvgsep & \phantom{0}1.1 & \phantom{0}9.4 & 0.0 \\
    llama3.1 8b & \phantom{0}0.0 & \phantom{0}7.6 & 0.0 & \phantom{0}0.0 & \phantom{0}6.1 & 0.0 & \phantom{0}7.7 & \phantom{0}4.9 & 0.0 & \ShadowAvgsep & \edit{\phantom{0}0.6} & \edit{\phantom{0}7.0} & 0.0 \\
    \midrule
    \multicolumn{14}{c}{\textit{\textbf{Lean-specialized methods}}} \\
    \midrule
    Kimina 7B$\to$8B & \phantom{0}0.0 & \phantom{0}5.9 & 0.0 & \phantom{0}0.0 & \phantom{0}5.4 & 0.0 & \phantom{0}0.0 & \phantom{0}9.6 & 0.0 & \ShadowAvgsep & \phantom{0}0.0 & \phantom{0}6.0 & 0.0 \\
    Goedel 8B$\to$32B & \phantom{0}0.0 & \phantom{0}6.6 & 0.0 & \phantom{0}0.0 & \phantom{0}6.0 & 0.0 & \phantom{0}0.0 & \phantom{0}1.7 & 0.0 & \ShadowAvgsep & \phantom{0}0.0 & \phantom{0}6.1 & 0.0 \\
    Goedel 8B$\to$8B & \phantom{0}0.0 & \phantom{0}6.1 & 0.0 & \phantom{0}1.9 & \phantom{0}8.9 & 0.0 & \phantom{0}7.7 & 14.6 & 0.0 & \ShadowAvgsep & \phantom{0}1.1 & \phantom{0}7.5 & 0.0 \\
    \bottomrule
  \end{tabular}}
  \caption{Per-difficulty automatic metric results (\%) on \ours. BEq+ and Judge are pass rates. BLEU is a token-level similarity score. The Average columns are task-weighted over 178 problems.}
  \label{tab:per-metric-by-difficulty}
\end{table*}

This section expands \Cref{sec:metric-correlation}.
\Cref{tab:per-metric-by-difficulty} reports BEq+, BLEU, and LLM-as-judge by difficulty.
The judge prompt is in \Cref{app:judge-prompt} and the cost profile in \Cref{app:judge-cost}.





\section{Cost and Compute}
\label{sec:appendix:cost_compute}
\label{app:compute-resources}
\label{app:cost}

\begin{table}[h!]
  \centering
  \footnotesize
  \begin{tabular}{lrrr}
    \toprule
    & \multicolumn{2}{c}{\code{max\_tokens}} & \\
    \cmidrule(lr){2-3}
    Bucket & 8192 & 16384 & Subtotal \\
    \midrule
    Closed (API)        & \$23.10 & \$30.46 & \$53.56 \\
    Open (via OpenRouter) & \$3.79  & \$4.70  & \$8.49 \\
    Open (self-hosted)  & \$0.00  & \$0.00  & \$0.00 \\
    \midrule
    \textbf{Total} & \$26.89 & \$35.16 & \$62.05 \\
    \bottomrule
  \end{tabular}
  \caption{Non-agentic generation cost across 33 runs and 178 prompts per run, taken from per-call sidecar logs.
  }
  \label{tab:cost-na}
\end{table}

This section reports the compute and API budget used for the experiments.
Measured costs come from per-call sidecar logs with \code{prompt\_tokens},
\code{completion\_tokens}, and OpenRouter-quoted \code{cost\_usd}.
Estimated costs use observed prompt and completion lengths with provider
pricing.
Self-hosted Lean-specialized models and Qwen3~235B were served on local GPUs,
so we report wall-clock time rather than cost.

\subsection{Non-Agentic Generation}

\Cref{tab:cost-na} aggregates per-call sidecar logs across the 33 non-agentic
runs.
Closed-source rows are billed by the model provider.
Open-source API rows are billed through OpenRouter.
Open-source self-hosted rows are local vLLM deployments and have zero API cost.

\subsection{Agentic Generation}

\begin{table*}[!b]
  \centering
  \footnotesize
  \begin{tabular}{lrrrr}
    \toprule
    Experiment & N tasks & Wall-clock (h) & Avg/task (min) & Est. cost (USD) \\
    \midrule
    Claude Code (Opus 4.6)         & \edit{178} & 20.5 &  6.9 & \$310 \\
    \quad $+$ Numina              & \edit{178} & 101.2 & \edit{34.1} & \$1{,}500 \\
    Claude Code (Opus 4.8)         & \edit{178} & 20.4 &  \edit{6.9} & \$310 \\
    \quad $+$ Numina              & \edit{178} & 51.8 & \edit{17.5} & \$780 \\
    Codex (GPT-5.4)                & \edit{178} &  2.8 &  0.9 & \$200 \\
    \quad $+$ Numina              & \edit{178} & 21.1 &  7.1 & \$330 \\
    Claude Code (Qwen3 235B)       & \edit{178} &  --  & --   & local GPU \\
    \quad $+$ Numina               & \edit{178} &  --  & --   & local GPU \\
    \midrule
    \textbf{Total}     &     & 217.8 &       & \$3{,}430 \\
    \bottomrule
  \end{tabular}
  \caption{Estimated API cost of agentic methods over the 178 problems of \ours.}
  \label{tab:cost-agentic}
\end{table*}

The agentic runs use Claude Code and Codex, each with and without Numina-Lean-Agent.
Early experiments used subscription access, while later runs incurred paid API charges.
For consistent accounting, \Cref{tab:cost-agentic} reports the estimated cost if all API-capable agentic runs were billed through API.
We estimate cost from per-task wall-clock time and provider token throughput.
The table reports wall-clock time and estimated API cost for the \edit{178}-problem run.
Claude Code with Numina-Lean-Agent is the most costly API-billed setting at \edit{34.1} minutes per problem.
Claude Code with Qwen3~235B uses the local vLLM server, so it has no API charge in this table.

\subsection{LLM-as-Judge}
\label{app:judge-cost}
\begin{table}[t!]
  \centering
  \footnotesize
  \begin{tabular}{lrr}
    \toprule
    Model & Calls & Cost \\
    \midrule
    \small{anthropic/claude-opus-4.7}          & \edit{9{,}207} & \edit{\$324} \\
    \small{google/gemini-3.1-pro-preview}      & \edit{9{,}207} & \edit{\$58} \\
    \small{openai/gpt-5.4}                     & \edit{9{,}207} & \edit{\$98} \\
    \midrule
    \textbf{Total judge spend}                 & \edit{27{,}621} & \edit{\$480} \\
    \bottomrule
  \end{tabular}
  \caption{Cost of LLM-as-judge.}
  \label{tab:cost-judge}
\end{table}

We score every cached generation with a three-judge majority vote (3 samples per judge, temperature 0.5).
The three judges are \texttt{anthropic/claude-opus-4.7}, \texttt{google/gemini-3.1-pro-preview}, and \texttt{openai/gpt-5.4}, accessed via OpenRouter.
At three samples per judge, each candidate triggers nine completions.

\subsection{Total}
Adding the three components above, the API spend for the experiments reported in the main body is approximately \edit{\$3{,}970} (\$62 NA $+$ \$3{,}430 agentic $+$ \edit{\$480} judge).
Self-hosted local GPU runs are not included in this API total.

\section{Per-Configuration Agreement with Expert Judgment}
\label{sec:appendix:agreement}

\begin{table}[!t]
  \centering
  \footnotesize
  \setlength{\tabcolsep}{4pt}
  \begin{tabular}{lcccc}
    \toprule
    Metric & Precision & Recall & F1 & Agreement \\
    \midrule
    Compile      & \edit{0.214} & \edit{\textbf{1.000}} & \edit{0.353} & \edit{0.214} \\
    BLEU         & \edit{0.265} & \edit{0.722} & \edit{0.388} & \edit{0.512} \\
    LLM-as-judge & \edit{0.375} & \edit{0.167} & \edit{0.231} & \edit{\underline{0.762}} \\
    BEq+         & \edit{0.364} & \edit{0.222} & \edit{0.276} & \edit{0.750} \\
    \ourrelaxed  & \edit{\underline{0.400}} & \edit{\underline{0.889}} & \edit{\underline{0.552}} & \edit{0.690} \\
    \ourmetric   & \edit{\textbf{1.000}} & \edit{\underline{0.889}} & \edit{\textbf{0.941}} & \edit{\textbf{0.976}} \\
    \bottomrule
  \end{tabular}
  \caption{\edit{Binary agreement of automatic metrics with expert judgment for outputs
  from Claude Code (Opus 4.6) with Numina-Lean-Agent on \ours.}
  The best value per column is bold, and the second best is underlined.}
  \label{tab:agree-claude-numina}
\end{table}

\begin{table}[!t]
  \centering
  \footnotesize
  \setlength{\tabcolsep}{4pt}
  \begin{tabular}{lcccc}
    \toprule
    Metric & Precision & Recall & F1 & Agreement \\
    \midrule
    Compile      & \edit{0.000} & \edit{--} & \edit{--} & \edit{0.000} \\
    BLEU         & \edit{0.000} & \edit{--} & \edit{--} & \edit{0.250} \\
    LLM-as-judge & \edit{0.000} & \edit{--} & \edit{--} & \edit{\underline{0.750}} \\
    BEq+         & \edit{--} & \edit{--} & \edit{--} & \edit{\textbf{1.000}} \\
    \ourrelaxed  & \edit{0.000} & \edit{--} & \edit{--} & \edit{0.375} \\
    \ourmetric   & \edit{--} & \edit{--} & \edit{--} & \textbf{1.000} \\
    \bottomrule
  \end{tabular}
  \caption{Binary agreement of automatic metrics with expert judgment for outputs from
  Claude Code (Opus 4.6) on \ours.
  The best value per column is bold, and the second best is underlined.}
  \label{tab:agree-claude-direct}
\end{table}

\begin{table}[!t]
  \centering
  \footnotesize
  \setlength{\tabcolsep}{4pt}
  \begin{tabular}{lcccc}
    \toprule
    Metric & Precision & Recall & F1 & Agreement \\
    \midrule
    Compile      & \edit{0.192} & \textbf{1.000} & \edit{0.322} & \edit{0.192} \\
    BLEU         & \edit{0.286} & \edit{0.842} & \edit{0.427} & \edit{0.566} \\
    LLM-as-judge & \edit{0.143} & \edit{0.053} & \edit{0.077} & \edit{0.758} \\
    BEq+         & \edit{0.444} & \edit{0.211} & \edit{0.286} & \edit{\underline{0.798}} \\
    \ourrelaxed  & \edit{\underline{0.452}} & \textbf{1.000} & \edit{\underline{0.623}} & \edit{0.768} \\
    \ourmetric   & \textbf{1.000} & \edit{\underline{0.947}} & \edit{\textbf{0.973}} & \edit{\textbf{0.990}} \\
    \bottomrule
  \end{tabular}
  \caption{Binary agreement of automatic metrics with expert judgment for outputs from
  Codex (GPT-5.4) with Numina-Lean-Agent on \ours.
  The best value per column is bold, and the second best is underlined.}
  \label{tab:agree-codex-numina}
\end{table}

\begin{table}[!t]
  \centering
  \footnotesize
  \setlength{\tabcolsep}{4pt}
  \begin{tabular}{lcccc}
    \toprule
    Metric & Precision & Recall & F1 & Agreement \\
    \midrule
    Compile      & \edit{0.133} & \edit{\textbf{1.000}} & \edit{0.235} & \edit{0.133} \\
    BLEU         & \edit{0.000} & \edit{0.000} & \edit{0.000} & \edit{0.567} \\
    LLM-as-judge & \edit{--} & \edit{0.000} & -- & \edit{\underline{0.867}} \\
    BEq+         & \edit{--} & \edit{0.000} & -- & \edit{\underline{0.867}} \\
    \ourrelaxed  & \edit{\underline{0.500}} & \edit{\textbf{1.000}} & \edit{\underline{0.667}} & \edit{\underline{0.867}} \\
    \ourmetric   & \textbf{1.000} & \edit{\textbf{1.000}} & \edit{\textbf{1.000}} & \edit{\textbf{1.000}} \\
    \bottomrule
  \end{tabular}
  \caption{Binary agreement of automatic metrics with expert judgment for outputs from
  Codex (GPT-5.4) on \ours.
  The best value per column is bold, and the second best is underlined.}
  \label{tab:agree-codex-direct}
\end{table}

\begin{table}[!t]
  \centering
  \footnotesize
  \setlength{\tabcolsep}{4pt}
  \begin{tabular}{lcccc}
    \toprule
    Metric & Precision & Recall & F1 & Agreement \\
    \midrule
    Compile      & 0.083 & \textbf{1.000} & 0.154 & 0.083 \\
    BLEU         & 0.000 & 0.000 & 0.000 & \edit{0.583} \\
    LLM-as-judge & \edit{--} & 0.000 & \edit{--} & \underline{0.917} \\
    BEq+         & \edit{--} & 0.000 & \edit{--} & \underline{0.917} \\
    \ourrelaxed  & \underline{0.500} & \textbf{1.000} & \underline{0.667} & \underline{0.917} \\
    \ourmetric   & \textbf{1.000} & \textbf{1.000} & \textbf{1.000} & \textbf{1.000} \\
    \bottomrule
  \end{tabular}
  \caption{Binary agreement of automatic metrics with expert judgment for outputs from
  Claude Code (Qwen3 235B) with Numina-Lean-Agent on \ours.
  The best value per column is bold, and the second best is underlined.}
  \label{tab:agree-qwen-numina}
\end{table}

\begin{table}[!t]
  \centering
  \footnotesize
  \setlength{\tabcolsep}{4pt}
  \begin{tabular}{lcccc}
    \toprule
    Metric & Precision & Recall & F1 & Agreement \\
    \midrule
    Compile      & \edit{0.111} & \edit{\textbf{1.000}} & \edit{0.200} & \edit{0.111} \\
    BLEU         & \edit{--} & \edit{0.000} & -- & \edit{\underline{0.889}} \\
    LLM-as-judge & \edit{--} & \edit{0.000} & -- & \edit{\underline{0.889}} \\
    BEq+         & \edit{--} & \edit{0.000} & -- & \edit{\underline{0.889}} \\
    \ourrelaxed  & \edit{\underline{0.500}} & \edit{\textbf{1.000}} & \edit{\underline{0.667}} & \edit{\underline{0.889}} \\
    \ourmetric   & \textbf{1.000} & \edit{\textbf{1.000}} & \edit{\textbf{1.000}} & \edit{\textbf{1.000}} \\
    \bottomrule
  \end{tabular}
  \caption{Binary agreement of automatic metrics with expert judgment for outputs from
  Claude Code (Qwen3 235B) on \ours.
  The best value per column is bold, and the second best is underlined.}
  \label{tab:agree-qwen-direct}
\end{table}

For expert judgment, two mathematics experts independently evaluate whether each generated theorem is semantically aligned with the source problem. 
The evaluation is finalized once both annotators reach consensus. 
The disagreements are resolved through discussion.

\Cref{tab:human-correlation} in the main text pools all six agentic
configurations. 
\Cref{tab:agree-claude-numina,tab:agree-claude-direct,tab:agree-codex-direct,tab:agree-codex-numina,tab:agree-qwen-direct,tab:agree-qwen-numina} report them individually.
\edit{Across every configuration, \ourmetric achieves a Precision of $1.0$ against expert judgment, with Recall of
$1.0$ except on two of the three Numina configurations, where it is $0.889$ and $0.947$.
Claude Code (Opus 4.6) without Numina-Lean-Agent produced only incorrect outputs, so Recall and F1 are undefined for it and Agreement is reported instead.}

\paragraph{False Negatives of \ourmetric.}
\label{sec:appendix:false-negatives}

\ourmetric misses three of the outputs the experts judged aligned.
All three state the intended theorem correctly but name their auxiliary
declarations differently from the reference.
Two rename or omit them, 
and the other one declares the expected name as a theorem where the reference declares a definition.
The LLM matching step of checker adaptation (\Cref{sec:curation}) did not
resolve these cases, so every forward check for those problems fails.

\section{Threshold Sensitivity}
\label{sec:appendix:threshold}

\begin{figure}[!t]
  \centering
  \includegraphics[width=\columnwidth]{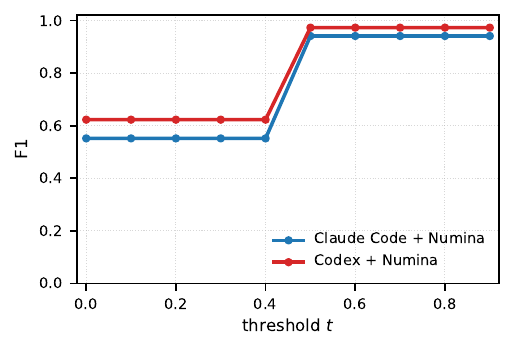}
  \caption{F1 of \ourrelaxed against expert judgment across the binarization threshold $t$.}
  \label{fig:sa-pass-threshold}
\end{figure}

The relaxed score \ourrelaxed is the fraction of passing shadow checks, and is thus non-binary.
To test whether the choice of threshold affects agreement with expert judgment, 
we sweep a threshold $t$ and count a statement as positive when its passing fraction exceeds $t$ (\Cref{fig:sa-pass-threshold}).
\edit{F1 is flat below $t=0.5$, at $0.552$ for Claude Code (Opus 4.6) with
Numina-Lean-Agent and $0.623$ for Codex with Numina-Lean-Agent, and rises to $0.941$ and
$0.973$ at $t=0.5$. A passing fraction above one half requires both directions to pass, so
\ourrelaxed coincides with \ourmetric from that point on.}
We therefore report \ourrelaxed with $t=0$, the setting most distinct from \ourmetric.

\begin{figure}[!t]
  \centering
  \includegraphics[width=\columnwidth]{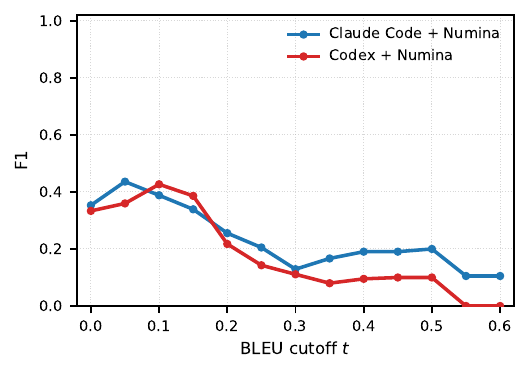}
  \caption{F1 of BLEU against expert judgment across the cutoff $t$.}
  \label{fig:bleu-threshold}
\end{figure}

\edit{For BLEU, F1 peaks at a low cutoff and then falls (\Cref{fig:bleu-threshold}). Claude Code
(Opus 4.6) with Numina-Lean-Agent peaks at $0.436$ with a cutoff of $0.05$, and Codex with
Numina-Lean-Agent at $0.427$ with a cutoff of $0.1$. We report BLEU at a cutoff of $0.1$, which
maximizes the mean F1 over the two configurations that have expert-labelled positives, so BLEU
is scored at its most favorable setting. No cutoff brings BLEU near the agreement \ourmetric
reaches.}

\section{Rank Correlation with Expert Judgment}
\label{sec:appendix:rank_correlation}

\begin{table}[h]
  \centering
  \footnotesize
  \setlength{\tabcolsep}{5pt}
  \begin{tabular}{lcc}
    \toprule
    Metric & $\rho$ ($\uparrow$) & $\tau$ ($\uparrow$) \\
    \midrule
    Compile & \edit{0.99} & \edit{0.97} \\
    BLEU & \edit{0.90} & \edit{0.79} \\
    LLM-as-judge & \edit{0.72} & \edit{0.52} \\
    BEq+ & \edit{0.76} & \edit{0.57} \\
    \ourrelaxed & \edit{0.82} & \edit{0.71} \\
    \ourmetric & \textbf{1.00} & \textbf{1.00} \\
    \bottomrule
  \end{tabular}
  \caption{
  Spearman $\rho$ and Kendall $\tau$ 
  between the ranking of the six agentic systems by each automatic metric and by expert judgment, over the 178 problems. 
  The best value per column is in bold.}
  \label{tab:appendix-rankcorr-system}
\end{table}

We measure whether each automatic metric ranks whole systems in the same order as a human expert.
Two human experts label every candidate Lean statement as \texttt{ok} or \texttt{reject}. 
We evaluate
six agentic systems: Claude Code and Codex, each run directly and with Numina-Lean-Agent, and Qwen3
235B run directly and with Numina-Lean-Agent. For each system, we aggregate every score over all \edit{178}
tasks and rank the systems by the expert pass rate.
\Cref{tab:appendix-rankcorr-system} reports Spearman $\rho$ and Kendall $\tau$ between the ranking
of the six systems induced by each metric and the expert ranking.

\ourmetric recovers the expert ranking exactly ($\rho=\tau=1.00$), and \edit{compile rate is next at
$\rho=0.99$, ahead of BLEU ($0.90$), \ourrelaxed ($0.82$), BEq+ ($0.76$), and LLM-as-judge ($0.72$).
With six systems whose compile rates range from $4.5\%$ to $55.6\%$, the system-level ranking is a
coarse test, and the per-task agreement in \Cref{tab:human-correlation} separates the metrics
further.}

\section{Public Challenge Track}
\label{sec:public-challenge}
\label{sec:appendix:challenge}

\begin{figure*}[t]
  \centering
  \includegraphics[width=2\columnwidth]{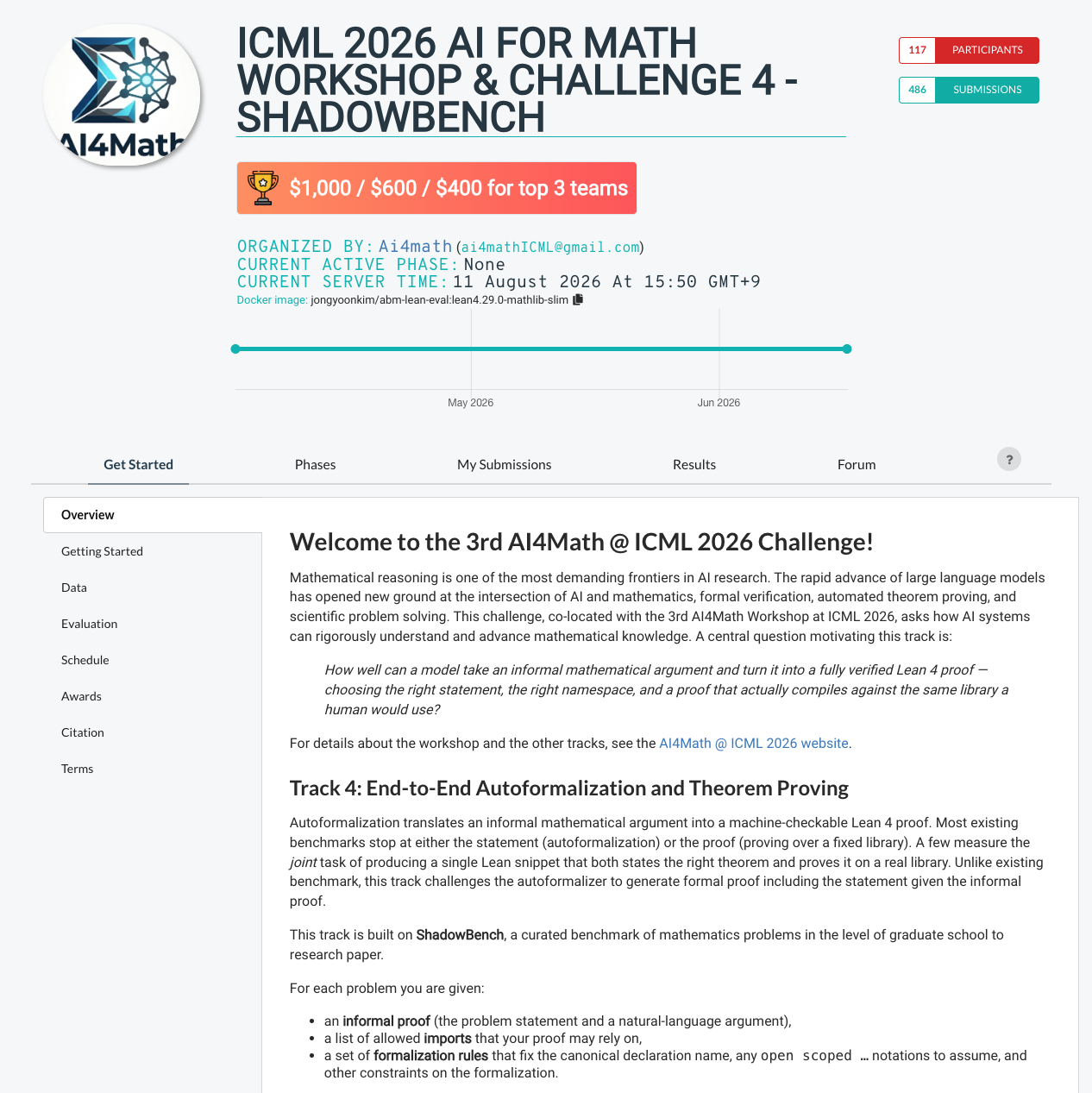}
  \caption{CodaBench page of \ours\ as Track 4 of the ICML 2026 AI4Math
  Challenge.}
  \label{fig:codabench-page}
\end{figure*}

We released an early 126-problem version of \ours as Track 4 of the ICML 2026 AI4Math Challenge.
\Cref{fig:codabench-page} shows the CodaBench competition page that hosted the 126-problem snapshot of \ours as Track 4 of the ICML 2026 AI4Math Challenge.
As we described in \Cref{sec:release}, informal theorems and formalization rules are accessible to the public, while the checker theorems for evaluation remain hidden.
About 117 teams registered, but only twenty-seven teams actually participated.

\begin{figure}[t]
  \centering
  \begin{subfigure}[b]{\columnwidth}
    \centering
    \includegraphics[width=\columnwidth]{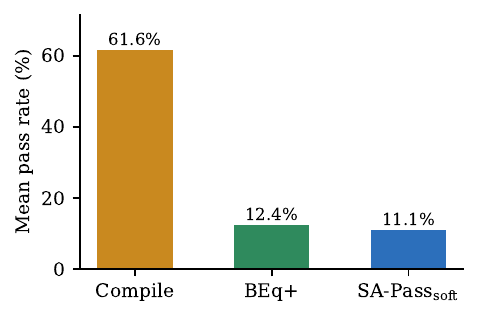}
    \caption{Mean pass rate under compilation, BEq+, and \ourrelaxed.}
    \label{fig:challenge-funnel-rates}
  \end{subfigure}

  \vspace{4pt}

  \begin{subfigure}[b]{\columnwidth}
    \centering
    \includegraphics[width=\columnwidth]{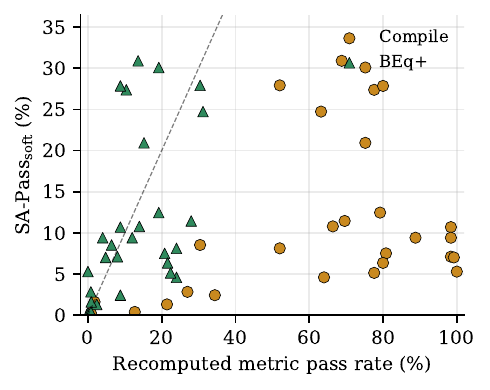}
    \caption{Per participant, the recomputed metric pass rate against \ourrelaxed.}
    \label{fig:challenge-funnel-scatter}
  \end{subfigure}
  \caption{
  Metric disagreement on the public challenge.
  In (\subref{fig:challenge-funnel-rates}), compilation accepts $5.5\times$ as many outputs on average compared to \ourrelaxed.
  In (\subref{fig:challenge-funnel-scatter}), points for compilation lie far
  from the diagonal, combining a high compile rate with a low \ourrelaxed
  score, which indicates over-acceptance.
  }
  \label{fig:challenge-funnel}
\end{figure}

The participants' results show similar trends to our experiments.
The \ourrelaxed distribution is skewed, where the median is $8.1\%$, and the average is $11.1\%$, while the compile rate averages $61.6\%$.
That is, there is about a $5.5\times$ gap between the compile rate and \ourrelaxed, underlining that the compile rate causes false positives in formalizations.
In many of these low-scoring cases, the submission replaces the target with a trivial statement closed by \code{trivial}, which compiles but cannot
be used by another theorem or declaration.
Such submissions give the compile rate many false positives.
They can even induce a ranking error, where the top five by compile rate share no member with the top five by \ourrelaxed.
Compilation and BEq+ 
produce rankings that differ substantially from \ourrelaxed
(\Cref{fig:challenge-funnel}), with the compile rate agreeing at only $\rho=0.41$ and BEq+ passing $12.4\%$.
This result over many independent participants is consistent with the metric-validity gap we measure against expert judgment in \Cref{sec:metric-correlation}.

The best submissions use agentic pipelines that decompose each goal into auxiliary lemmas, \
and we observe the same trend in our experiments, 
where Claude Code (Opus 4.8) with Numina-Lean-Agent achieves the best result.
We examine two factors, agentic behavior and decomposition.
We label a submission as agentic when it runs more than three turns, and we measure decomposition by the number of declarations, such as lemma, def,
or theorem.
The verified score correlates with the number of declarations at Spearman's $\rho$ of $0.59$ and with the number of turns at $0.56$, emphasizing
that \ours requires multi-turn solving through decomposition.
$59.7\%$ of submissions add at least one auxiliary lemma or definition ($3.6$ on average), 
which is close to the $4.5$ in the reference proofs (\Cref{tab:dataset-statistics}).
\Cref{tab:challenge-leaderboard} lists the top ten participants and their submission profiles.

\begin{table*}[t]
  \centering
  \footnotesize
  \setlength{\tabcolsep}{5pt}
  \begin{tabular}{c r r r r r r}
    \toprule
    Rank & \ourrelaxed \% & Compile \% & No Sorry \% & Length (Char) & Auxiliary & Multiple declaration \% \\
    \midrule
    1  & 30.9 & 68.8 & 99.2  & 3,045 & 5.38 & 92.0 \\
    2  & 30.1 & 75.2 & 99.2  & 2,952 & 5.42 & 90.4 \\
    3  & 27.9 & 52.0 & 96.8  & 5,133 & 9.58 & 58.2 \\
    4  & 27.8 & 80.0 & 99.2  & 2,365 & 4.40 & 75.2 \\
    5  & 27.4 & 77.6 & 99.2  & 2,899 & 5.71 & 88.0 \\
    6  & 24.7 & 63.2 & 95.2  & 4,113 & 6.56 & 70.8 \\
    7  & 20.9 & 75.2 & 94.4  & 969   & 2.43 & 68.9 \\
    8  & 12.5 & 79.2 & 100.0 & 2,567 & 2.69 & 47.6 \\
    9  & 11.4 & 69.6 & 100.0 & 2,434 & 3.53 & 45.2 \\
    10 & 10.8 & 66.4 & 91.1  & 8,052 & 6.89 & 78.8 \\
    \bottomrule
  \end{tabular}
  \caption{
  Top ten teams on the \ours public challenge track, anonymized and ranked by \ourrelaxed.
  Compile is the fraction of the 126 problems whose submitted solution compiles standalone.
  No Sorry counts the number of problems solved without sorry.
  Length indicates the average length of submission in characters.
  Auxiliary is the average number of auxiliary declaration per each problem solution beyond single canonical solution.
  Multiple declaration counts the number of solutions that used more than one declaration.
  The median \ourrelaxed across all 27 participants is $8.1\%$.}
  \label{tab:challenge-leaderboard}
\end{table*}

\section{Annotation Interface}
\label{sec:appendix:annotation_interface}

\begin{figure*}[t!]
  \centering
  \includegraphics[width=\linewidth]{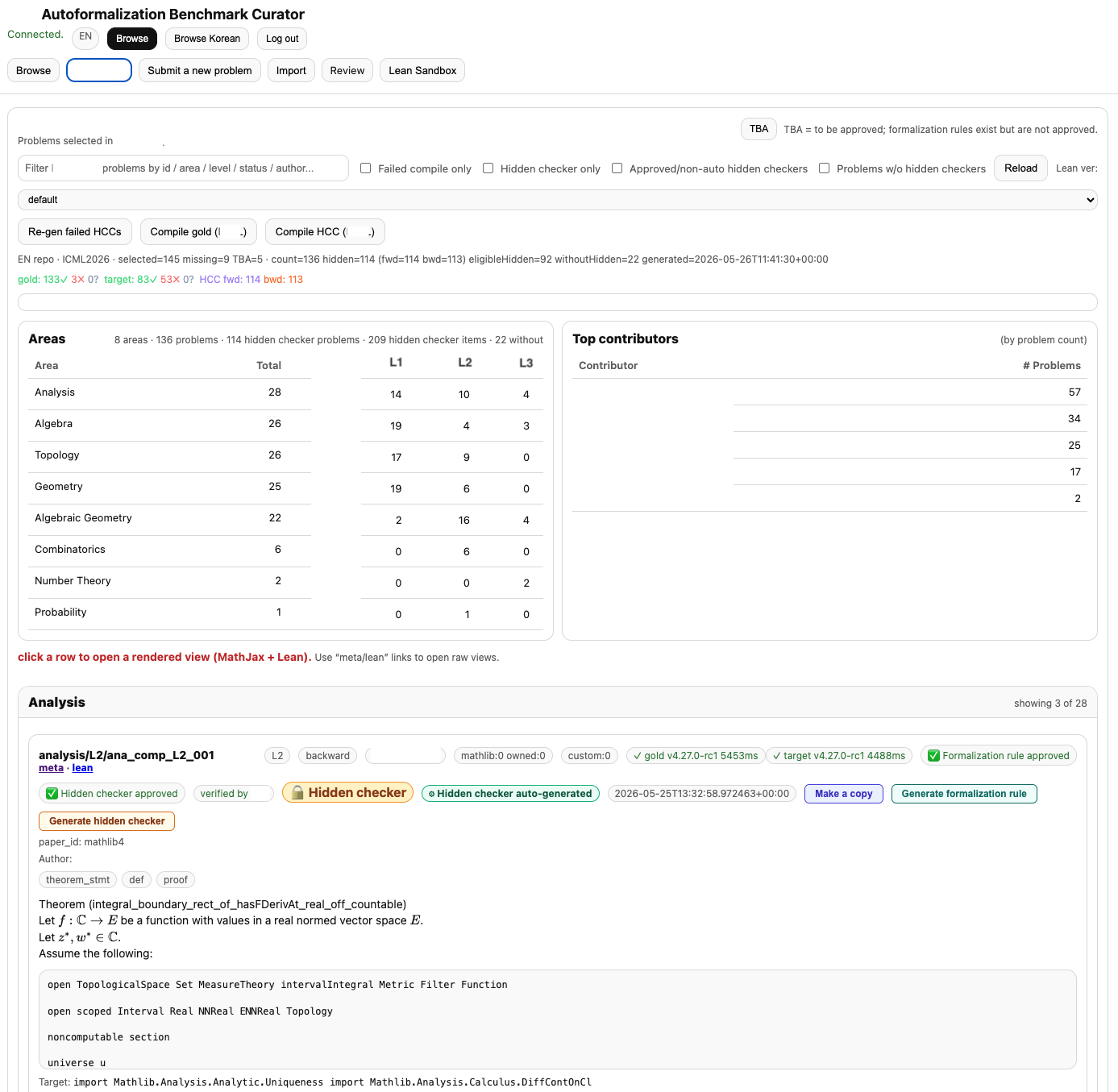}
  \caption{Data annotation landing page. The left panel summarises
    area by level coverage. The right panel ranks
    contributors by the number of problems they have annotated.
    Selecting a row in the left panel opens the problem detail in
    \Cref{fig:annotation-example}.}
  \label{fig:annotation-main}
\end{figure*}

\begin{figure*}[t!]
  \centering
  \includegraphics[width=\linewidth]{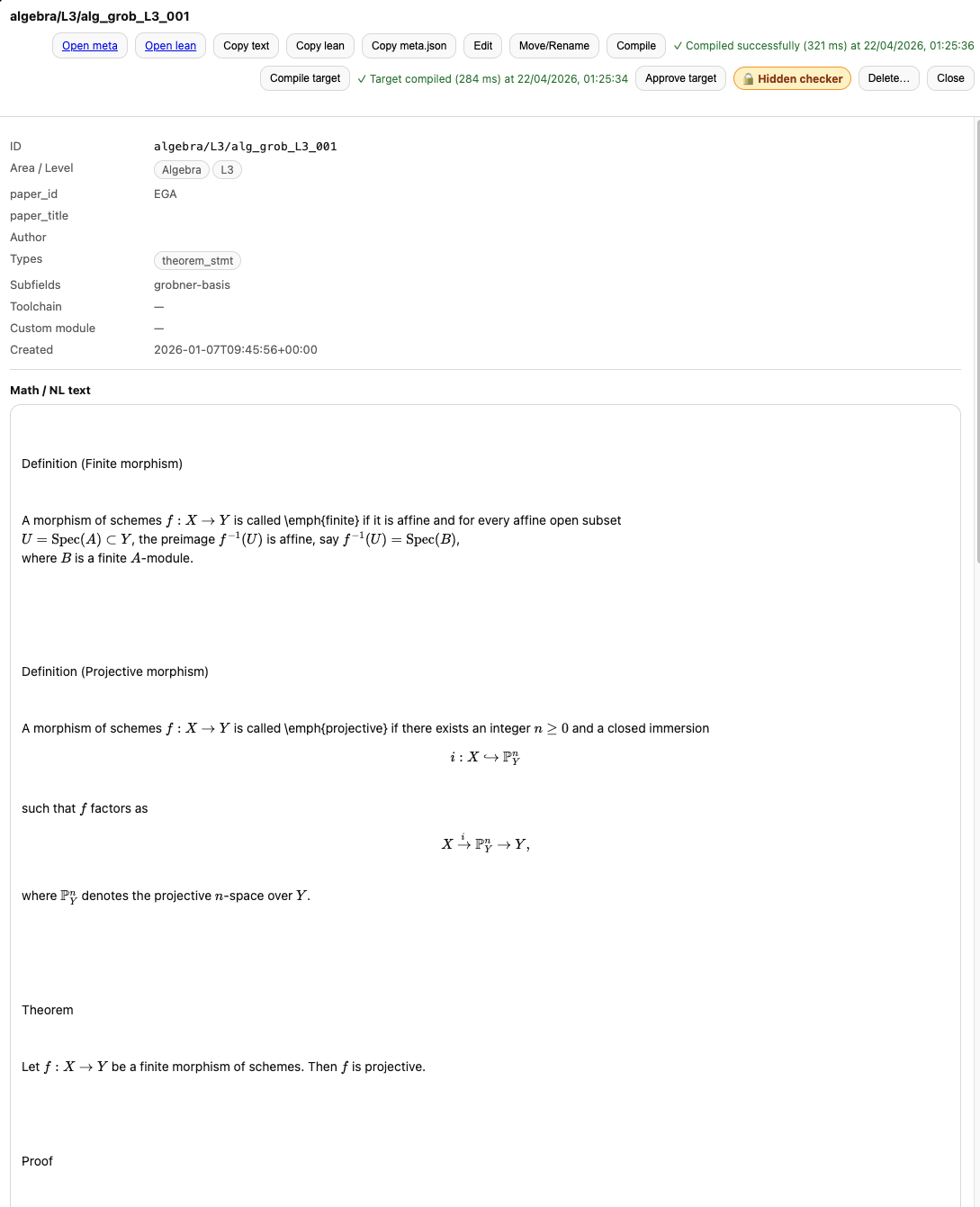}
  \caption{Problem detail view (\texttt{algebra/L3/alg\_grob\_L3\_001}).
    Header buttons surface the underlying \texttt{meta.json} and
    \texttt{gold.lean}, and the ``Compile'' / ``Compile target''
    actions rerun the Lean type checker through the evaluation
    Docker image. The body renders the LaTeX natural language
    statement and proof.}
  \label{fig:annotation-example}
\end{figure*}

\begin{figure*}[t!]
  \centering
  \includegraphics[width=1\linewidth]{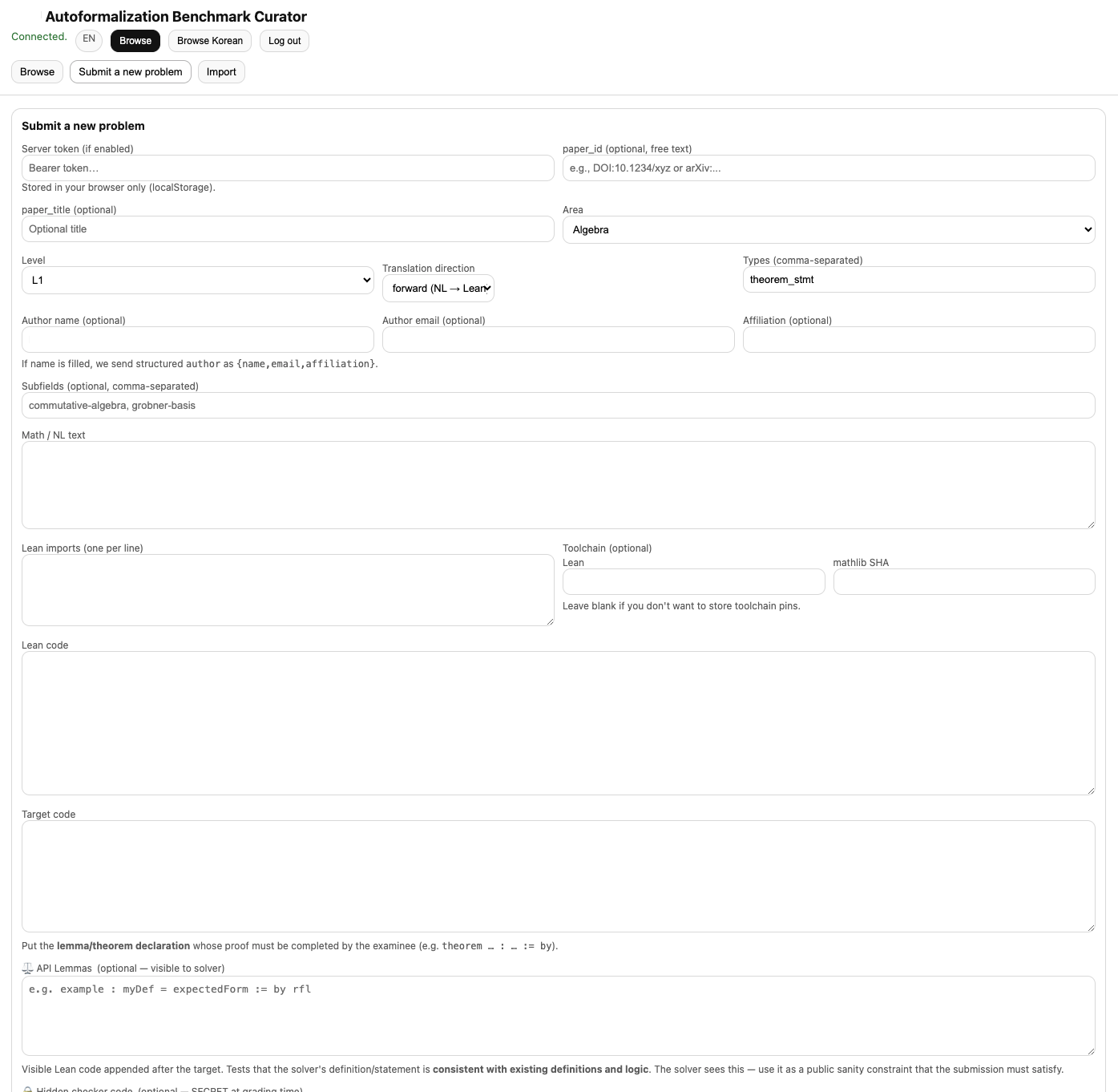}
  \caption{Submission form for a new problem. The form pairs
    natural language text (\texttt{Math / NL text}) with Lean code,
    accepts optional auxiliary lemma imports, and separates the public
    target code from an optional hidden checker block used for semantic
    evaluation.}
  \label{fig:annotation-write}
\end{figure*}

The \edit{178} problems were annotated through a custom browser-based tool.
The tool exposes three views, each captured below.
\Cref{fig:annotation-main} is the main page, which shows the number of problems
annotated for each area and level.
Selecting a problem opens the detail page in \Cref{fig:annotation-example}.
The detail view renders the natural language statement and proof, and exposes
metadata such as area, level, paper of origin, subfields, and type tags.
This page also shows whether the problem is valid by compiling with the reference Lean code through the same Docker image used at evaluation time.
To submit the new problem, curators fill out the form in \Cref{fig:annotation-write}.
This page accepts paired natural language text and Lean code, optional auxiliary imports,
reference code that the examinee must complete,
and hidden shadow checks used to evaluate semantic correctness.

\section{Prompt Templates}
\label{app:prompts}

Every prompt used in this paper is reproduced below verbatim, one per
figure. Placeholders of the form \texttt{\{\{ \ldots{} \}\}} are
filled per problem by the serving pipeline. Raw text sources live
under \texttt{prompts/} and \texttt{scripts/judge\_prompt.py} in the
code repository. The structured environments here mirror the
\texttt{system}, \texttt{user}, and \texttt{model} roles in the API
call.

\paragraph{Non-agentic prompt.}
\label{app:prompt-naive}
We use the prompt shown in \Cref{fig:prompt-autoformalize} for non-agentic models.

\paragraph{Agentic baseline.}
\label{app:prompt-claude-code}
\Cref{fig:prompt-claude-code} shows the direct agentic prompt used
without Numina-Lean-Agent. The Claude Code rows invoke this prompt
through the standard Claude Code CLI on the target Lean file. Codex
uses the same task instruction in its own wrapper.

\paragraph{Agentic setting (Numina-Lean-Agent).}
\label{app:prompt-agentic}
\Cref{fig:prompt-medium-mode} is used for the
\textit{+ Numina} rows of \Cref{tab:main-results}.
Numina-Lean-Agent runs the coordinator prompt below, which instructs the
agent to select one target \texttt{sorry} per session, sketch
incrementally, and verify via \texttt{lean\_diagnostic\_messages}.

\subsection{LLM-as-Judge Prompt}
\label{app:judge-prompt}

The judge prompt drives the Judge column in \Cref{tab:per-metric-by-difficulty}. For each candidate
the builder fills \texttt{\{\{ informal\_block \}\}} with the problem's
NL statement (and NL proof, when the ablation condition provides
one), \texttt{\{\{ formal\_ref \}\}} with the reference Lean~4
statement with \texttt{sorry}, and \texttt{\{\{ candidate \}\}} with
the model's generated Lean~4 theorem plus its proof.

The response parser accepts bare JSON, fenced JSON, and a permissive
regex fallback. Verdicts are normalised to
\{\texttt{correct}, \texttt{incorrect}, \texttt{uncertain}\}. All
three are recorded per vote, then reduced by majority.

\onecolumn
\begin{systemprompt}
    You are a helpful assistant that can interact multiple times with a computer shell to solve programming tasks.
    However, for this specific interaction, you are acting as an expert Lean 4 mathematician tasked with implementing a formal Lean 4 declaration (theorem or definition) based on the provided context.

    Your response must contain exactly ONE lean code block.

    Include a THOUGHT section before your code where you explain your reasoning process.
    Format your response as shown in \texttt{<format\_example>}.

    \texttt{<format\_example>}
    \begin{textcode}
THOUGHT: Your reasoning and analysis here, explaining how you deduce the implementation from the context.
    \end{textcode}

    \begin{leancode}
your_formal_lean_code_here
    \end{leancode}
    \texttt{</format\_example>}

    Failure to follow these rules will cause your response to be rejected.
\end{systemprompt}

\begin{userprompt}
    \texttt{<task\_description>}
    \\
    Your task is to provide the full formal Lean 4 code for the specific declaration named \texttt{\{\{ target\_code\_name \}\}}.
    The declaration is part of a larger code sequence, and you must implement it to fit seamlessly into the provided context.

    The Code Context is provided below.
    \begin{itemize}
        \item \texttt{Header}: Imports and open namespaces.
        \item \texttt{Code Before}: Code appearing before the target.
        \item \texttt{Target Code Name}: The name of the definition or theorem you need to implement.
        \item \texttt{Natural Language Description}: A description of the target code in natural language.
        \item \texttt{Code After}: Code appearing after the target.
    \end{itemize}

    Header:
    \begin{leancode}
{{ header }}
    \end{leancode}

    Code Before:
    \begin{leancode}
{{ before_target_code }}
    \end{leancode}

    Target Code Name:
    \begin{textcode}
{{target_code_name}}
    \end{textcode}

    Natural Language Description:
    \begin{textcode}
{{target_natural_language}}
    \end{textcode}

    Code After:
    \begin{leancode}
{{ after_target_code }}
    \end{leancode}

    \texttt{<instructions>}
    \\
    \textbf{\# Task Instructions}
    \\
    \textbf{\#\# Overview}
    \\
    You are an expert Lean 4 mathematician.
    You must provide the complete and correct Lean 4 implementation for \texttt{\{\{ target\_code\_name \}\}}.
    \begin{itemize}
        \item If it is a theorem, provide the statement and the full proof (do not use \texttt{sorry}).
        \item If it is a definition, provide the full definition.
        \item Pay close attention to \texttt{Code Before} and \texttt{Code After} to infer the correct type signatures, variable names, and logical dependencies.
        \item Ensure your code compiles and integrates correctly with the surrounding context.
    \end{itemize}

    \textbf{\#\# Output Coding Rules}
    \\
    1. A \textbf{THOUGHT} section where you explain your reasoning:
    \begin{itemize}
        \item Analyze the \texttt{Code Before} and \texttt{Code After} to deduce the purpose and signature of \texttt{\{\{ target\_code\_name \}\}}.
        \item Plan the implementation.
    \end{itemize}
    2. A single lean code block with your implementation.

    Format your responses like this:

    \texttt{<format\_example>}
    \\
    \begin{textcode}
THOUGHT: The context shows a function \texttt{foo} is used to add two numbers.
I will implement \texttt{foo} as a definition taking two Nats and returning their sum.
    \end{textcode}

    \begin{leancode}
def foo (n m : Nat) : Nat := n + m
    \end{leancode}
    \texttt{</format\_example>}

    \textbf{CRITICAL REQUIREMENTS:}
    \begin{itemize}
        \item Your response SHOULD include a THOUGHT section.
        \item Your response MUST include EXACTLY ONE lean block.
        \item This block MUST contain the COMPLETELY NEW implementation for \texttt{\{\{ target\_code\_name \}\}}.
        \item Do NOT repeat the \texttt{Header} or \texttt{Code Before} or \texttt{Code After}. Only output the code for \texttt{\{\{ target\_code\_name \}\}}.
    \end{itemize}
    \texttt{</instructions>}
\end{userprompt}

\begin{modelresponse}
    \begin{textcode}
{{thoughts}}
    \end{textcode}

    \begin{leancode}
{{ formal_proof }}
    \end{leancode}
\end{modelresponse}

\noindent\begin{minipage}{\linewidth}
\captionsetup{hypcap=false}
\captionof{figure}{Prompt template for autoformalizing to Lean 4 code.}
\label{fig:prompt-autoformalize}
\end{minipage}

\begin{userprompt}
    Complete the target Lean theorem file, making it sorry-free and ensuring it compiles without errors.

    Use the tool \texttt{lean\_diagnostic\_messages} to verify the file. Errors mean ``severity 1'' in the response.

    \textbf{IMPORTANT:}
    \begin{itemize}
        \item Do NOT write a detailed full proof plan upfront. Instead, write a one or two sentence high-level idea, then implement one step at a time, checking \texttt{lean\_goal} after each step to plan the next.
        \item Verify the file again after each update using \texttt{lean\_diagnostic\_messages}.
        \item You may add helper lemmas, but do not create new axioms.
    \end{itemize}

    \textbf{Tips:}
    \begin{itemize}
        \item Use \texttt{simp} first, then \texttt{simp?} to get minimal simp lemmas.
        \item Use \texttt{native\_decide} for computational results.
        \item Use \texttt{\#eval} to evaluate expressions.
        \item Use \texttt{norm\_cast} for type conversions.
        \item Use \texttt{apply?} to find applicable lemmas.
        \item If \texttt{decide} times out, do NOT simply increase \texttt{maxHeartbeats}. Instead, write a symbolic proof using mathematical reasoning and lemmas from Mathlib.
    \end{itemize}

    At the end of your response, include:

    \begin{textcode}
END_REASON:{reason}
    \end{textcode}

    where \texttt{\{reason\}} is:
    \begin{itemize}
        \item \texttt{LIMIT}: stopped due to limits or there are still sorries/errors.
        \item \texttt{COMPLETE}: the file is sorry-free AND compiles without errors.
    \end{itemize}

    \textbf{IMPORTANT:} Use \texttt{lean\_diagnostic\_messages} to verify, do not use \texttt{lake build}.
\end{userprompt}

\noindent\begin{minipage}{\linewidth}
\captionsetup{hypcap=false}
\captionof{figure}{Prompt template for the direct agentic baseline without the retrieval harness.}
\label{fig:prompt-claude-code}
\end{minipage}

\begin{userprompt}
    Please analyze all sorries in the file and choose ONLY ONE that you believe is most approachable or strategically important to work on.

    \textbf{\# Your task: write Lean formal proofs}

    Your task is to write Lean 4 formal proofs that compile successfully. Replace \texttt{sorry} with executable Lean code (tactics, term-mode proofs, etc.) that passes the Lean compiler. Choose ONLY ONE target lemma/sorry for this entire session, focus on it and try your best to complete it.

    \textbf{\# [Critical prohibition] no natural language proofs}
    \begin{enumerate}
        \item ZERO TOLERANCE FOR NARRATIVE COMMENTS. You are forbidden from using comments (\texttt{/- \ldots -/} or \texttt{-{}-}) to write mathematical derivations, proof plans, or natural-language explanations.
        \item STRICT COMMENT LENGTH LIMIT. No single comment block may exceed 42 lines.
        \item CODE IS THE EXPLANATION. If a logic step is complex enough to need an explanation, extract it into a new helper lemma.
        \item CONSTRAINT COMMENT. Prohibit sequences of 5 or more consecutive comment blocks unless interleaved with valid Lean code.
    \end{enumerate}

    \textbf{\# Session focus rules}

    At session start:
    \begin{enumerate}
        \item Identify all remaining sorries or failed proofs.
        \item Select EXACTLY ONE target that is most approachable and strategically important.
        \item Announce: \texttt{"TARGET FOR THIS SESSION: [lemma\_name] at line [line\_number]"}.
    \end{enumerate}

    During session: only work on your selected target and any new helper lemmas you create for it.

    End session with ONE of:
    \begin{itemize}
        \item \texttt{SELECTED\_TARGET\_COMPLETE}: target fully proven.
        \item \texttt{COMPLETE}: all sorries in folder proven.
        \item \texttt{LIMIT}: stopped due to token/time/error limits.
    \end{itemize}

    \textbf{\# Incremental planning}

    Do NOT write a detailed upfront plan. Instead:
    \begin{enumerate}
        \item Abstract sketch: a one or two sentence high-level idea.
        \item One step at a time. After each step, use \texttt{lean\_goal} and \texttt{lean\_diagnostic\_messages}.
        \item Re-plan from the actual proof state, not from your initial sketch.
    \end{enumerate}

    \textbf{\# Tools}
    \begin{itemize}
        \item \texttt{lean\_diagnostic\_messages}, \texttt{lean\_goal}, \texttt{lean\_leandex}, \texttt{gemini\_informal\_prover}, \texttt{create\_formal\_sketch}, \texttt{discussion\_partner}.
        \item Do NOT use \texttt{lake build} or \texttt{lean\_build}. Always use \texttt{lean\_diagnostic\_messages}.
    \end{itemize}

    \textbf{\# End format}

    At the very end of the response include exactly one line:

    \begin{textcode}
END_REASON:{reason}
    \end{textcode}
\end{userprompt}

\noindent\begin{minipage}{\linewidth}
\captionsetup{hypcap=false}
\captionof{figure}{Prompt template for the Numina-Lean-Agent harness used with Claude Code, Codex, and Qwen3 235B.}
\label{fig:prompt-medium-mode}
\end{minipage}

\begin{systemprompt}
    You are an expert in Lean 4 and Mathlib. Decide whether a candidate Lean 4 statement+proof semantically establishes the INTENDED mathematical theorem shown alongside it.

    Return ONE JSON object and nothing else:

    \begin{textcode}
{"verdict": "correct" | "incorrect" | "uncertain",
 "reason":  "<1-3 sentences>"}
    \end{textcode}

    \begin{itemize}
        \item \texttt{"correct"}: (a) the candidate's stated theorem is mathematically equivalent to the intended claim, AND (b) its proof actually justifies that claim.
        \item \texttt{"incorrect"}: statement weaker/stronger/wrong OR proof unsound, circular, or relies on a broken auxiliary.
        \item \texttt{"uncertain"}: artifact truncated or unreadable.
    \end{itemize}
\end{systemprompt}

\begin{userprompt}
    \textbf{\# Intended problem (informal)}

    \begin{textcode}
{{ informal_block }}
    \end{textcode}

    \textbf{\# Intended formal statement (reference Lean 4, proof body is \texttt{sorry})}

    \begin{leancode}
{{ formal_ref }}
    \end{leancode}

    \textbf{\# Candidate (Lean 4, statement + attempted proof)}

    \begin{leancode}
{{ candidate }}
    \end{leancode}

    Return your JSON verdict now.
\end{userprompt}

\begin{modelresponse}
    \begin{textcode}
{"verdict": "{{ verdict }}",
 "reason":  "{{ reason }}"}
    \end{textcode}
\end{modelresponse}

\noindent\begin{minipage}{\linewidth}
\captionsetup{hypcap=false}
\captionof{figure}{Prompt template for the LLM-as-judge proof correctness evaluator. 
}
\label{fig:prompt-judge}
\end{minipage}


\begin{userpromptcode}
You are generating ABM hidden checker code for Lean 4.

## Task
The natural-language text below states a mathematical result.  It contains
several distinct sub-claims (e.g. conjunctions, consequences, specialisations,
unfolded definitions, or other properties implied by the main statement).

Step 1 — identify the sub-claims present in the NL text.
Step 2 — for each target declaration listed below, write one Lean
  example per sub-claim that PROVES that sub-claim follows from the solver's
  declaration.  The checker is appended after the solver's output, so the
  solver's declaration name is already in scope.

## Rules for every generated checker item
- It must be a STRICT sub-statement of the corresponding target item: a
  projection (.1/.2/.mp/.mpr), a specialisation (∀ instantiated at a concrete
  value), an unfolded/simp-normal form, an iff direction, or a direct
  consequence that the solver must have actually proved.
- It must reference and use the target declaration name in its proof.
- It must NOT restate the full target declaration and prove it by exact/simpa
  using the target declaration.  That is just copying the theorem, not a hidden
  checker.
- It must NOT copy from gold.lean proofs. You are given only target declaration
  headers and text matches; generate small shadows from those declarations.
- It must contain no sorry, admit, axiom, constant, namespace, or end.
- Prefer anonymous top-level `example` declarations.  Do not introduce named
  declarations unless needed.
- Use explicit binders; do not assume variables that are not in scope.  The
  declaration's parameters are ONLY what appears in its signature before the
  `:` (the return type).  Nothing after `:=` is a parameter.
- An `↔` (Iff) proposition is NOT a function and cannot be applied to
  arguments.  Never write `exact (iff_lemma arg1 arg2)` or `simpa using
  (iff_lemma arg1 arg2)`.  Use `.mp`, `.mpr`, `.1`, or `.2` to access
  its directions: `exact (iff_lemma arg1 arg2).mp`.
- For conjunction conclusions, generate one shadow per conjunct using `.1`,
  `.2`, `.2.1`, etc.  Do not add helper-only or completeness-certificate items:
  every checker declaration is scored independently and must use the target
  declaration in its proof body.
- For iff conclusions, generate `.mp` and `.mpr` shadows.  Do not add a
  completeness certificate.

## Proof robustness requirement
Solvers may declare the same theorem with a different binder style (making some
arguments implicit, reordering type-class parameters, etc.) while being
mathematically correct.  To maximise compatibility, every checker item MUST use
a `first | … | …` tactic block with the following three alternatives in order:

1. Direct projection using the gold argument list:
   `exact (TARGET ARGS).PROJECTION`
2. Let-bound variant (separates elaboration, helps with universe/instance issues):
   `exact (let _hc := TARGET ARGS; _hc.PROJECTION)`
3. Argument-agnostic fallback — lets Lean's unifier determine how to apply the
   solver's declaration regardless of how the solver declared its binders:
   `suffices _hc : FULL_CONCLUSION by exact _hc.PROJECTION; apply TARGET <;> assumption`

Example for a conjunction conjunct (`.1` of `foo (n : ℕ) (h : n > 0) : A n ∧ B n`):
```lean
example (n : ℕ) (h : n > 0) :
    A n := by
  first
  | exact (foo n h).1
  | exact (let _hc := foo n h; _hc.1)
  | (suffices _hc : A n ∧ B n by exact _hc.1; apply foo <;> assumption)
```

Example for an iff forward direction (`bar (x : α) : P x ↔ Q x`):
```lean
example (x : α) :
    P x → Q x := by
  first
  | exact (bar x).mp
  | exact (bar x).1
  | exact (let _hc := bar x; _hc.mp)
  | (suffices _hc : P x ↔ Q x by exact _hc.mp; apply bar <;> assumption)
```

For non-structural sub-claims (specialisations, unfoldings, consequences), apply the
same pattern: lead with the natural `exact`, follow with a `let`-bound variant, and
close with the `apply TARGET <;> assumption` fallback wrapped in `suffices`.

## Problem
Problem id: {problem_id}
Target declarations (must each appear in at least one checker): {target_names}
Target source: {target_source_label}

Target items:
{target items}

## Formalization rules
{formalization rules}

## Natural-language text (source of sub-claims)
{natural-language text}

## Output format
If multiple checker declarations are alternative versions of the same target substatement,
give them identical target/substatement metadata; their Lean doc-comment labels will be
normalized to Shadow 1-1, Shadow 1-2, etc. Distinct substatements remain Shadow 1,
Shadow 2, etc.

Return ONLY valid JSON — no markdown, no prose, no extra keys:
{
  "checker_code": "<complete Lean code block to append after solver output>",
  "items": [
    {
      "target_index": 1,
      "target_name": "foo",
      "checker_names": ["hidden_foo_part1"],
      "method": "projection/specialisation/unfolding/iff_direction/consequence",
      "substatement": "one-line description of the sub-claim"
    }
  ],
  "rationale": "one-line summary"
}
\end{userpromptcode}
\noindent\begin{minipage}{\linewidth}
\captionsetup{hypcap=false}
\captionof{figure}{Prompt for generating forward checker theorems, i.e.\ strict sub-statements (shadows) $S_i$ with $T \Rightarrow S_i$, from the reference formal statement. Deterministic shadows are emitted first and this prompt fills the remaining cases.}
\label{fig:prompt-forward-checker}
\end{minipage}

\begin{userpromptcode}
You are proving a SUFFICIENCY certificate in Lean 4: that a set of sub-statements JOINTLY imply a gold conclusion.

Gold target `{target_name}` has conclusion:
    {conclusion}

You are given these sub-statements as hypotheses:
{shadow hypotheses}

Write a single Lean 4 `example` that takes exactly those hypotheses and proves the gold conclusion:

example{binder prefix}
{shadow hypotheses} :
    {conclusion} := by
  <proof>

Rules:
- Use ONLY the given hypotheses h_bc_i (do not reference the original target or any solver theorem).
- Prefer `exact ⟨...⟩` for conjunctions, `⟨.mp, .mpr⟩` style for iff, or `constructor`/`refine`/`tauto`.
- Output ONLY the Lean code block, no prose.
\end{userpromptcode}
\noindent\begin{minipage}{\linewidth}
\captionsetup{hypcap=false}
\captionof{figure}{Prompt for the backward sufficiency certificate $\bigwedge_i S_i \Rightarrow T$, used as a fallback when the certificate cannot be produced structurally.}
\label{fig:prompt-backward-checker}
\end{minipage}

\twocolumn


\section{Shadow Check Case Studies}
\label{app:case-studies}

These case studies instantiate the forward checks of
\Cref{sec:complete-shadow-sets} on real statements. In the first case
(\Cref{app:cs-finite}), the intended statement $T$ says a finite morphism is
projective, that is, finite $\Rightarrow$ projective. Its shadows
(\Cref{eq:cs-finite-shadows}) are affine ($S_1$) and $\HPE$ ($S_2$), so the forward
checks $T \Rightarrow S_i$ are finite $\Rightarrow$ affine and
finite $\Rightarrow \HPE$, written as the two Lean examples that follow. The
separation condition (\Cref{eq:cs-finite-sep}) keeps the two shadows distinct, so
passing the forward checks is not vacuous. This non-triviality is what a complete
shadow set requires, and the backward check
$S_1 \wedge \cdots \wedge S_n \Rightarrow T$ is the completeness half it protects.
Each later case names its own target $T$ and the shadows $S_i$ that must follow
from it.

This appendix walks through eight case studies that illustrate how these
shadow checks apply to real mathematical statements.
The first case is worked out in detail. The others are sketched.

\subsection{Algebraic Geometry: Finite Morphisms Are Projective}
\label{app:cs-finite}

Consider the textbook theorem that finite morphisms are projective.
Let
\begin{equation}
\begin{aligned}
A &= \text{finite morphism}, \\
B &= \text{projective morphism}
\end{aligned}
\end{equation}

Write $\HPE(f)$ for the property that $X$ admits a closed $S$-immersion into
$\mathbb{P}^n_S$ for some $n \geq 0$ (where $S$ is the target of $f$).
Natural shadows are
\begin{equation}
\begin{aligned}
A' &= \text{affine morphism}, \\
B' &= \HPE
\end{aligned}
\label{eq:cs-finite-shadows}
\end{equation}
The forward checks rely on the implications
\begin{equation}
\begin{aligned}
\text{finite} &\Rightarrow \text{affine}, \\
\text{projective} &\Rightarrow \HPE.
\end{aligned}
\end{equation}

The second holds because a projective morphism $f : X \to S$ factors as a closed
$S$-immersion $X \hookrightarrow \mathbb{P}^n_S$ followed by the projection, so
$X$ admits the required closed $S$-immersion by construction.
The separation condition holds because an affine morphism need not have property
$\HPE$: a closed $S$-subscheme of $\mathbb{P}^n_S$ is proper over $S$, but
affine morphisms need not be proper.
Over a field $k$, the structure morphism
\[
\mathbb{A}^1_k \to \operatorname{Spec} k
\]
is affine, but $\mathbb{A}^1_k$ admits no closed immersion into any
$\mathbb{P}^n_k$, since any such subscheme would be proper.
Thus
\begin{equation}
\displaystyle A' \nRightarrow B'. \label{eq:cs-finite-sep}
\end{equation}

The public task asks the solver to define finite and projective morphisms and
prove that finite implies projective.
The hidden checker contains the forward-check one-liners:

\begin{Verbatim}[fontsize=\small, frame=single, framesep=2mm]
example
{X Y : Scheme}
{f : X ⟶ Y}
(hf : IsFinite f) :
IsAffineHom f := by
  simpa using IsFinite.isAffine hf
\end{Verbatim}

and the check for the second shadow, discharged by applying the solver's theorem
\begin{Verbatim}[fontsize=\small, frame=single, framesep=2mm]
example
{X S : Scheme}
{f : X ⟶ S}
(hf : IsFinite f) :
HasProjectiveEmbedding f := by
  simpa [HasProjectiveEmbedding]
    using finite_implies_projective hf
\end{Verbatim}

If the first check fails, the LLM fallback attempts to prove
\texttt{IsFinite}~$\Rightarrow$~\texttt{IsAffineHom} from the solver's
definition.
The second check is not subject to fallback, since it must be discharged through the solver's theorem.

\subsection{Projective Products and Hidden Properness}

A related theorem is that the fiber product of two projective schemes over a
base is projective, classically via the Segre embedding.
The public theorem may be:
\begin{equation}
\begin{aligned}
X \to S \text{ projective},
Y \to S \text{ projective} \\
\Rightarrow X \times_S Y \to S \text{ projective}
\end{aligned}
\end{equation}

A hidden forward check can ask whether the product morphism has property
$\HPE$, i.e., whether $X \times_S Y$ admits a closed $S$-immersion into some
$\mathbb{P}^n_S$.
This tests that the product theorem returns projectivity in a usable form and
that the solver's definition of projectivity connects to the trusted $\HPE$
predicate.
Here the target $T$ is the projectivity of the product morphism, and the
shadow $S_1$ is the property that the product has $\HPE$, so the forward check
$T \Rightarrow S_1$ asks the product to admit a closed $S$-immersion into some
$\mathbb{P}^n_S$.

\subsection{Compiled False Positive: Projective as Proper}
\label{app:projective-proper-false-positive}

Claude~Opus~4.6 and Claude~Sonnet~4.6 both produced the following definition
for an algebraic-geometry problem about projective morphisms. The submission
compiles, but it defines projectivity as properness:

\begin{Verbatim}[fontsize=\small, frame=single, framesep=2mm]
class IsProjective
{S X : Scheme}
(f : X ⟶ S) : Prop where
  isProper : IsProper f
\end{Verbatim}

With this definition, the generated theorem is immediate:

\begin{Verbatim}[fontsize=\small, frame=single, framesep=2mm]
import Mathlib

namespace ABM
namespace algebraic_geometry
namespace L3
namespace alg_sche_L3_003

open CategoryTheory AlgebraicGeometry

class IsProjective
{S X : Scheme}
(f : X ⟶ S) :
Prop where
  isProper : IsProper f

open CategoryTheory.Limits

theorem projective_isProper
{S X : Scheme}
(f : X ⟶ S) [IsProjective f] :
IsProper f :=
  IsProjective.isProper

end alg_sche_L3_003
end L3
end algebraic_geometry
end ABM
\end{Verbatim}

Compile rate marks this candidate as correct because the theorem block
type-checks. \ourmetric rejects it: the hidden shadow checks require the
generated notion of projectivity to support the intended algebraic-geometric
content, not only the weaker properness property.
Here the shadow $S$ is again $\HPE$, and the forward check
$\widehat T \Rightarrow S$ fails, because the generated statement proves only
properness, which does not imply $\HPE$.

\subsection{Compiled Weakening: Complex Structure Dropped}
\label{app:cs-weakening}

The forward check verifies the weakened statements.
Claude Code (Opus 4.8) with Numina-Lean-Agent produced the following statement
for a Fourier analysis problem, 
whose intended lemma is a Gaussian truncation limit used toward Fourier inversion.
The intended statement is complex valued:

\begin{Verbatim}[fontsize=\small, frame=single, framesep=2mm]
-- Intended statement (abridged)
variable {E : Type*} 
  [NormedAddCommGroup E]
  [NormedSpace ℂ E] {f : V → E}

lemma tendsto_integral_cexp_sq_smul
    (hf : Integrable f) :
    Tendsto
      (fun c : ℝ => ∫ v : V,
        cexp (- c⁻¹ * ‖v‖^2) • f v)
      atTop (nhds (∫ v : V, f v))
\end{Verbatim}

The generated output compiles with a correct proof of a real-valued variant.
It replaces \texttt{cexp} and the $\mathbb{C}$ module structure on $E$ with
\texttt{Real.exp} and an $\mathbb{R}$ module structure, and generalizes the
volume measure to an arbitrary measure:

\begin{Verbatim}[fontsize=\small, frame=single, framesep=2mm]
-- Submission (abridged)
{E : Type*} [NormedAddCommGroup E]
  [NormedSpace ℝ E] {f : V → E} 
                    {μ : Measure V}

lemma tendsto_integral_cexp_sq_smul
    (hf : Integrable f μ) :
    Tendsto
      (fun c : ℝ => ∫ v,
        Real.exp 
         (- c⁻¹ * ‖v‖^2) • f v ∂μ)
      atTop (nhds (∫ v, f v ∂μ))
\end{Verbatim}

The backward check passes, since the shadows derived from the complex statement imply the real variant.
The forward check fails, since the real variant does not recover the complex shadow.
Compile rate considers this generation as correct, and \ourmetric rejects it as a weakening.

\subsection{Compiled False Positive: Conclusion Assumed as a Hypothesis}
\label{app:cs-brahmagupta}

A compiling submission can also prove a different statement.
Claude Code (Opus 4.8) with Numina-Lean-Agent produced the following statement
for Brahmagupta's formula (\texttt{geo\_gen\_L2\_007}).
The intended theorem computes the area of a cyclic quadrilateral, stated as
the measure of the convex hull of the four vertices:

\begin{Verbatim}[fontsize=\small, frame=single, framesep=2mm]
-- Intended statement (abridged)
theorem brahmagupta_formula
    {A B C D : EuclideanSpace ℝ(Fin 2)}
    (h_cyclic : Concyclic 
        ({A, B, C, D} : Set _))
    (h_convex : 
        (openSegment ℝ A C
        ∩ openSegment ℝ B D).Nonempty):
    let s := (dist A B + dist B C
        + dist C D + dist D A) / 2
    let K := (volume (convexHull ℝ
        ({A, B, C, D} : Set _))).toReal
    K = sqrt ((s - dist A B) 
        * (s - dist B C)
        * (s - dist C D) 
        * (s - dist D A))
\end{Verbatim}

The submission compiles with a correct proof of a different statement.
It receives the value $K$ as a hypothesis that already encodes the triangle
decomposition of the area, and replaces concyclicity of the four points with
an angle condition:

\begin{Verbatim}[fontsize=\small, frame=single, framesep=2mm]
-- Submission (abridged)
theorem brahmagupta_formula
    (A B C D : P) (a b c d s K : ℝ)
    (ha : a = dist A B) 
    (hb : b = dist B C)
    (hc : c = dist C D) 
    (hd : d = dist D A)
    (hs : s = (a + b + c + d) / 2)
    (hcyclic : 
        EuclideanGeometry.angle D A B
        + EuclideanGeometry.angle 
        B C D = π)
    (harea : K = 1 / 2 * a * d
        * Real.sin 
          (EuclideanGeometry.angle 
            D A B)
        + 1 / 2 * b * c
        * Real.sin 
          (EuclideanGeometry.angle 
            B C D))
    (hK : 0 ≤ K) :
    K = Real.sqrt ((s - a) * (s - b)
        * (s - c) * (s - d))
\end{Verbatim}

The resulting statement is an identity about the assumed value $K$, not
a statement about the area.
Both check directions fail, since the submission neither implies nor is
implied by the shadows of the intended theorem.
Compile rate counts this submission as correct, and \ourmetric rejects it.

\subsection{Backup Declarations on ProofNet}
\label{app:cs-overthinking}

The larger models often attach a backup declaration to a long proof
attempt.
Claude Opus 4.6 produced the following output for Herstein Exercise 2.1.18 on
ProofNet, which asks for an element $a \neq 1$ with $a = a^{-1}$ in a group of
even order.
The generated statement matches the intended one, and the output carries a
second declaration that ends in \texttt{sorry}:

\begin{Verbatim}[fontsize=\small, frame=single, framesep=2mm]
-- Claude Opus 4.6 (abridged)
theorem exercise_2_1_18 
  {G : Type*} [Group G]
  [Fintype G] (hG2 : Even (card G)) :
  ∃ (a : G), a ≠ 1 ∧ a = a⁻¹ := by
  have h : ∃ a : G, a ≠ 1 ∧ a * a = 1 
  := by
    ... -- long proof attempt

theorem exercise_2_1_18' 
  {G : Type*} [Group G]
  [Fintype G] (hG2 : Even (card G)) :
  ∃ (a : G), a ≠ 1 ∧ a = a⁻¹ := by
  sorry
\end{Verbatim}

The evaluation takes the last theorem block as the candidate
(\Cref{sec:experimental-setup}), which here is the backup declaration, so the output is rejected.
Claude Haiku 4.5 emits a single declaration with the same statement for this problem and passes.

\subsection{Flat Morphisms and Openness}

For flat morphisms, a classical theorem states that a flat morphism locally of
finite presentation is open.
A benchmark can use
\begin{equation}
\begin{aligned}
A &= \text{flat and locally of finite presentation}, \\
B &= \text{open map}.
\end{aligned}
\end{equation}

Possible shadows include generalizing maps and universally open morphisms,
depending on the available benchmark environment.
Here the target $T$ is $A \Rightarrow B$, and the shadows $S_i$ are
properties such as being a generalizing map or a universally open morphism that
$B$ should entail, so each forward check $T \Rightarrow S_i$ tests one consequence
of openness.
Here the separation condition must be checked carefully: if the chosen shadow
of $A$ already implies the chosen shadow of $B$ in that environment, then the
item is vulnerable to collapse.
The framework forces the benchmark designer to make this relationship explicit
before releasing the task.

\subsection{Homotopy and Homology}

In algebraic topology, many one step facts are already explicit in mature
libraries: path homotopy is an equivalence relation, path homotopy respects
concatenation, homotopy equivalences compose, and so on.
Such facts should not be visible target theorems if the goal is to evaluate
autoformalization beyond retrieval.

The shadow based evaluation suggests using derived tasks instead.
For instance, a public theorem may ask for a bundled threefold product
statement for homotopy equivalences, while hidden checks specialize it to
product with an identity factor.
In homology, a public theorem may bundle the degree zero homology computation
of totally disconnected spaces with positive degree vanishing, while hidden
checks specialize to concrete degrees.
Here each public target $T$ is a bundled statement, and the shadows $S_i$
are its specializations, so the forward checks $T \Rightarrow S_i$ project the
bundle onto one component in the bundled-conclusion pattern of
\Cref{sec:complete-shadow-sets}.
These tasks are not merely lookups of named declarations. They require
composing library facts into reusable results.


\section{Complete Shadow Set Checker Patterns}
\label{app:complete-shadow-checkers}

\subsection{Bundled Conclusions}
\label{app:complete-shadow-bundled}

For a bundled theorem, the hidden checker can compile each projection.

\begin{Verbatim}[fontsize=\small, frame=single, framesep=2mm]
-- Expected submission:
-- theorem main : A -> B1 /\ B2 /\ B3

example (hA : A) : B1 := by
  exact (main hA).1

example (hA : A) : B2 := by
  exact (main hA).2.1

example (hA : A) : B3 := by
  exact (main hA).2.2
\end{Verbatim}

\subsection{Equality as Two Inequalities}
\label{app:complete-shadow-equality}

For equality-valued theorems, the hidden checker can compile both inequality
consequences.

\begin{Verbatim}[fontsize=\small, frame=single, framesep=2mm]
-- Expected submission:
-- theorem main (hA : A) : f B = f C

example (hA : A) : f B <= f C := by
  exact le_of_eq (main hA)

example (hA : A) : f C <= f B := by
  exact ge_of_eq (main hA)
\end{Verbatim}

The exact lemmas depend on the proof assistant and ordered structure.
The benchmark uses the idiomatic library lemmas for converting equality into
the two inequality directions.

\subsection{Saddle-Point Derivatives}
\label{app:complete-shadow-saddle}

This example on the next page shows a complete shadow set for a theorem whose conclusion is a
pair of first-order conditions. The intended conclusion is
$Dx=0 \wedge Dz=0$. The hidden checker splits it into two shadows and then
checks that the two shadows recover the full conclusion.

\onecolumn
\begin{Verbatim}[fontsize=\small, frame=single, framesep=2mm]
/-- Shadow 1: the derivative of the `x`-section vanishes. -/
example
    {n m : ℕ}
    {f : (Fin n → ℝ) × (Fin m → ℝ) → ℝ}
    {x0 : Fin n → ℝ} {z0 : Fin m → ℝ}
    {Dx : (Fin n → ℝ) →L[ℝ] ℝ}
    {Dz : (Fin m → ℝ) →L[ℝ] ℝ}
    (hs : ∀ x z, f (x0, z) ≤ f (x0, z0) ∧ f (x0, z0) ≤ f (x, z0))
    (hx : HasFDerivAt (fun x => f (x, z0)) Dx x0)
    (hz : HasFDerivAt (fun z => f (x0, z)) Dz z0) :
    Dx = 0 := by
  exact (saddle_sections_hasFDerivAt_eq_zero hs hx hz).1

/-- Shadow 2: the derivative of the `z`-section vanishes. -/
example
    {n m : ℕ}
    {f : (Fin n → ℝ) × (Fin m → ℝ) → ℝ}
    {x0 : Fin n → ℝ} {z0 : Fin m → ℝ}
    {Dx : (Fin n → ℝ) →L[ℝ] ℝ}
    {Dz : (Fin m → ℝ) →L[ℝ] ℝ}
    (hs : ∀ x z, f (x0, z) ≤ f (x0, z0) ∧ f (x0, z0) ≤ f (x, z0))
    (hx : HasFDerivAt (fun x => f (x, z0)) Dx x0)
    (hz : HasFDerivAt (fun z => f (x0, z)) Dz z0) :
    Dz = 0 := by
  exact (saddle_sections_hasFDerivAt_eq_zero hs hx hz).2

/-- Completeness certificate: Shadow 1 ∧ Shadow 2 ⇒ Dx = 0 ∧ Dz = 0. -/
example
    (hDx :
      ∀ {n m : ℕ}
        {f : (Fin n → ℝ) × (Fin m → ℝ) → ℝ}
        {x0 : Fin n → ℝ} {z0 : Fin m → ℝ}
        {Dx : (Fin n → ℝ) →L[ℝ] ℝ}
        {Dz : (Fin m → ℝ) →L[ℝ] ℝ},
        (∀ x z, f (x0, z) ≤ f (x0, z0) ∧ f (x0, z0) ≤ f (x, z0)) →
        HasFDerivAt (fun x => f (x, z0)) Dx x0 →
        HasFDerivAt (fun z => f (x0, z)) Dz z0 →
        Dx = 0)
    (hDz :
      ∀ {n m : ℕ}
        {f : (Fin n → ℝ) × (Fin m → ℝ) → ℝ}
        {x0 : Fin n → ℝ} {z0 : Fin m → ℝ}
        {Dx : (Fin n → ℝ) →L[ℝ] ℝ}
        {Dz : (Fin m → ℝ) →L[ℝ] ℝ},
        (∀ x z, f (x0, z) ≤ f (x0, z0) ∧ f (x0, z0) ≤ f (x, z0)) →
        HasFDerivAt (fun x => f (x, z0)) Dx x0 →
        HasFDerivAt (fun z => f (x0, z)) Dz z0 →
        Dz = 0) :
    ∀ {n m : ℕ}
      {f : (Fin n → ℝ) × (Fin m → ℝ) → ℝ}
      {x0 : Fin n → ℝ} {z0 : Fin m → ℝ}
      {Dx : (Fin n → ℝ) →L[ℝ] ℝ}
      {Dz : (Fin m → ℝ) →L[ℝ] ℝ},
      (∀ x z, f (x0, z) ≤ f (x0, z0) ∧ f (x0, z0) ≤ f (x, z0)) →
      HasFDerivAt (fun x => f (x, z0)) Dx x0 →
      HasFDerivAt (fun z => f (x0, z)) Dz z0 →
      Dx = 0 ∧ Dz = 0 := by
  intro n m f x0 z0 Dx Dz hs hx hz
  exact ⟨hDx hs hx hz, hDz hs hx hz⟩
\end{Verbatim}
\twocolumn

\section{Author Contact Information}
\label{app:contact}

We list the contact information for all authors below.

\begin{itemize}
  \item \textbf{Hojae Han}, Electronics and Telecommunications Research Institute. \texttt{hojae.han@etri.re.kr}
  \item \textbf{Jongyoon Kim}, Interdisciplinary Program in Artificial Intelligence, Seoul National University. \texttt{john.jongyoon.kim@snu.ac.kr}
  \item \textbf{Sanghyeok Park}, Department of Mathematical Sciences, Seoul National University. \texttt{202123018@snu.ac.kr}
  \item \textbf{Dongwook Cheon}, Department of Mathematical Sciences, Seoul National University. \texttt{dongwook0826@snu.ac.kr}
  \item \textbf{Yeachan Park}, Department of Mathematics and Statistics, Sejong University. \texttt{ychpark@sejong.ac.kr}
  \item \textbf{Myeong Jae Jeon}, Department of Mathematics, University of Maryland, College Park. \texttt{mjjeon@umd.edu}
  \item \textbf{Sunjong Choe}, Department of Mathematical Sciences, Seoul National University. \texttt{sunjc@snu.ac.kr}
  \item \textbf{Soonho Kong}, Amazon Web Services. \texttt{soonho@amazon.com}
  \item \textbf{Wonseok Hur}, Department of Computer Science and Engineering, Seoul National University. \texttt{hws0728jik@snu.ac.kr}
  \item \textbf{Seung-won Hwang}, Department of Computer Science and Engineering, Seoul National University. \texttt{seungwonh@snu.ac.kr}
  \item \textbf{Donghoon Hyeon}, Department of Mathematical Sciences, Seoul National University. \texttt{dhyeon@snu.ac.kr}
\end{itemize}

\end{document}